\documentclass[10pt,twocolumn,letterpaper]{article}

\usepackage{iccv}
\usepackage{times}
\usepackage{epsfig}
\usepackage{graphicx}
\usepackage{amsmath}
\usepackage{amssymb}

\usepackage{titling}
\usepackage[font={small}]{caption}
\usepackage[font={small}]{subcaption}

\usepackage[breaklinks=true,bookmarks=false]{hyperref}

\iccvfinalcopy % *** Uncomment this line for the final submission

\def\iccvPaperID{1921} % *** Enter the ICCV Paper ID here

\begin{document}

%%%%%%%%% TITLE
\title{Stereo DSO: \\Large-Scale Direct Sparse Visual Odometry with Stereo Cameras}
\author{Rui Wang\thanks{These authors contributed equally.} , Martin Schw{\"o}rer$^{\ast} $, Daniel Cremers\\
Technical University of Munich\\
{\tt\small\{wangr, schwoere, cremers\}@in.tum.de}}

\date{\vspace{-3ex}}
\maketitle

%%%%%%%%% ABSTRACT
\begin{abstract}
We propose Stereo Direct Sparse Odometry (Stereo DSO) as a novel
method for highly accurate real-time visual odometry estimation of
large-scale environments from stereo cameras. It jointly optimizes
for all the model parameters within the active window, including the
intrinsic/extrinsic camera parameters of all keyframes and the depth
values of all selected pixels. In particular, we propose a novel
approach to integrate constraints from static stereo into the bundle
adjustment pipeline of temporal multi-view stereo. Real-time
optimization is realized by sampling pixels uniformly from image
regions with sufficient intensity gradient. Fixed-baseline stereo
resolves scale drift. It also reduces the sensitivities to large
optical flow and to rolling shutter effect which are known
shortcomings of direct image alignment methods. Quantitative
evaluation demonstrates that the proposed Stereo DSO outperforms
existing state-of-the-art visual odometry methods both in terms of
tracking accuracy and robustness. Moreover, our method delivers a
more precise metric 3D reconstruction than previous dense/semi-dense
direct approaches while providing a higher reconstruction density
than feature-based methods. 

\end{abstract}

%%%%%%%%% BODY TEXT
\section{Introduction}
 
\subsection{Real-time Visual Odometry}
\begin{figure}[t]
\begin{center}
 \includegraphics[width=1.0\linewidth]{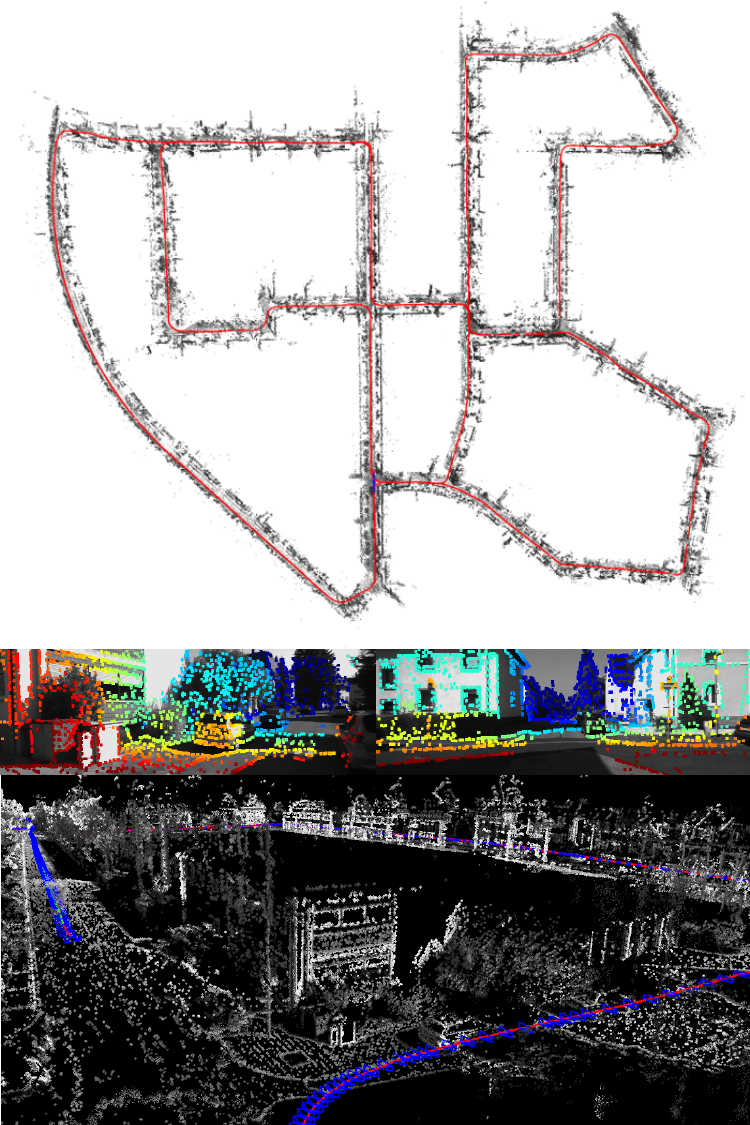}
\end{center}
\vspace{-0.7em}
\caption{Results of Stereo Direct Sparse Odometry on sequence 00 of
  the KITTI Dataset. On top is the estimated camera trajectory. To
  distinguish the sparsity from previous dense or semi-dense direct
  methods, we show the depth maps used for tracking in the middle. An example of a reconstructed scene is shown at bottom.}
\vspace{-1.2em}
\end{figure}
While traditionally robotic systems such as self-driving cars have
been largely relying on Laser or Lidar to actively sense their
environment and perform self localization and mapping, more recently
camera-based SLAM and odometry algorithms have witnessed a drastic
boost in performance. Although such passive sensors require a
sufficiently illuminated and textured scene
to infer 3D structure and motion, they comprise several advantages
including higher resolution, higher sensing range and sensing rate as well as lower weight, size and hardware cost. This makes them far more
versatile -- the smaller size and weight, for example, enable
applications such as visual-inertial autonomous navigation of
nanocopters \cite{dunkley14iros}.
As a result, there is a huge demand for real-time capable visual
odometry and visual SLAM algorithms. Among the most desirables
properties are maximal accuracy, robustness (to changes in
scene structure, lighting and faster motion) and density of
the reconstructed environment. In this paper, we propose what we
believe to be the currently most accurate and robust real-time visual
odometry method.

\subsection{Related Work}

The first real-time capable visual SLAM and odometry systems were
pioneered around 2000
\cite{chiuso2002structure,nister2004visual,davison2007monoslam,civera2008inverse}.
The key idea underlying these structure and motion techniques is to
select a set of keypoints (typically corner-like structures), track
them across frames and jointly infer their 3D location and the camera
motion. More recently, we have witnessed a boost of increasingly
performant solutions stemming from numerous advances both in computing
hardware and in algorithmic sophistication. For example, Klein and
Murray~\cite{klein2007parallel} parallelized the motion and 3D structure
estimations. Strasdat et al.~proposed to
expand the concept of keyframes to integrate
scale~\cite{strasdat2010scale} and proposed a double window
optimization~\cite{strasdat2011double}. 
Mei et al.~\cite{mei2011rslam} developed a relative SLAM approach for
constant-time estimation.
More recently,
ORB-SLAM~\cite{mur2015orb, mur2016orb} introduced an efficient visual
SLAM solution based on ORB features and innovations such as map reuse.
It has gained a lot of popularity due to its high tracking accuracy
and robustness and is among the state-of-the-art methods for visual
SLAM. 

While the traditional structure and motion/visual SLAM algorithms were
based on heuristically selected keypoints, more recently a number of
so-called direct methods were proposed
\cite{stuhmer2010real,newcombe2011dtam,engel2013semi,forster2014svo,engel2014lsd}.
These aim at computing geometry and motion directly from the 
images thereby skipping the intermediate keypoint selection step.
Algorithmically they typically rely on robust cost functions and
optimization through Gauss-Newton iteration as done also for RGB-D
based SLAM in \cite{kerl2013dense, kerl2013robust, klose2013efficient}.
While the methods
\cite{stuhmer2010real,newcombe2011dtam} rely on variational methods
and total variation regularization to generate dense reconstructions
in real-time (on powerful GPUs), the latter
works \cite{engel2013semi,forster2014svo} abstain from dense solutions in order to not oversmooth or
hallucinate geometric structures but rather generate semi-dense or sparse
reconstructions without the need for GPU support. The extension of
direct monocular SLAM algorithms to large-scale environments was
proposed in LSD-SLAM \cite{engel2014lsd,engel2015large}. The key idea
is to incrementally track the camera and simultaneously perform a
pose graph optimization in order to keep the entire camera trajectory
globally consistent. While this kind of stripped-down version of
bundle adjustment (which includes motion, but excludes structure) does not remove the drift, it appears to spread
it out across the computed trajectory. Since then, the semi-dense
direct VO/SLAM has been further extended by support for omnidirectional
cameras~\cite{caruso2015large} and tightly coupling with IMU~\cite{usenko2016direct}. A rolling shutter calibration
method for LSD-SLAM has been recently proposed
in~\cite{kim2016direct}.

While we are witnessing an ongoing competition between
keypoint based algorithms and direct algorithms, recently Direct
Sparse Odometry (DSO) \cite{engel2016direct} was shown to outperform
the state-of-the-art keypoint based monocular SLAM algorithm ORB-SLAM
\cite{mur2015orb} in terms of both accuracy and robustness on a fairly
large dataset for monocular camera tracking
\cite{engel2016monodataset}.

While this seems to indicate a certain advantage of direct methods,
DSO has several shortcomings as a technique for visual
odometry or SLAM: Firstly, the mentioned performance gain was
demonstrated on a photometrically calibrated dataset. In the absense
of this photometric calibration (many datasets do not provide it), the
performance of direct methods like DSO substantially degrades.
Secondly, being a pure monocular system, DSO invariably cannot
estimate the scale of the reconstructed scene or the units of camera
motion. Furthermore, the estimated trajectory suffers from  substantial
scale drift such that even manually providing the best scale
does not resolve the problem -- see Fig~\ref{fig:qualitative_kitti}.
Thirdly, as shown in~\cite{engel2016direct} DSO is quite sensitive to
geometric distortions as those induced by fast motion and rolling
shutter cameras. While techniques for calibrating rolling shutter exist for direct
SLAM algorithms~\cite{kim2016direct},
these are often quite involved and far from real-time capable.

\subsection{Contribution}

In this work, we propose {\em Stereo DSO} as a novel direct visual
odometry method for highly accurate and robust motion
and 3D structure estimation in real-time from a moving stereo camera. 
It addresses the aforementioned shortcomings of previous approaches by
leveraging the additional sensor information. Thus it provides an
accurate and (due to the stereo initialization) much faster converging scale
estimation and is less sensitive to missing photometric calibration or
the effects of rolling shutter. In particular:
\begin{itemize}
\item We derive a stereo version of DSO. To this end, we detail the
  proposed combination of temporal multi-view stereo and static stereo
  and their integration with marginalization using the Schur complement.
  Unlike previous extension of monocular direct approach~\cite{engel2014lsd}
  to stereo~\cite{engel2015large} (both apply filtering approaches to the geometry that do not involve bundle adjustment), we propose a novel way to extend the energy function and the entire bundle adjustment procedure in a manner that real-time capability is assured.
\item  We perform systematic quantitative evaluations on the KITTI dataset
  and on the Cityscapes dataset. Comparisons to alternative
  methods like Stereo ORB-SLAM and Stereo LSD-SLAM demonstrate that the
  proposed Stereo DSO is superior to these techniques, in particular
  when evaluated on the KITTI testing set, indicating that the method generalizes better to unknown settings.
\item A quantitative evaluation over longer ranges
  demonstrates that the proposed Stereo DSO algorithm without loop
  closure optimization outperforms Stereo ORB-SLAM which applies global pose graph optimization and bundle adjustment.
\end{itemize}
\section{Direct Sparse VO with Stereo Cameras}

\begin{figure}[t]
\begin{center}
\includegraphics[width=1.0\linewidth]{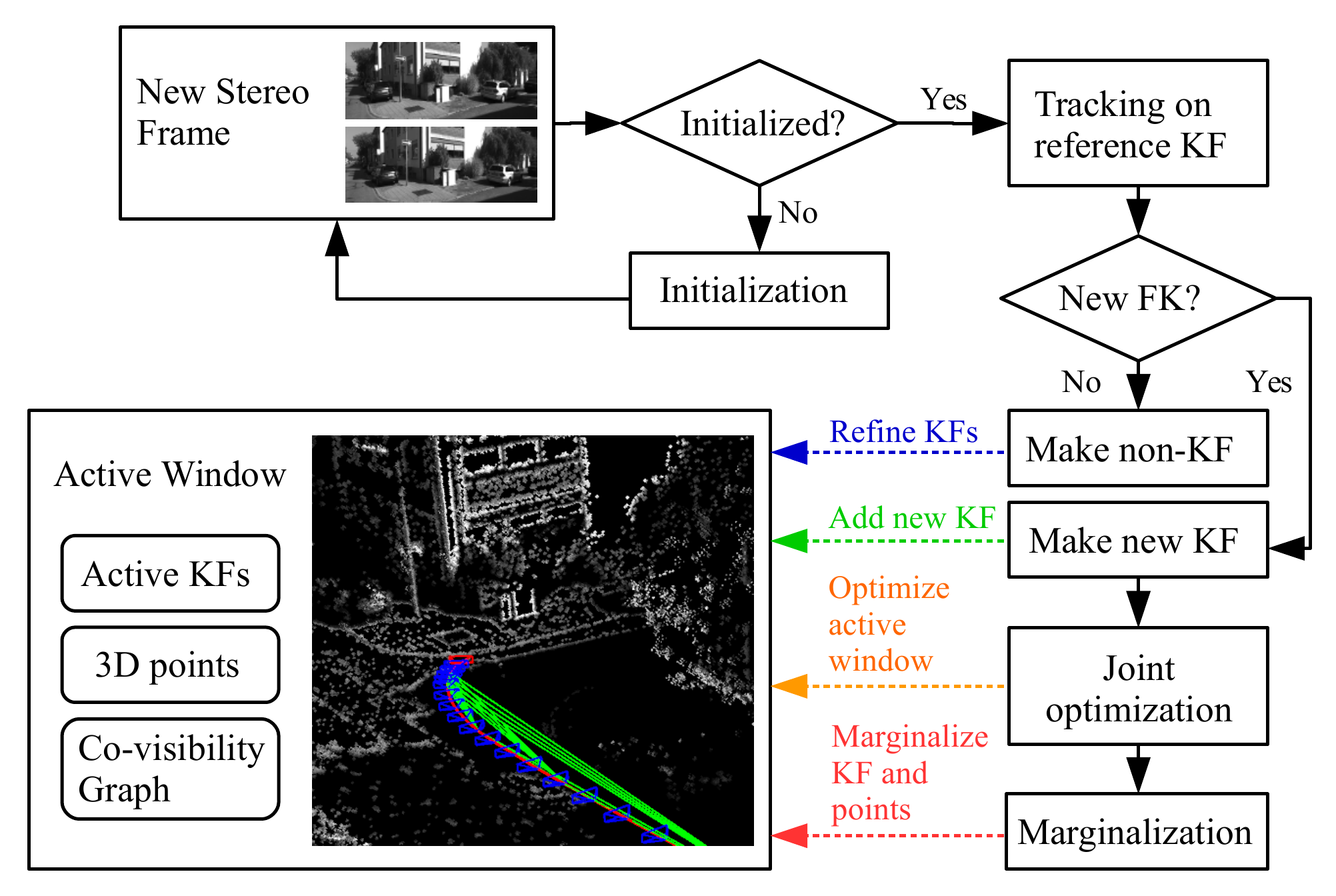}
\end{center}
\vspace{-1.0em}
   \caption{System overview.}\label{fig:system_overview}
   \vspace{-1.0em}
\end{figure}

% System overview
Our Stereo DSO is a system that combines static stereo with multi-view stereo.
As is demonstrated in~\cite{engel2015large}, 
such hybrid approach brings several advantages over each of the separate one:
\begin{itemize}
\setlength\itemsep{0em}
\item Absolute scale can be directly calculated from static stereo from the known baseline of the stereo camera.
\item Static stereo can provide initial depth estimation for multi-view stereo.
\item Due to the fixed baseline, static stereo can only accurately triangulate 3D points within a limited depth range. This limit is resolved by temporal multi-view stereo.
\item They can complement each other in degenerate cases where edges are parallel to epipolar lines.
\end{itemize}
An overview of our system is shown in Fig~\ref{fig:system_overview}. 
Instead of using random depth for initialization~\cite{engel2013semi, engel2014lsd, 
engel2016direct}, our system uses depth estimation from static stereo matching 
(Sec~\ref{sec:tracking}). Based on the direct image alignment formulation 
(Sec~\ref{sec:direect_alignment}), new stereo frames are first tracked with respect to 
their reference keyframe in a coarse-to-fine manner (Sec~\ref{sec:tracking}). 
The obtained pose estimate is used to refine the depth of recently selected points.
Then our system checks whether a new keyframe is needed by the current active window. 
If not, a non-keyframe will be created, otherwise a new keyframe will be generated and added to the active window
(Sec~\ref{sec:frame_management}). For all keyframes in the active window, a joint 
optimization of their poses, affine brightness parameters, as well as the depths of all the observed 3D 
points and camera intrinsics is performed. To maintain the size of the active window, 
old keyframes and 3D points are marginalized out using the Schur complement (Sec~\ref{sec:windowed_optimization}).

% =====================================================================================
\subsection{Notation}
% =====================================================================================
Throughout this paper we use light, bold lower-case letters and bold upper-case letters 
to denote scalars ($u$), vectors ($\mathbf{t}$) and matrices ($\mathbf{R}$) respectively. 
Light upper-case letters are used to represent functions ($I$).

Camera calibration matrices are denoted by $\mathbf{K}$. 
Camera poses are represented by matrices of the special Euclidean group $\mathbf{T}_{i} \in SE(3)$, 
which transform a 3D coordinate from the camera coordinate 
system to the world coordinate system. $\mathbf{\Pi_{K}}$ and $\mathbf{\Pi^{-1}_{K}}$ 
are used to denote camera projection and back-projection functions. In this paper, a 3D 
point is represented by its image coordinate $\mathbf{p}$ and inverse depth $d_\mathbf{p}$ relative to its host keyframe.
The host keyframe is the frame the point got selected from.
The inverse depth parameterization has been demonstrated to be advantageous when errors in images 
are modeled as Gaussian distributions~\cite{davison2007monoslam, civera2008inverse}. 

% =====================================================================================
\subsection{Direct Image Alignment Formulation}\label{sec:direect_alignment}
% =====================================================================================
% Affine illumination correction in this subsection
Suppose a point set $\mathcal{P}_{i}$ in a reference frame $I_i$ is observed in another frame $I_j$, 
the basic idea of direct image alignment can be formulated as
\begin{equation}\label{eq:direct_alignment_basic}
E_{ij}=\sum_{\mathbf{p}\in{\mathcal{P}_{i}}} \omega_{\mathbf{p}} \Big\| I_j[\mathbf{p'}] - I_{i}[\mathbf{p}] \Big\|_{\gamma},
\end{equation}
where $\|\cdot\|_{\gamma}$ is the Huber norm and $\omega_{\mathbf{p}}$ is a weighting that down-weights high image gradients
\begin{equation}
\omega_{\mathbf{p}}=\frac{c^{2}}{c^{2}+\|\nabla I_{i}(\mathbf{p})\|^{2}_{2}},
\end{equation}
with some constant $c$. $\mathbf{p'}$ is the projection of $\mathbf{p}$ in $I_j$ calculated by
\begin{equation}\label{eq:projection}
\mathbf{p'}=\mathbf{\Pi_{K}}\big(\mathbf{T}_{ji}\mathbf{\Pi^{-1}_{K}}(\mathbf{p}, d_{\mathbf{p}})\big),
\end{equation}
with $d_{\mathbf{p}}$ the inverse depth of $\mathbf{p}$ and $\mathbf{T}_{ji}$ the transformation 
that transforms a point from frame $i$ to frame $j$:
\begin{equation}
\mathbf{T}_{ji}:=\begin{bmatrix}\mathbf{R}_{ji} & \mathbf{t}\\0 & 1\\\end{bmatrix}=\mathbf{T}^{-1}_{j}\mathbf{T}_{i}.
\end{equation}

Conventional direct methods, both dense and semi-dense, tend to use as many pixels from 
each image as possible. While bringing heavy computational burden to the system, its 
benefit saturates fast. Therefore, in~\cite{engel2016direct} the authors proposed a 
strategy to select a fixed number of points from each frame, uniformly across all the 
regions with sufficient gradient. For each selected point, a small neighborhood around 
it is used to calculate the photometric error in (\ref{eq:direct_alignment_basic}). In 
this paper we follow the same approach, but use the stereo image pair to verify the 
selected points and assist the depth initialization. More details are provided in 
Sec~\ref{sec:frame_management}.

As the photometric error is calculated directly on pixel intensities, it is very 
sensitive to sudden illumination changes between consecutive frames. Ideally the exposure
time of each frame, as well as the camera response function (which can be highly non-linear) 
are directly accessible from the hardware~\cite{engel2016monodataset}, which can be used 
to correct such effect. When this information is not available (as for most existing datasets),
similar to~\cite{engel2016direct}, 
we introduce two parameters $a_{i}$, $b_{i}$ for each image to model an affine brightness 
change. The energy function in (\ref{eq:direct_alignment_basic}) is then modified 
to 
\begin{equation}\label{eq:direct_alignment_energy}
E_{ij}=\sum_{\mathbf{p}\in{\mathcal{P}_{i}}} 
       \sum_{\tilde{\mathbf{p}}\in{\mathcal{N}_{\mathbf{P}}}} 
       \omega_{\tilde{\mathbf{p}}} \bigg\| I_j[\tilde{\mathbf{p}}']-b_{j} - 
                                           \frac{e^{a_{j}}}{e^{a_{i}}}(I_{i}[\tilde{\mathbf{p}}]-b_{i}) \bigg\|_{\gamma},
\end{equation}
with $\mathcal{N}_{\mathbf{P}}$ the 8-point pattern of $\mathbf{p}$ as defined in~\cite{engel2016direct} and $\tilde{\mathbf{p}}'$ the projection of the pattern point $\tilde{\mathbf{p}}$ into $I_j$.
$a_{i}$, $a_{j}$, $b_{i}$, $b_{j}$ are estimated in the windowed optimization as is shown in Sec~\ref{sec:windowed_optimization}.

% =====================================================================================
\subsection{Tracking}\label{sec:tracking}
% =====================================================================================
Each time a new stereo frame is fed into the system, direct image alignement~\cite{engel2013semi} is used to track it with respect to the newest keyframe in 
the active window: a constant motion model is used to assign an initial 
pose to the new frame to be tracked. All the points inside the active window are projected into the new frame. Then the pose of the new frame is optimized by minimizing the energy function (\ref{eq:direct_alignment_energy}) while keeping the depth values fixed. The optimization is performed with Gauss-Newton on an image pyramid in a coarse-to-fine order.

To initialize the whole system, i.e., to track the second frame with respect to the initial one using 
(\ref{eq:direct_alignment_energy}), the inverse depth values of the points in the first 
frame are required. Previous monocular direct VO approaches use random depth values for 
initialization~\cite{engel2013semi,engel2014lsd,engel2016direct}, thus usually need a 
certain pattern of the initial camera movement. In this work we use static stereo 
matching to estimate a semi-dense depth map for the first frame. As at this stage the affine
brightness transfer factors between the stereo image pair are unknown, correspondences are 
searched along the horizontal epipolar line using the NCC of the 
$3\times5$ neighborhood.

% Say something about tracking fail?

% =====================================================================================
\subsection{Frame Management}\label{sec:frame_management}
% =====================================================================================
If a new stereo frame is successfully tracked, we use the same criteria as in
~\cite{engel2016direct} to determine if a new keyframe is required. The basic idea is 
to check if the scene or the illumination has sufficiently changed. Scene changing is 
evaluated by the mean squared optical flow, as well as the mean squared optical flow
without rotation between the current frame and the last 
keyframe in the active window. Illumination change is quantized by the relative 
brightness factor $|a_{j}-a_{i}|$.

To create a new keyframe, a sparse set of points is selected from the image, which 
will be called candidate points in the rest of the paper. To select points evenly distributed across the image and only points that have sufficient image gradient, the
image is divided into small blocks and for each block an adaptive threshold is calculated.
Instead of using square blocks of fixed size~\cite{engel2016direct}, we use blocks with size that is proportional to the image size. We find this helpful for images
with dissimilar width and height like the ones from KITTI.
A point is selected if it surpasses the threshold of the block and it has the largest
absolute gradient in its neighborhood.

Before a candidate point is activated and optimized in the windowed optimization, its 
inverse depth is constantly refined by the following non-keyframes. In the monocular 
case, the candidate point is usually initialized to have a depth range from 0 to infinity,
corresponding to a large depth variance. In our case, we use static stereo matching with NCC to obtain a better depth initialization
for the candidate points, which significantly increases the tracking accuracy.

\begin{figure}
\begin{center}
\includegraphics[width=1.0\linewidth]{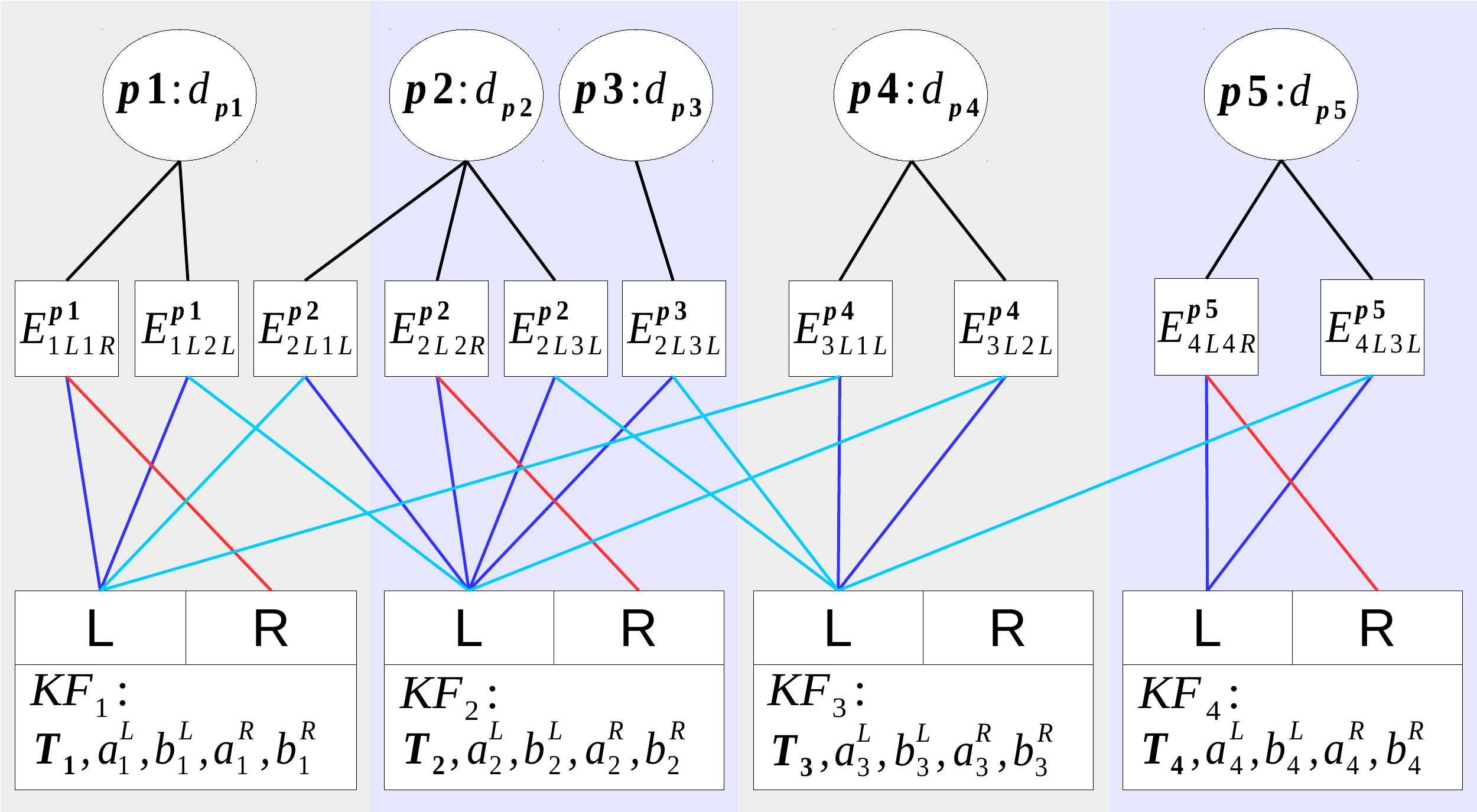}
\end{center}
\vspace{-0.5em}
   \caption{Factor graph of the energy function. In this example, 5 points are observed
            by 4 keyframes. Each energy factor is related to one point and two keyframes,
            thus depends on the inverse depth of the point, the camera poses of the two keyframes 
            and their affine brightness correction factors, as well as the camera intrinsic
            parameters (we assume the same for the left and the right cameras, and omit it
            here for simplicity). Constraints from host keyframes and static stereo are
            shown in dark blue and red respectively.
            Remaining constraints in light blue are the ones from the keyframes the points
            are observed.}\label{fig:factor_graph}
   \vspace{-1.0em}
\end{figure}

When old points are removed from the active window by marginalization
(Sec~\ref{sec:windowed_optimization}), candidate points are activated and added to the 
joint optimization. Each activated point is hosted in one keyframe and is observed by 
several other keyframes in the active window. Each time an active point is observed
in another keyframe, it creates a photometric energy factor defined as the inner part of
(\ref{eq:direct_alignment_energy}):
\begin{equation}\label{eq:point_residual}
E^{\mathbf{p}}_{ij}=\omega_{\mathbf{p}} \bigg\| I_j[\mathbf{p}']-b_{j} - 
                                                \frac{e^{a_{j}}}{e^{a_{i}}}(I_{i}[\mathbf{p}]-b_{i}) \bigg\|_{\gamma}.
\end{equation}
For simplicity of notation, we omit the summation over the neighborhood $\mathcal{N}_{\mathbf{P}}$
in the above equation as well as in the rest of this paper. 
A factor graph of the energy function is shown in Fig~\ref{fig:factor_graph}, where each 
factor (represented as small square in the middle) depends on the inverse depth of the
point, the cameras poses of the host keyframe and the 
keyframe that observes this point, as well as their affine brightness correction factors. The constraints from static stereo (denoted with red lines) introduce scale information into the system. Moreover, they also provide good geometric priors to temporal multi-view stereo. 

% Energy formulation (factor graph)

% =====================================================================================
\subsection{Windowed Optimization}\label{sec:windowed_optimization}
% =====================================================================================
Putting all the energy factors together, the final energy function to be minimized in the 
windowed optimization is
\begin{equation}\label{eq:final_energy}
E=\sum_{i\in{\mathcal{F}}} 
  \sum_{\mathbf{p}\in{\mathcal{P}_{i}}}
  \sum_{j\in{obs(\mathbf{p})}}  
  E^{\mathbf{p}}_{ij},
\end{equation}
where $\mathcal{F}$ is the set of the keyframes in the current window and $obs(\mathbf{p})$ is the set of the keyframes in $\mathcal{F}$ that can observe $\mathbf{p}$. The energy is optimized iteratively using Gauss-Newton algorithm:
\begin{equation}
\delta\boldsymbol{\xi} = -({\mathbf{J}^{T}\mathbf{WJ}})^{-1}\mathbf{J^{T}Wr},
\end{equation}
\begin{equation}
\boldsymbol{\xi}^{new} = \delta\boldsymbol{\xi} \boxplus \boldsymbol{\xi},
\end{equation}
where $\mathbf{r}$ contains the stacked residuals, $\mathbf{J}$ is the Jacobian and $\mathbf{W}$
is the diagonal weight matrix. The parameters we want to optimize are enclosed in
\begin{equation}
\begin{split}
\boldsymbol{\xi} = ( & \mathbf{T}_{0,...,N_{f}-1}, d_{0,...,N_{p}-1}, \mathbf{c}, \\
                      & a^{L}_{0,...,N_{f}-1}, b^{L}_{0,...,N_{f}-1}, \\
                      & a^{R}_{0,...,N_{f}-1}, b^{R}_{0,...,N_{f}-1}),
\end{split}
\end{equation}
where $\mathbf{c}$ is the vector containing the global camera intrinsics, $L$ and $R$ denote the parameter of the left and right camera frame, and $N_{f}$ and $N_{p}$ are the numbers of keyframes and active points in the current window, respectively.
The $\boxplus$-operator on $\mathbf{T}$ is
defined using Lie Algebra $se(3)$ as
\begin{equation}
\boxplus : se(3) \times SE(3) \rightarrow SE(3), \mathbf{x} \boxplus \mathbf{T} := exp(\hat{\mathbf{x}})\mathbf{T},
\end{equation}
whereas on the rest parameters it is simply the conventional addition.

\noindent{\textbf{Temporal Multi-View Stereo.}} Each residual from temporal multi-view stereo 
is defined as
\begin{equation}\label{eq:residual}
r^{t}_k =  I_j[\mathbf{p}'(\mathbf{T}_{i}, \mathbf{T}_{j}, d, \mathbf{c})]-b_{j} - 
                  \frac{e^{a_{j}}}{e^{a_{i}}}(I_{i}[\mathbf{p}]-b_{i}).
\end{equation}
Similar to~\cite{engel2016direct} the Jacobian is defined as
\begin{equation}\label{eq:Jacobian}
\mathbf{J}^{t}_{k} = \bigg[ \frac{\partial I_j}{\partial \mathbf{p}'} \frac{\partial \mathbf{p}'(\delta\boldsymbol{\xi} \boxplus \boldsymbol{\xi})}{\partial \delta \boldsymbol{\xi}^{t}_{geo}}, 
                        \frac{\partial r_k(\delta\boldsymbol{\xi} \boxplus \boldsymbol{\xi})}{\partial \delta\boldsymbol{\xi}^{t}_{photo}} 
                 \bigg],
\end{equation}
where the geometric parameters $\boldsymbol{\xi}^{t}_{geo}$ are $(\mathbf{T}_{i}, \mathbf{T}_{j}, d, \mathbf{c})$
and the photometric parameters $\boldsymbol{\xi}^{t}_{photo}\!=\!(a_i, a_j, b_i, b_j)$.\\

\noindent{\textbf{Static Stereo.}} For static stereo the residual is modified to 
\begin{equation}\label{eq:residual_static}
r^{s}_k = I^R_i[\mathbf{p}'(\mathbf{T}_{ji}, d, \mathbf{c})]-b^R_{i} - 
                  \frac{e^{a^R_{i}}}{e^{a^L_{i}}}(I_{i}[\mathbf{p}]-b^L_{i}).
\end{equation}
The Jacobian has the same form as in (\ref{eq:Jacobian}) but now with less geometric 
parameters $\boldsymbol{\xi}_{geo}\!=\!(d, \mathbf{c})$, because the relative transformation between
the left and right cameras $\mathbf{T}_{ji}$ is fixed. Therefore, $\mathbf{T}_{ji}$ is not optimized in the
windowed optimization. \\

\noindent{\textbf{Stereo Coupling.}} To balance the relative weights of temporal
multi-view and static stereo, we introduce a coupling factor $\lambda$ to weight
the constraints from static stereo differently. The energy function in (\ref{eq:final_energy})
thus can be further formulated as
\begin{equation}
E = \sum_{i\in{\mathcal{F}}} 
    \sum_{\mathbf{p}\in{\mathcal{P}_{i}}}
    \Big(\sum_{j\in{obs^{t}(\mathbf{p})}} E^{\mathbf{p}}_{ij}
    +\lambda E^{\mathbf{p}}_{is}\Big),
\end{equation}
where $obs^{t}(\mathbf{p})$ are the observations of $\mathbf{p}$ from
temporal multi-view stereo, and $E^{\mathbf{p}}_{is}$ the energy belonging to the static stereo residuals. The effects of the coupling factor are detailed in Sec~\ref{sec:evaluation_kitti}. \\

\noindent{\textbf{Marginalization.}} To keep the active window of bounded size, old keyframes are removed by 
marginalization using the Schur complement~\cite{leutenegger2015keyframe, engel2016direct}. 
Before marginalizing a keyframe, we first marginalize all active points that are not observed 
by the two latest keyframes together with all active points hosted in the keyframe. 
Afterwards, the keyframe is marginalized and moved out of the active window. Let $\mathbf{H}=\mathbf{J}^{T}\mathbf{WJ}$ and
$\mathbf{b}=\mathbf{J}^{T}\mathbf{W}\mathbf{r}$ be the
Gauss-Newton system containing only the variables to marginalize and the variables connected to them in the factor graph. % , where $\mathbf{r}$ is the stacked residual vector, $\mathbf{J}$ its Jacobian and $\mathbf{W}$ the diagonal matrix consisting of the weights.
If we use $\alpha$ and $\beta$ to respectively denote the variables to keep and to
marginalize, the Gauss-Newton system can be rearranged to
\begin{equation}
\begin{bmatrix} \mathbf{H}_{\alpha\alpha} & \mathbf{H}_{\alpha\beta} \\
                \mathbf{H}_{\beta\alpha} & \mathbf{H}_{\beta\beta} \end{bmatrix}
\begin{bmatrix} \mathbf{x}_{\alpha} \\ \mathbf{x}_{\beta} \end{bmatrix} = 
\begin{bmatrix} \mathbf{b}_{\alpha} \\ \mathbf{b}_{\beta} \end{bmatrix}.                
\end{equation}
Multiplying the second line by $\mathbf{H}_{\alpha\beta}\mathbf{H}^{-1}_{\beta\beta}$ and
subtracting it from the first leads to
\begin{equation}
\underbrace{(\mathbf{H}_{\alpha\alpha}-\mathbf{H}_{\alpha\beta}\mathbf{H}^{-1}_{\beta\beta}\mathbf{H}^{T}_{\alpha\beta})}_{\widehat{\mathbf{H}}_{\alpha\alpha}}\mathbf{x}_{\alpha} = 
\underbrace{\mathbf{b}_{\alpha}-\mathbf{H}_{\alpha\beta}\mathbf{H}^{-1}_{\beta\beta}\mathbf{b}_{\beta}}_{\widehat{\mathbf{b}}_{\alpha}}.
\end{equation}
The resulting system $\widehat{\mathbf{H}}_{\alpha\alpha}\mathbf{x}_{\alpha}=\widehat{\mathbf{b}}_{\alpha}$ only depends on the variables to keep $\mathbf{x}_{\alpha}$ and is added as prior to the subsequent optimizations.

\section{Evaluation}\label{sec:evaluation}
We evaluate our method on two popular datasets: the KITTI Odometry Benchmark and the
Cityscapes Dataset. Both datasets provide synchronized stereo sequences with rectified
high resolution images. We compare our method thoroughly with state-of-the-art stereo VO methods, both
feature-based and direct method on KITTI.
On Cityscapes, we create several sequences with ground truth camera poses calculated from the provided GPS
coordinates. We show both the tracking and 3D reconstruction results on this dataset and demonstrate that our method can be used for large-scale camera tracking and 3D reconstruction.

% =====================================================================================
\subsection{KITTI Visual Odometry Benchmark}\label{sec:evaluation_kitti}
% =====================================================================================
\begin{figure}
\begin{center}
\includegraphics[width=0.96\linewidth]{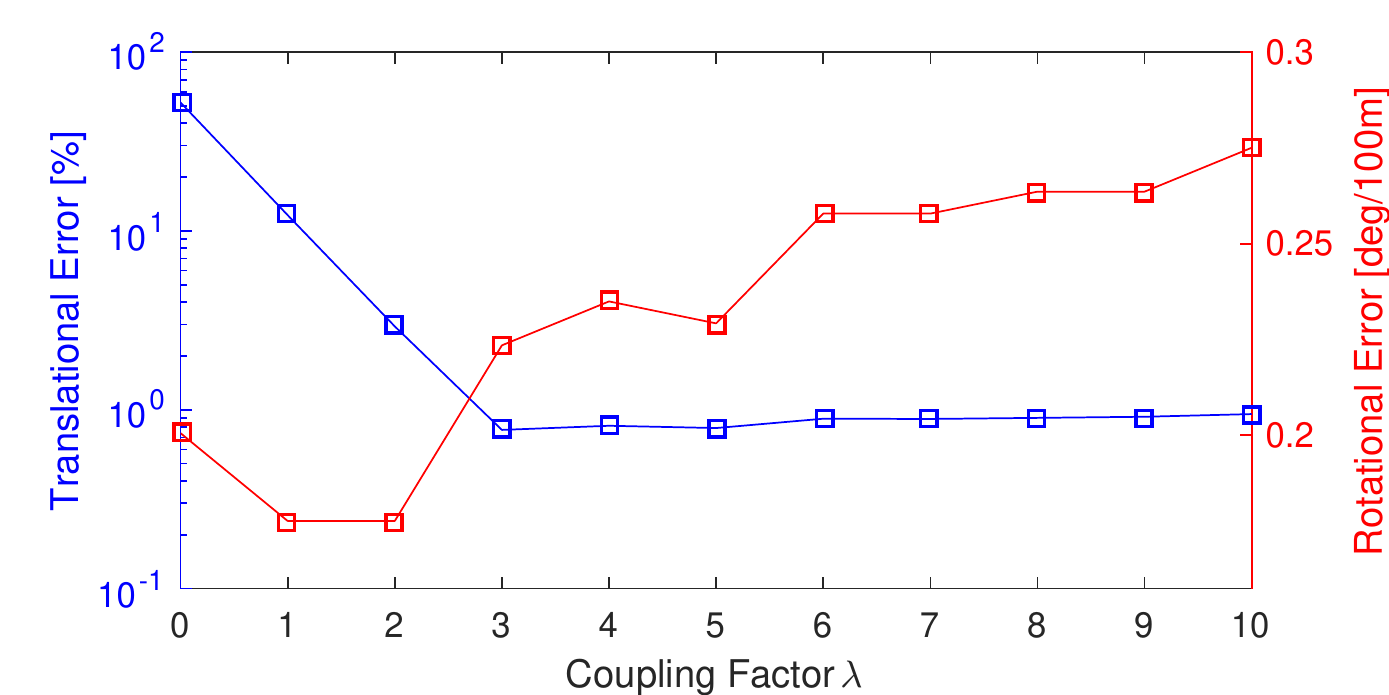}
\end{center}
\vspace{-0.7em}
   \caption{Average translational and rotational errors on Seq. 06 for different coupling factors. 
   }\label{plot:coupling_06}
   \vspace{-1.2em}
\end{figure}
We evaluate our method on the KITTI Odometry Benchmark~\cite{geiger2013vision}, where altogether 22 driving 
sequences are provided. The first 11 (00-10) sequences are provided with ground truth 6D 
poses as training set, whereas the latter 11 sequences comprise the testing set. 

We first test the influence of the stereo coupling factor on an example sequence (Seq. 06).
The translational and rotational errors obtained by using different coupling factors are
shown in Fig~\ref{plot:coupling_06}: Introducing constraints from 
static stereo with certain weightings ($\lambda=1,2$) significantly reduces both translational and rotational errors. Further increasing the weighting ($\lambda>3$) makes the method more sensitive to incorrect matchings from static stereo and thus degrade the performance. The estimated trajectories
for $\lambda = 0-3$ are shown in the supplementary material. 

The comparison of the VO accuracy of different stereo methods on the training set are shown in Table~\ref{tab:kitti_training}.
We compare our method to Stereo LSD-SLAM and ORB-SLAM2, which are currently the state-of-the-art direct and 
feature-based stereo VO methods respectively.
The results for Stereo LSD-SLAM are cited from~\cite{engel2015large} (VO only), while the ones for
ORB-SLAM2 are obtained by running their code with default settings. For fair comparison we turned off its 
loop-closure detection and global bundle adjustment. It can be seen that our results are almost 
always better than LSD-SLAM. 
Compared to ORB-SLAM2, all our rotational errors are better, but 
translational errors are slightly mixed. We claim that this might result from the relatively low frame rate
of the dataset.

\begin{table}
\small
\begin{center}
\begin{tabular}{|c|c|c|c|c|c|c|}
\hline
     & \multicolumn{2}{c|}{St. DSO} & \multicolumn{2}{c|}{ORB-SLAM2} & \multicolumn{2}{c|}{St. LSD-VO} \\
\hline
Seq. & \multicolumn{1}{c}{$t_{rel}$} & \multicolumn{1}{c|}{$r_{rel}$} & \multicolumn{1}{c}{$t_{rel}$} & \multicolumn{1}{c|}{$r_{rel}$} & \multicolumn{1}{c}{$t_{rel}$} & \multicolumn{1}{c|}{$r_{rel}$} \\
\hline
\hline
00   & \multicolumn{1}{c}{0.84}          & \multicolumn{1}{c|}{\textbf{0.26}}          & \multicolumn{1}{c}{\textbf{0.83}} & \multicolumn{1}{c|}{0.29} & \multicolumn{1}{c}{1.09}          & \multicolumn{1}{c|}{0.42}  \\
01   & \multicolumn{1}{c}{1.43}          & \multicolumn{1}{c|}{\textbf{0.09}} & \multicolumn{1}{c}{\textbf{1.38}} & \multicolumn{1}{c|}{0.20}          & \multicolumn{1}{c}{2.13}          & \multicolumn{1}{c|}{0.37}  \\
02   & \multicolumn{1}{c}{\textbf{0.78}}          & \multicolumn{1}{c|}{\textbf{0.21}} & \multicolumn{1}{c}{0.81} & \multicolumn{1}{c|}{0.28}          & \multicolumn{1}{c}{1.09}          & \multicolumn{1}{c|}{0.37}  \\
03   & \multicolumn{1}{c}{0.92}          & \multicolumn{1}{c|}{\textbf{0.16}} & \multicolumn{1}{c}{\textbf{0.71}} & \multicolumn{1}{c|}{0.17}          & \multicolumn{1}{c}{1.16}          & \multicolumn{1}{c|}{0.32}  \\
04   & \multicolumn{1}{c}{0.65}          & \multicolumn{1}{c|}{\textbf{0.15}} & \multicolumn{1}{c}{0.45}          & \multicolumn{1}{c|}{0.18}          & \multicolumn{1}{c}{\textbf{0.42}} & \multicolumn{1}{c|}{0.34}  \\
05   & \multicolumn{1}{c}{0.68}          & \multicolumn{1}{c|}{\textbf{0.19}} & \multicolumn{1}{c}{\textbf{0.64}} & \multicolumn{1}{c|}{0.26}          & \multicolumn{1}{c}{0.90}          & \multicolumn{1}{c|}{0.34}  \\
06   & \multicolumn{1}{c}{\textbf{0.67}} & \multicolumn{1}{c|}{\textbf{0.20}} & \multicolumn{1}{c}{0.82}          & \multicolumn{1}{c|}{0.25}          & \multicolumn{1}{c}{1.28}          & \multicolumn{1}{c|}{0.43}  \\
07   & \multicolumn{1}{c}{0.83}          & \multicolumn{1}{c|}{\textbf{0.36}}          & \multicolumn{1}{c}{\textbf{0.78}} & \multicolumn{1}{c|}{0.42} & \multicolumn{1}{c}{1.25}          & \multicolumn{1}{c|}{0.79}  \\
08   & \multicolumn{1}{c}{\textbf{0.98}} & \multicolumn{1}{c|}{\textbf{0.25}} & \multicolumn{1}{c}{1.07}          & \multicolumn{1}{c|}{0.31}          & \multicolumn{1}{c}{1.24}          & \multicolumn{1}{c|}{0.38}  \\
09   & \multicolumn{1}{c}{0.98}          & \multicolumn{1}{c|}{\textbf{0.18}} & \multicolumn{1}{c}{\textbf{0.82}} & \multicolumn{1}{c|}{0.25}          & \multicolumn{1}{c}{1.22}          & \multicolumn{1}{c|}{0.28}  \\
10   & \multicolumn{1}{c}{\textbf{0.49}} & \multicolumn{1}{c|}{\textbf{0.18}} & \multicolumn{1}{c}{0.58}          & \multicolumn{1}{c|}{0.28}          & \multicolumn{1}{c}{0.75}          & \multicolumn{1}{c|}{0.34}  \\
\hline
\hline
mean & \multicolumn{1}{c}{0.84} & \multicolumn{1}{c|}{0.20} & \multicolumn{1}{c}{0.81} & \multicolumn{1}{c|}{0.26} & \multicolumn{1}{c}{1.14} & \multicolumn{1}{c|}{0.40}  \\
\hline
\end{tabular}
\end{center}
\vspace{-0.7em}
\caption{Comparison of accuracy on KITTI training set. 
         $t_{rel}$ translational RMSE ($\%$), $r_{rel}$ rotational RMSE (degree per 100$m$).
         Both are average over 100$m$ to 800$m$ intervals. Best results are shown as bold numbers.}\label{tab:kitti_training}
         \vspace{-1.2em}
\end{table}

\begin{figure}[h!]
\begin{center}
\includegraphics[width=0.9\linewidth]{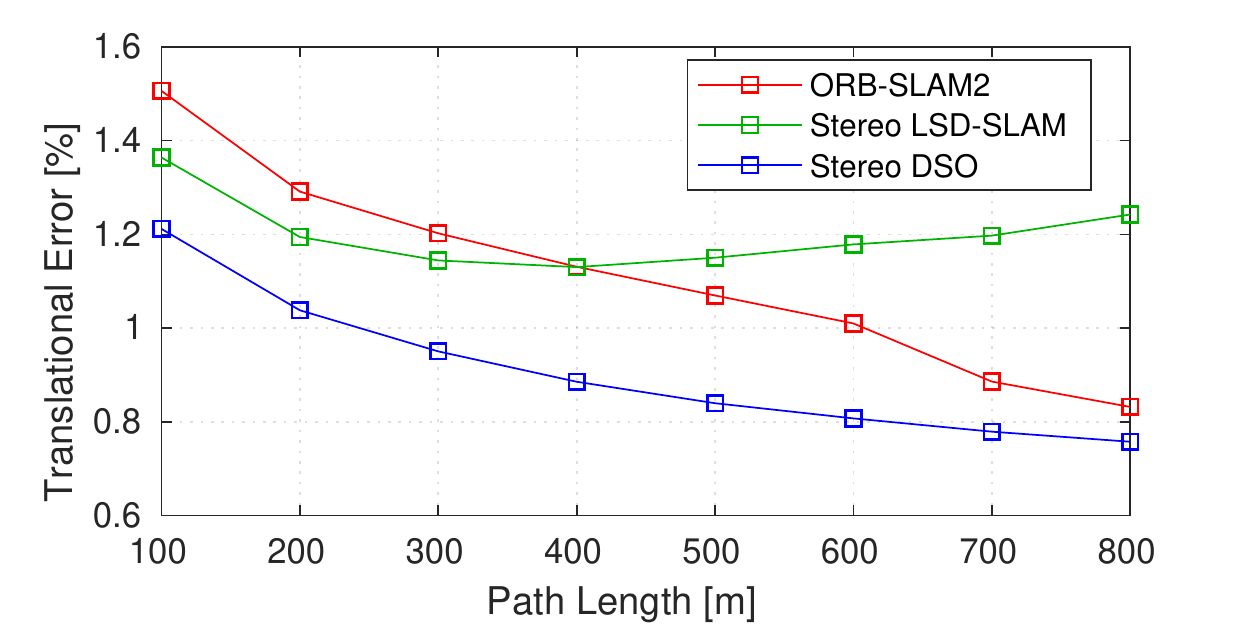}
\includegraphics[width=0.9\linewidth]{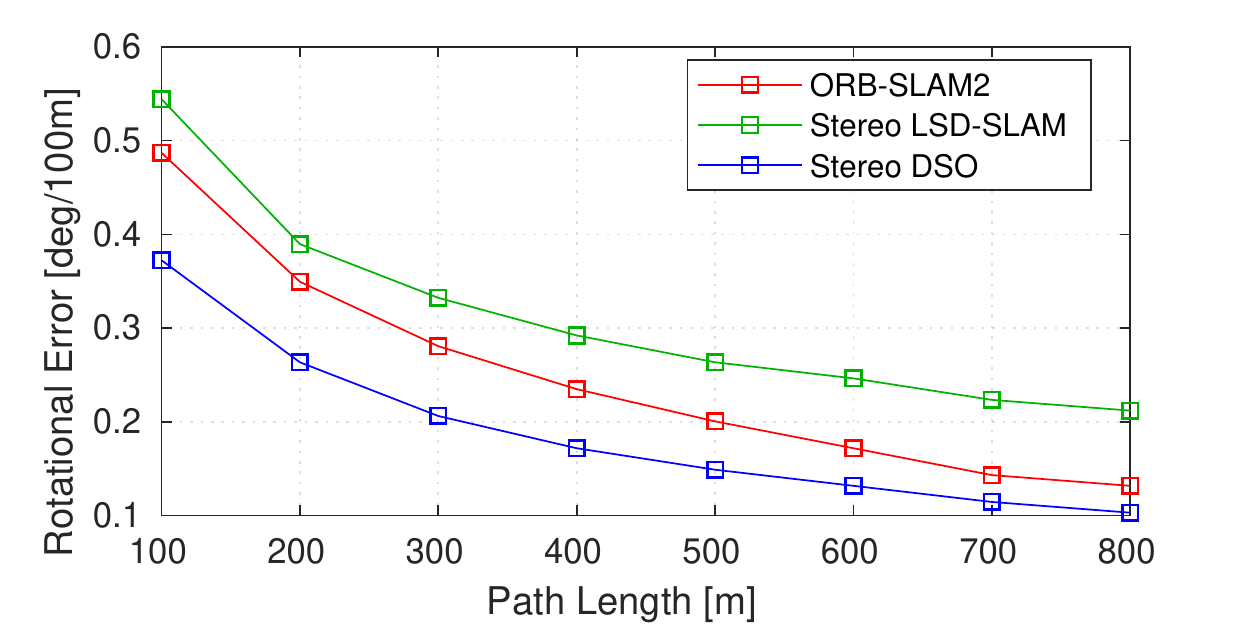}
\includegraphics[width=0.9\linewidth]{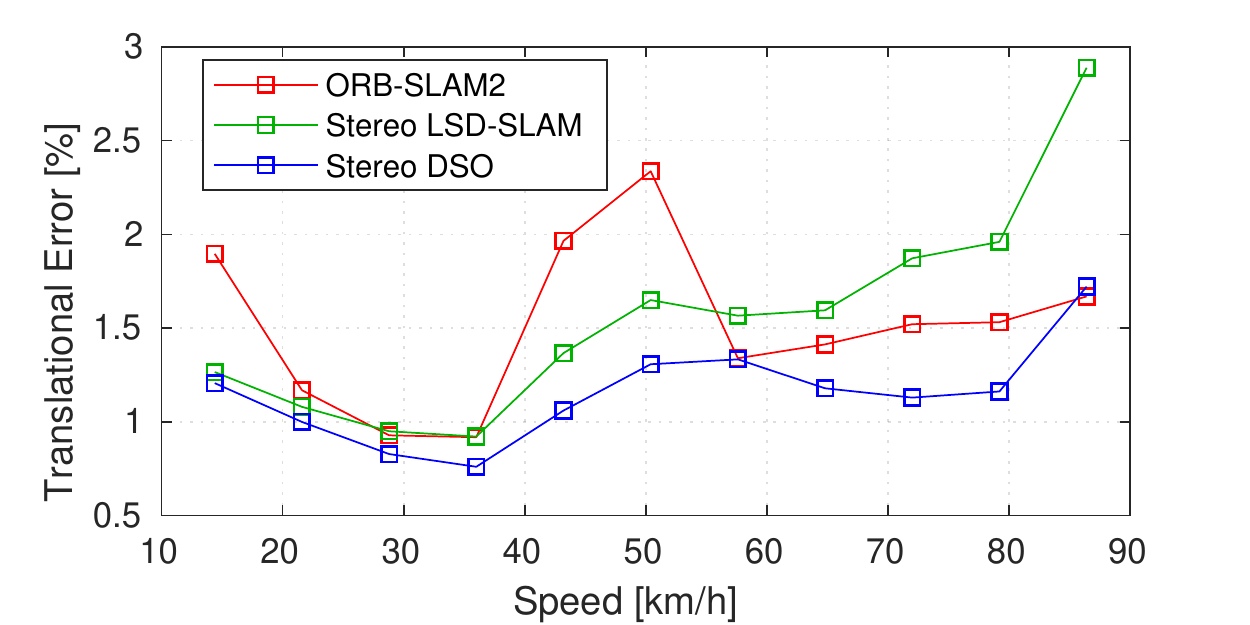}
\includegraphics[width=0.9\linewidth]{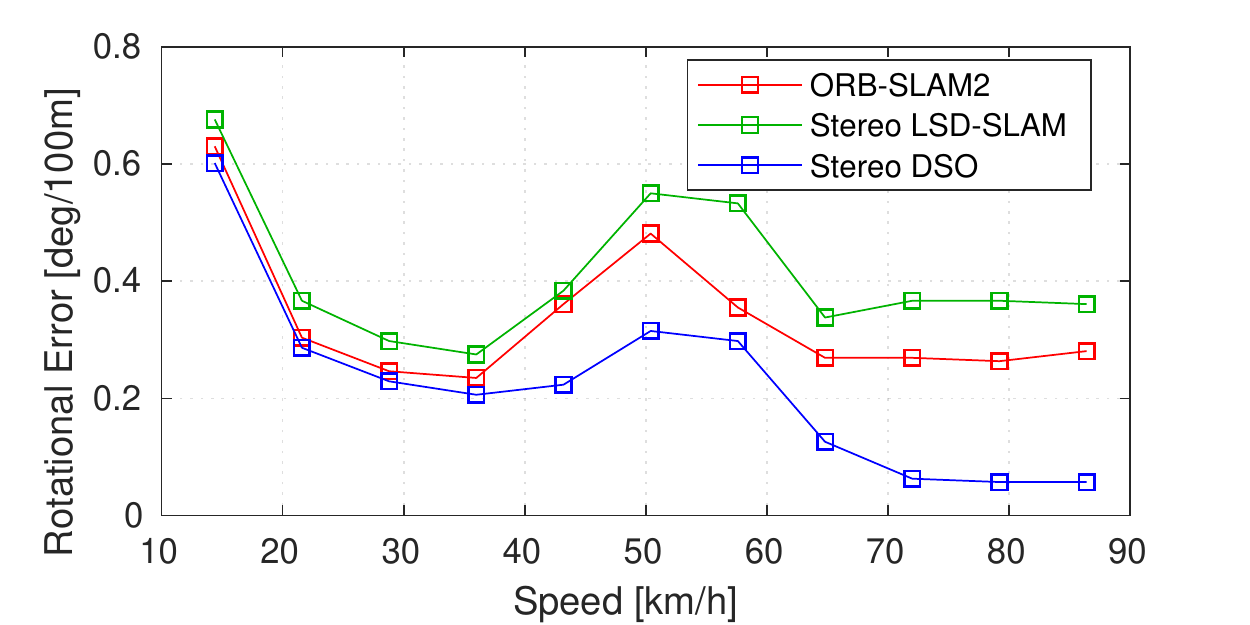}
\end{center}
\vspace{-0.8em}
   \caption{Average translational and rotational errors with respect to driving intervals (top two) 
   and driving speed (bottom two) on KITTI testing set (Seq. 11-21).
   In all the cases, our VO results (without loop closure) are better than the SLAM results (with loop closure, for ORB-SLAM2 also with global bundle adjustment) of LSD-SLAM and ORB-SLAM2. It is surprising to see that even for large intervals our method
   performs better, as in such cases loop closure usually reduces the errors significantly.}\label{plot:kitti_testing}
   \vspace{-1.5em}
\end{figure}

In Fig~\ref{plot:kitti_testing} we show our results on the testing set.
We show the same plots as recommended by the benchmark,
where translational errors and rotational errors with respect to different distance 
intervals and driving speeds over the entire set are plotted. It is worth noting that the results for the 
other two methods are obtained by their SLAM system with loop closure (for ORB-SLAM2 also
with global bundle adjustment), while ours are from pure VO. As can be seen from the plots,
our method performs the best under all settings.

\begin{figure}
\centering
\begin{subfigure}[]{0.98\linewidth}
\includegraphics[width=0.48\linewidth]{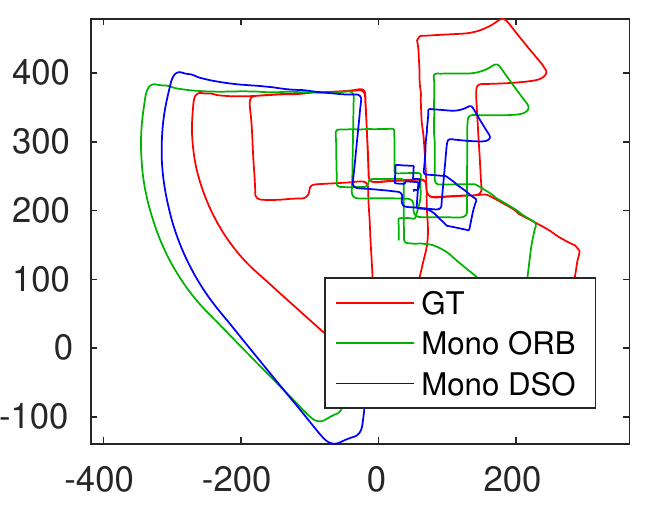}
\includegraphics[width=0.48\linewidth]{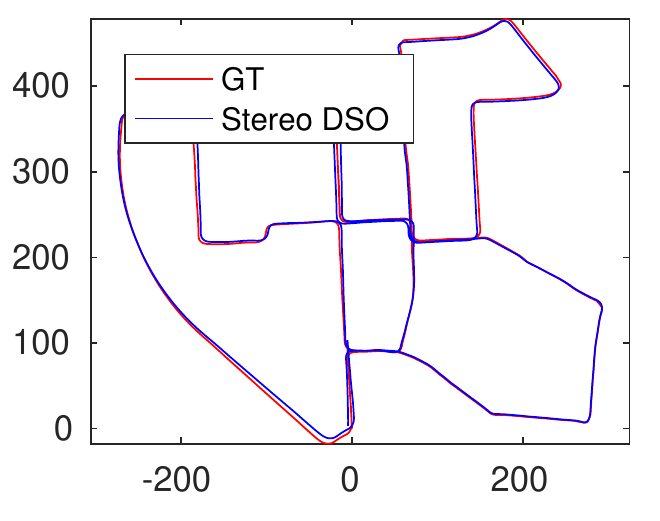}
\end{subfigure} 
\begin{subfigure}[]{0.98\linewidth}
\includegraphics[width=0.48\linewidth]{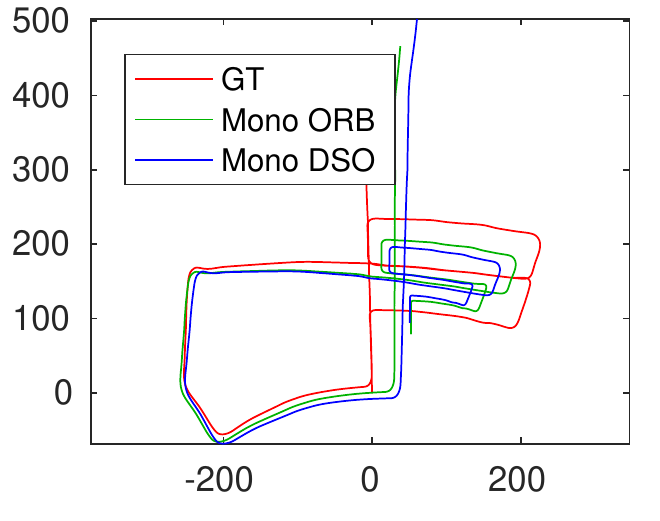}
\includegraphics[width=0.48\linewidth]{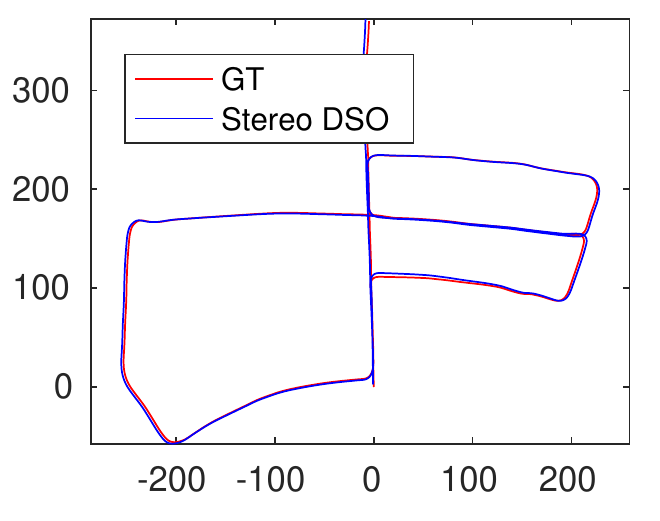}
\end{subfigure} 
\vspace{-0.3em}
   \caption{Qualitative results on KITTI sequence 00 and 05. For the monocular VO results we perform 
   a similarity alignment to the ground truth. }\label{fig:qualitative_kitti}
\end{figure}

\begin{figure}
\centering
\begin{subfigure}[]{0.98\linewidth}
\includegraphics[width=0.97\linewidth]{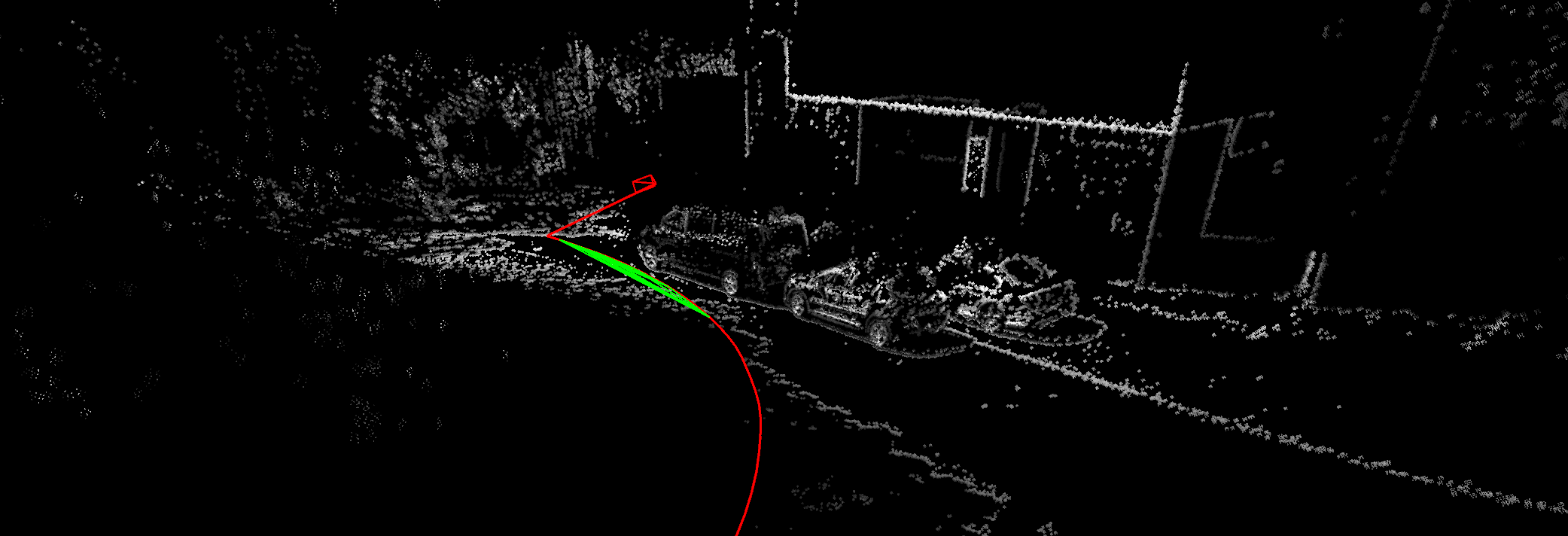}
\end{subfigure}
\begin{subfigure}[]{0.98\linewidth}
\includegraphics[width=0.48\linewidth]{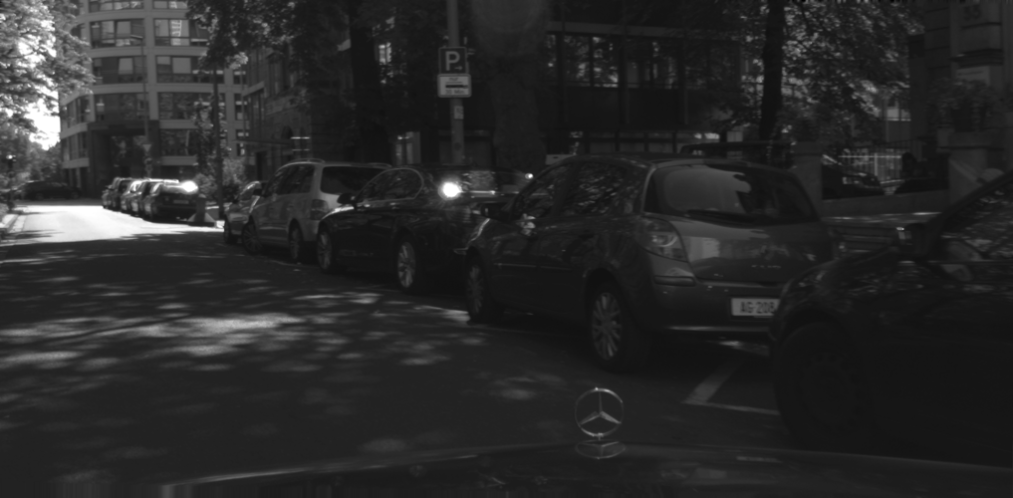}
\includegraphics[width=0.48\linewidth]{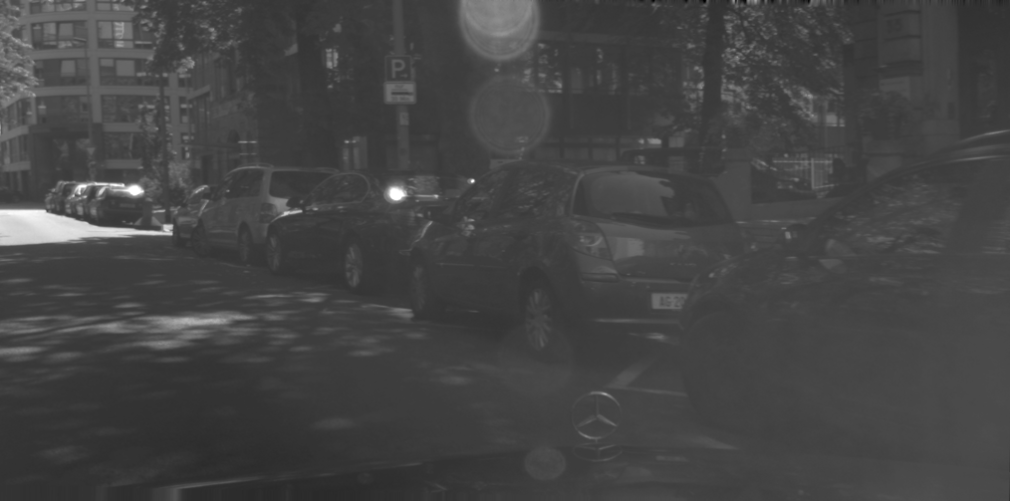}
\end{subfigure}
\begin{subfigure}[]{0.98\linewidth}
\includegraphics[width=0.48\linewidth]{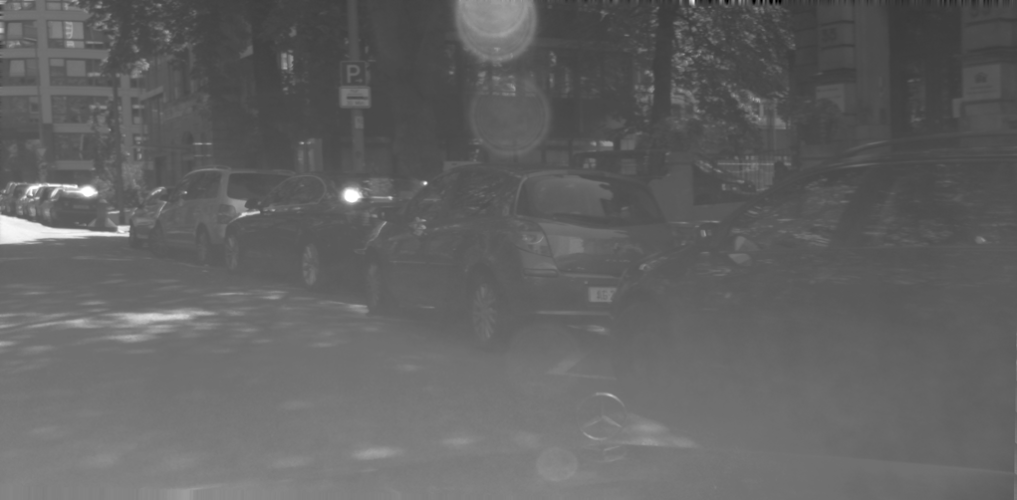}
\includegraphics[width=0.48\linewidth]{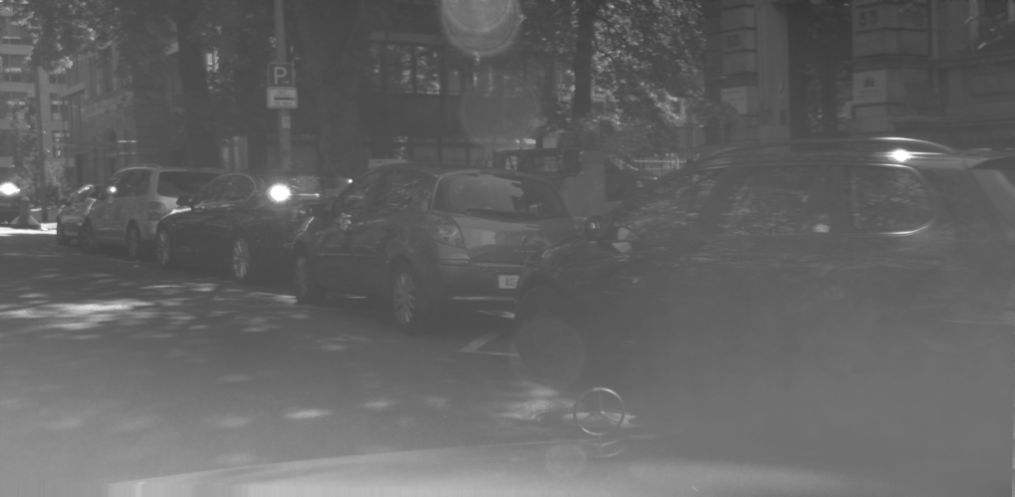}
\end{subfigure}
   \caption{An example of severe brightness change due to camera auto exposure/gain control.
   Despite the affine brightness correction our method still fails here due to
   the extreme brightness change as well as the strong rotational motion of the car.
   }\label{fig:cs_illumination}
   \vspace{-1.0em}
\end{figure}

To show the benefit of using a stereo camera, Fig~\ref{fig:qualitative_kitti} qualitatively compares the results of different monocular VO methods with our method on the sequences 00 and 06. Both state-of-the-art feature-based and direct monocular VO methods can not handle the scale drift properly. Results on all training sequences can be found in the supplementary material.

% =====================================================================================
\subsection{Cityscapes Dataset}
% =====================================================================================
To evaluate our method on more realistic data, we further run our Stereo DSO on the Cityscapes
Dataset~\cite{cordts2016cityscapes}. Although this dataset is dedicatedly designed for scene 
understanding and image segmentation, it also provides a long sequence (over 100,000
frames) captured using a stereo camera system. We choose this dataset as it provides
industrial level images with high dynamic-range (HDR) and fairly high frame rate (17Hz). Moreover,
in contrast to most existing datasets that provide street view sequences with global shutters,
the images of Cityscapes were captured using rolling shutter cameras, which has been considered as a main challenge to direct methods.
Although we do not address the rolling shutter calibration problem specifically in this paper,
our method implicitly makes use of two advantages of modern stereo camera systems:
Firstly, they have very fast pixel clock which reduces the rolling shutter effect; Secondly, the corresponding
rows of the left and right images are synchronized. Therefore, static stereo can compensate the errors
introduced by the rolling shutter effect across multiple views (multi-view stereo). 

\begin{figure*}[t!]
\begin{center}
\includegraphics[width=0.98\linewidth]{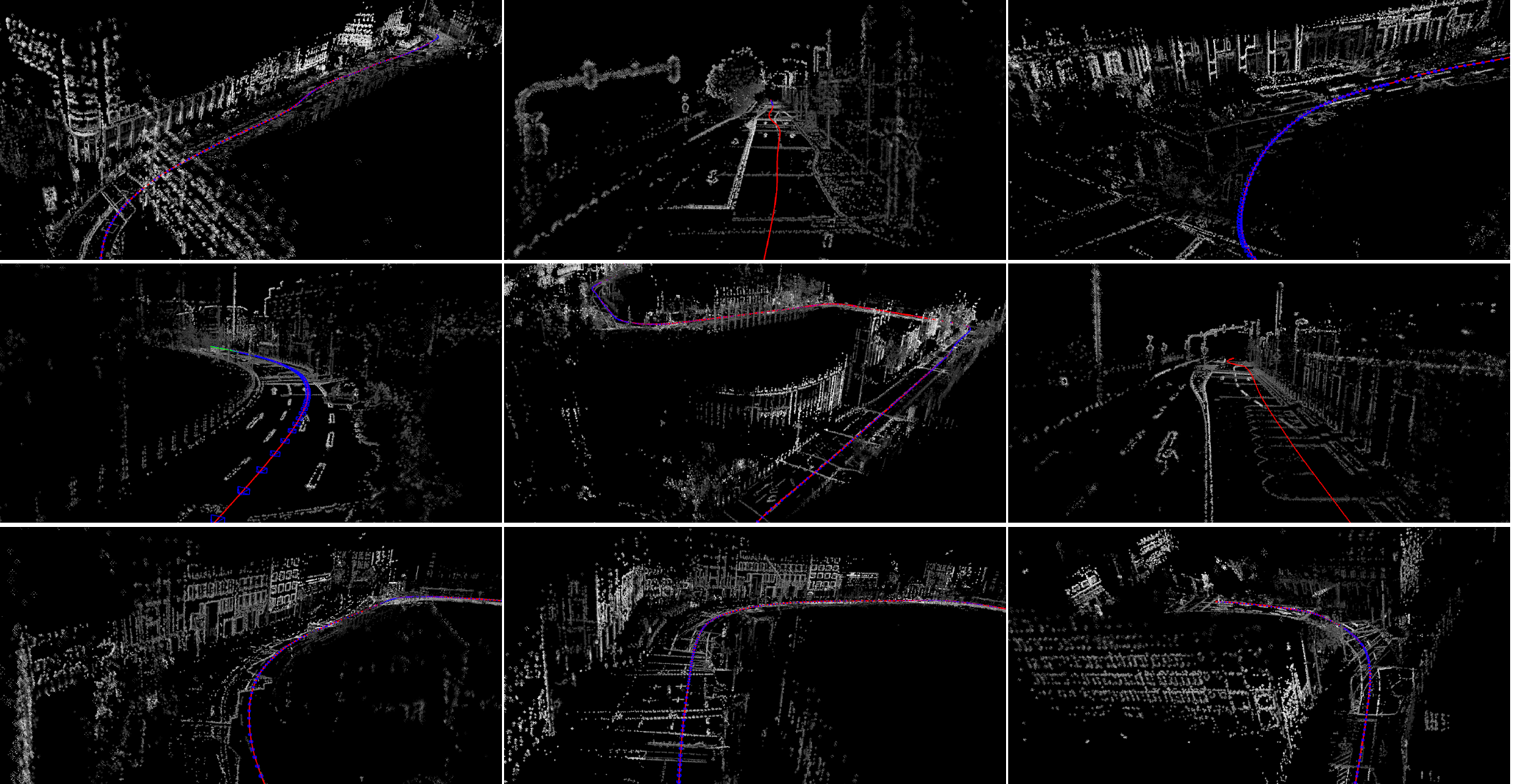}
\end{center}
\vspace{-0.5em}
   \caption{Examples of qualitative results on the Cityscapes Dataset.}\label{fig:qualitative_cs}
\vspace{-1.0em}
\end{figure*}

As for this dataset only inaccurate GPS/vehicle odometry information is provided,
it is hard to evaluate the VO performance accurately. Besides, this dataset has lots of 
moving objects such as cars driving right in front of the camera as well as severe uncalibrated brightness
changes (Fig~\ref{fig:cs_illumination}). Although our method has considered brightness changes
in the energy function, it seems the simple affine brightness modeling is not sufficient to handle
extreme cases. Our method fails in the scene
as shown in Fig~\ref{fig:cs_illumination}, where the severe brightness change is combined
with a strong rotational movement of the car. 

\begin{figure}
\centering
\begin{subfigure}[]{0.98\linewidth}
\includegraphics[width=0.48\linewidth]{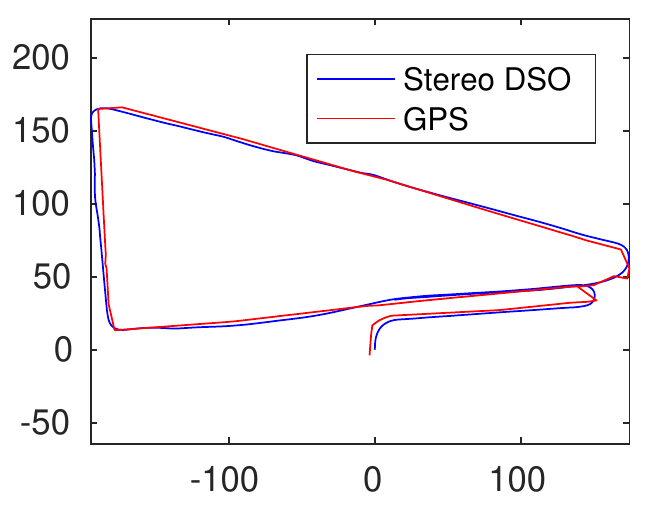}
\includegraphics[width=0.48\linewidth]{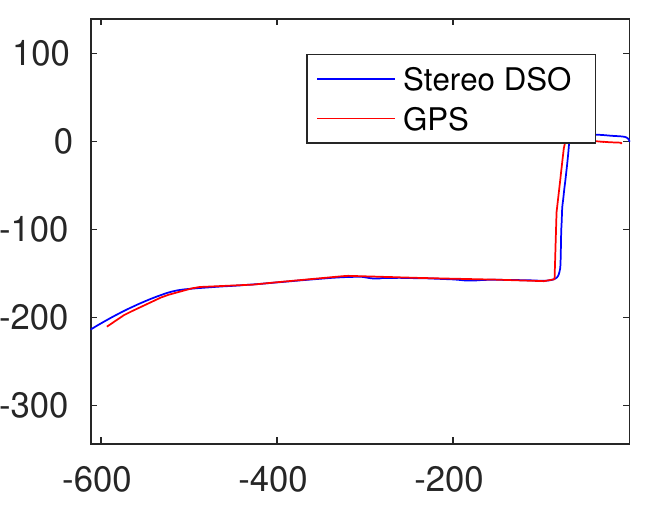}
\end{subfigure}
\begin{subfigure}[]{0.98\linewidth}
\includegraphics[width=0.48\linewidth]{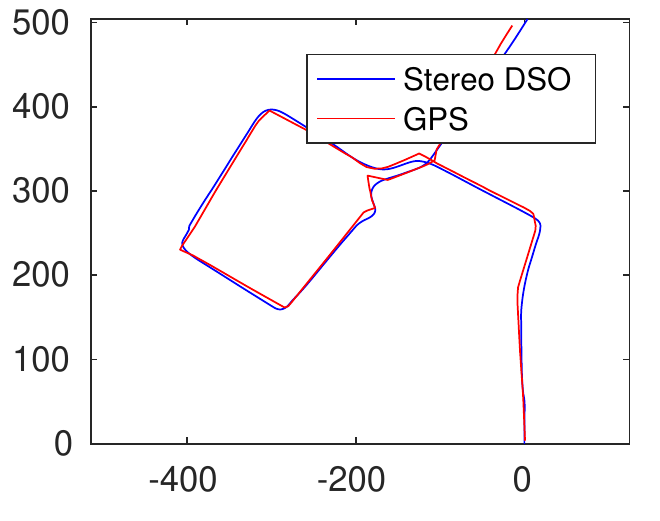}
\includegraphics[width=0.48\linewidth]{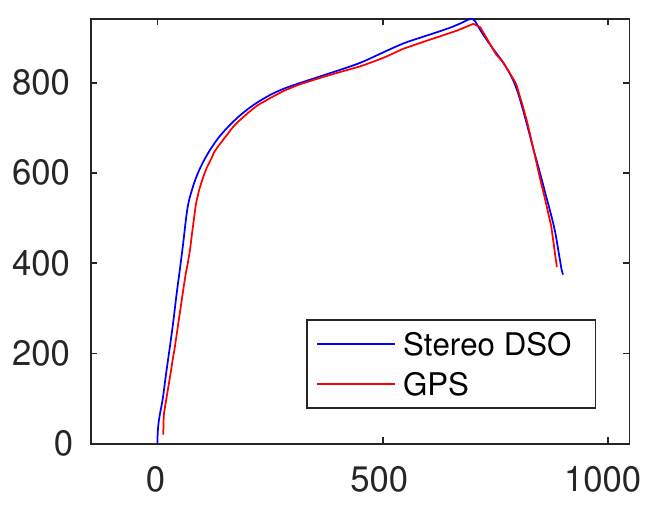}
\end{subfigure}
\vspace{-0.3em}
   \caption{Estimated trajectories on the Cityscapes Frankfurt stereo sequence. The sequences are obtained by dividing the long sequence into several smaller segments of 5000 to 6000 frames. 
   }\label{fig:cs_traj}
   \vspace{-1.0em}
\end{figure}

To evaluate our tracking accuracy, we divide the long sequence into several smaller ones with
length of 3000 to 6000 frames. For each small sequence, we calculate the ground truth camera
poses from the GPS coordinates using the Mercator projection and align these poses
to our trajectories using a SE(3) transformation. Some results of the estimated camera trajectories
can be found in Fig~\ref{fig:cs_traj}. As a big advantage of our method over
the feature-based methods, our VO approach creates precise and much denser 3D reconstructions.
Fig~\ref{fig:qualitative_cs} shows some reconstruction results on the Frankfurt sequence.
Although the reconstructions are
sparser than the ones from previous dense or semi-dense approaches, they are much more
accurate due to the bundle adjustment in the windowed optimization. More evaluation results
can be found in the supplementary material.

\section{Conclusion}
In this work, we introduced Stereo Direct Sparse Odometry as a
direct large-scale capable method for accurately tracking and mapping
from a stereo camera in real-time. We detailed the technical
implementation including the integration of temporal multi-view stereo
and static stereo within a marginalization framework using the Schur complement. Thorough qualitative and quantitative evaluations on the KITTI dataset
and the Cityscapes dataset demonstrate that Stereo DSO is the
currently most accurate and robust method for tracking a stereo camera
in challenging real-world scenarios. In particular, an evaluation on
the KITTI testing set showed that even without closing large loops,
Stereo DSO provides more accurate results than Stereo ORB-SLAM2 with loop closuring and global bundle adjustment.  

%In our view, this work is another indicator
%that on the long run direct visual SLAM algorithms will outperform the
%more classical keypoint based approaches, because they take into
%account all available image information (rather than a heuristically
%selected subset) and because they can pick up even subtle brightness
%changes on otherwise featureless areas.

In future work, we plan to extend our approach to a full SLAM system
by adding loop closuring and a database for map maintenance. Besides, we
also consider explicit dynamic object handling to further
boost the VO accuracy and robustness.\\ 

\noindent{\textbf{Acknowledgments}. This work was supported by the ERC
Consolidator Grant ``3D Reloaded''. We would like to thank Jakob Engel
and Vladyslav Usenko for their supports and fruitful discussions. }

{\small
\bibliographystyle{ieee}
\bibliography{Bib}
}

\title{Supplementary Material \\Stereo DSO: \\Large-Scale Direct Sparse Visual Odometry with Stereo Cameras}
\author{Rui Wang$^{\ast} $ , Martin Schw{\"o}rer$^{\ast} $, Daniel Cremers\\
Technical University of Munich\\
{\tt\small\{wangr, schwoere, cremers\}@in.tum.de}}
\date{\vspace{-3ex}}
\maketitle

\pagenumbering{arabic}
\setcounter{page}{1}
\setcounter{section}{0}
\setcounter{figure}{0}
\setcounter{table}{0}
\setcounter{equation}{0}

\begin{abstract}
In this \textbf{supplementary document}, we first show how weighting the constraints from static stereo differently influences the tracking accuracy. Next,
we provide the full
trajectory estimations on the training set of KITTI with comparisons to state-of-the-art
monocular VO methods. Afterwards we show more results on the
Cityscapes Frankfurt sequence, which qualitatively demonstrates the tracking
accuracy of our method. We also provide a \textbf{supplementary video}\footnote{\url{https://youtu.be/A53vJO8eygw}.} to show the performance of
our method on the selected datasets, as well as the quality of the
delivered 3D reconstructions.
\end{abstract}
\section{Effect of Stereo Coupling Factor}
\begin{figure}
\begin{center}
\includegraphics[width=0.98\linewidth]{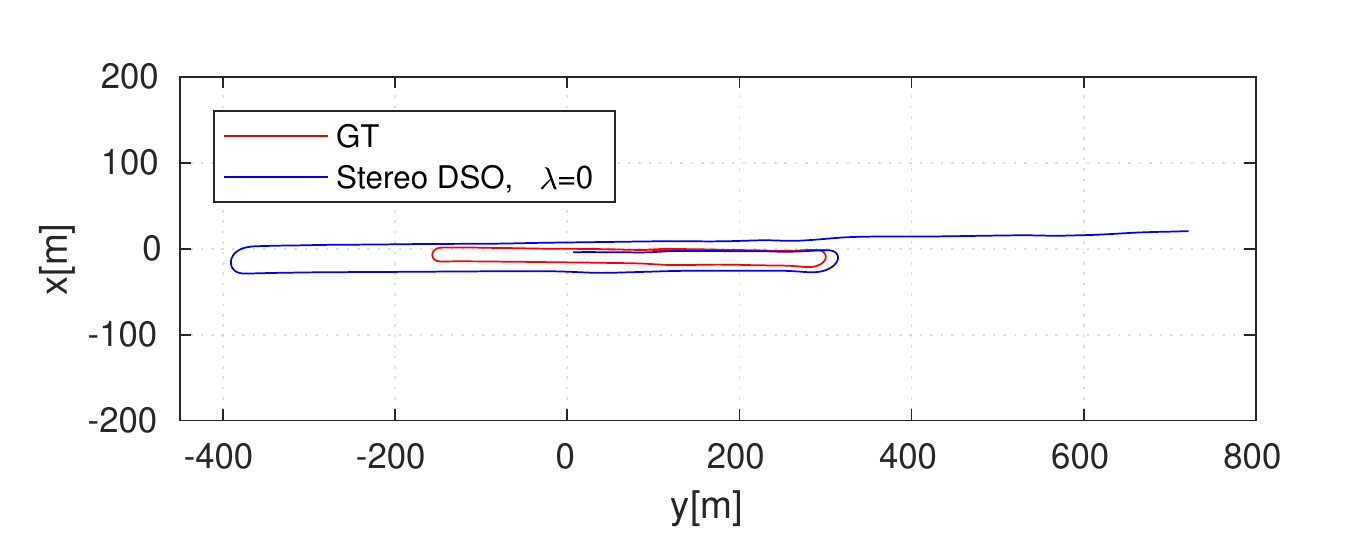}\vspace{-0.5em}
\includegraphics[width=0.98\linewidth]{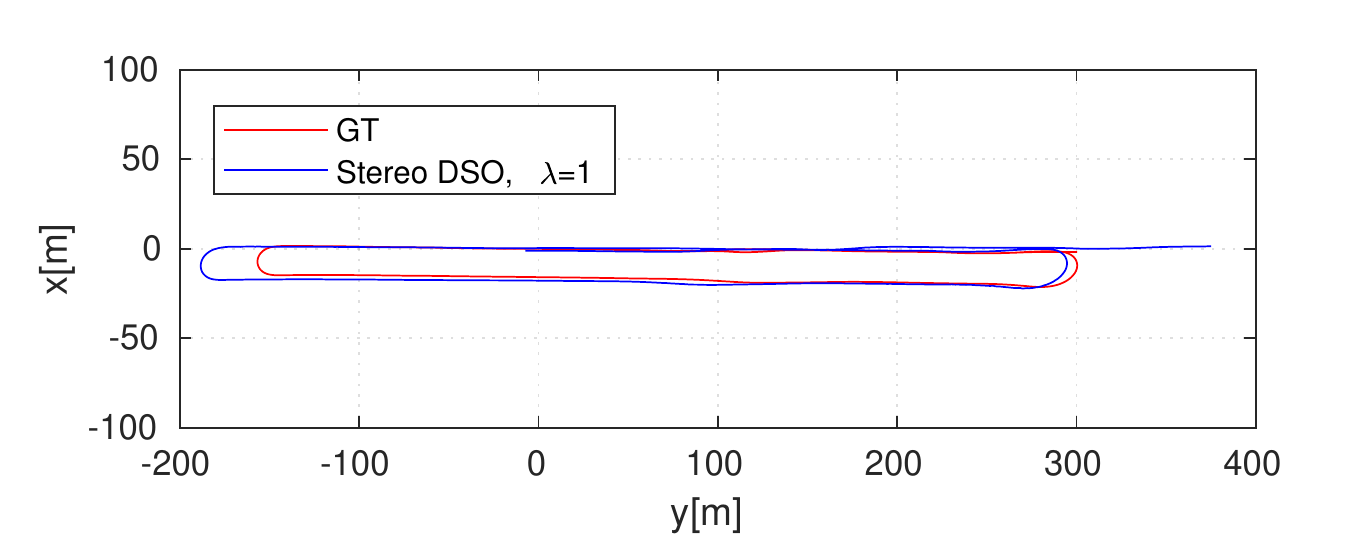}\vspace{-0.5em}
\includegraphics[width=0.98\linewidth]{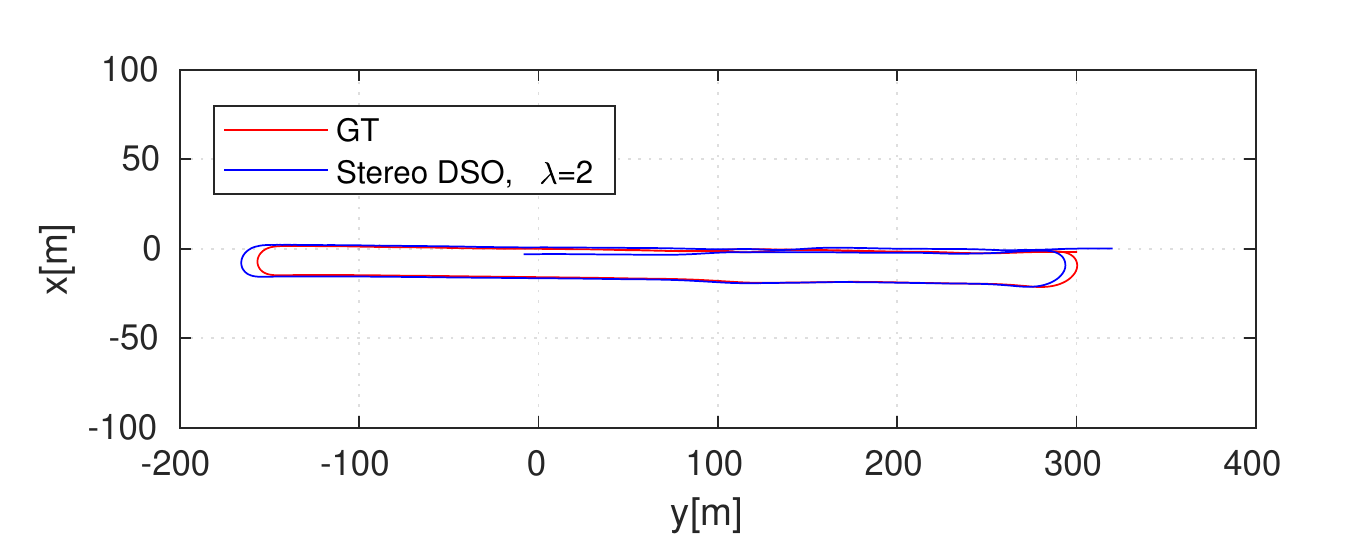}\vspace{-0.5em}
\includegraphics[width=0.98\linewidth]{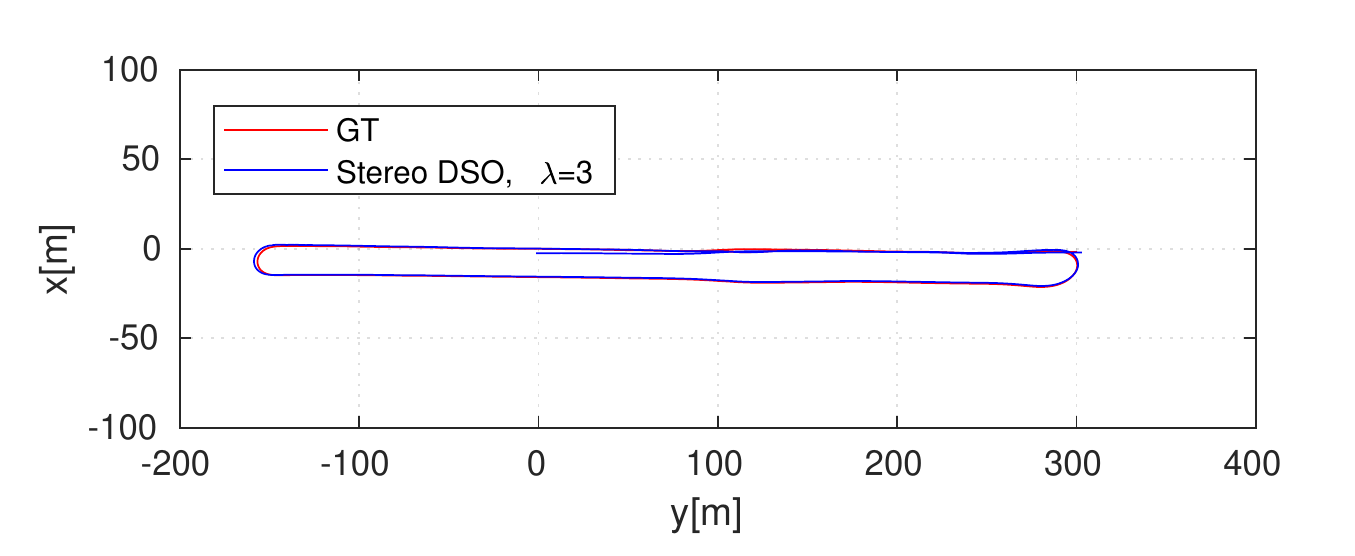}
\end{center}
   \caption{Trajectories on KITTI Seq. 06 using coupling
   factors ($\lambda$) 0-3. Increasing the weighting of the static stereo constraints significantly reduces the translational drift.}\label{plot:coupling_traj}
\end{figure}

The estimated trajectories on KITTI Seq. 06 using coupling factors ($\lambda$) 0-3
are shown in Fig~\ref{plot:coupling_traj}.

%\begin{figure}
%\begin{center}
%\includegraphics[width=0.98\linewidth]{figures/coupling.pdf}
%\end{center}
%   \caption{Average translational and rotational errors on Seq. 06 of using different
%   coupling factors. The translational errors are plotted with logarithmic scale.
%   Introducing constraints from static stereo with certain weightings ($\lambda=1,2$) significantly
%   reduces both translational and rotational errors. Further increasing the weighting ($\lambda>3$)
%   makes the method more sensitive to incorrect matchings from static stereo. 
%   }\label{plot:coupling_06}
%\end{figure}

\section{Full Results on KITTI}
%\begin{figure}
%\begin{center}
%\includegraphics[width=0.9\linewidth]{figures/ts.pdf}
%\includegraphics[width=0.9\linewidth]{figures/rs.pdf}
%\end{center}
%   \caption{Average translational and rotational errors with respect to driving speed 
%   on KITTI testing set (Seq. 11-21). In most of the cases, our VO results (without
%   loop closure) are better than the SLAM results of LSD-SLAM and ORB-SLAM2 (with loop closure, for ORB-SLAM2 also
%   with global bundle adjustment).}\label{plot:kitti_testing}
%\end{figure}
Fig~\ref{plot:kitti_trajectories} shows our trajectory estimates for all training sequences of KITTI (left) 
and their comparisons to the ground truth (right). To show the improvements over monocular methods, the results of the monocular ORB-SLAM (VO only) and monocular DSO are shown in the middle. 
The trajectory estimates of the monocular methods are aligned
to the ground truth using a similarity transformation (7DoF), while the results
of our method are aligned using a rigid-body transformation (6DoF). Obviously, scale
drift is the main problem of the monocular methods, which can be resolved by using stereo cameras.
In addition, monocular DSO seems to have larger scale drift than monocular ORB-SLAM.
We believe this results from the sensitivity of direct methods to the low frame rate, large optical flow, as well as other unmodeled effects in the image domain, such as non-lambertian reflectance and illumination changes that
have not been corrected sufficiently.

%In
%Fig \ref{} we show our results on the testing set. Since no ground truth poses
%are provided, we only show the estimated trajectories with their plots.

\begin{figure}[!t]
\begin{center}
\includegraphics[width=1.0\linewidth]{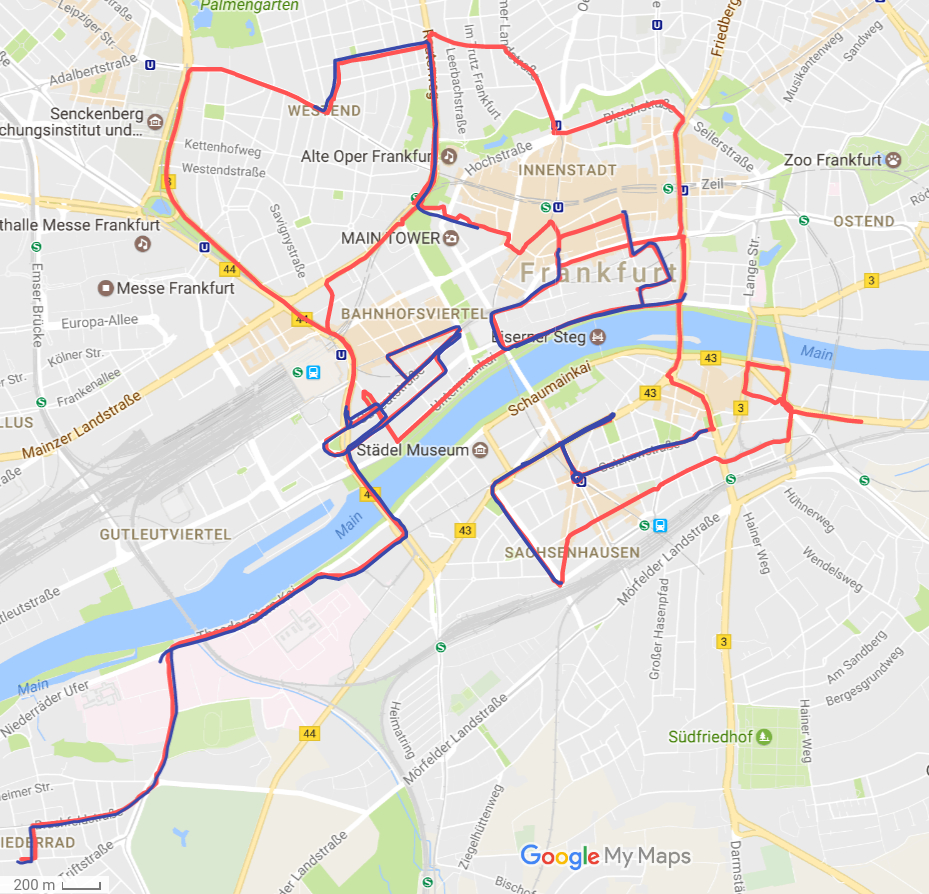}
\end{center}
   \caption{The full Frankfurt trajectory superimposed on the corresponding Google Map
   scene. The trajectory estimates of the subsections (blue) are aligned to the ground truth trajectory (orange) using a rigid-body transformation (6DoF). Best viewed printed.}\label{fig:frankfurt}
\end{figure}

\section{More Results on Cityscapes}
\begin{figure*}
    \centering
    \begin{subfigure}[t]{0.32\textwidth}
        \centering
        \includegraphics[width=\textwidth]{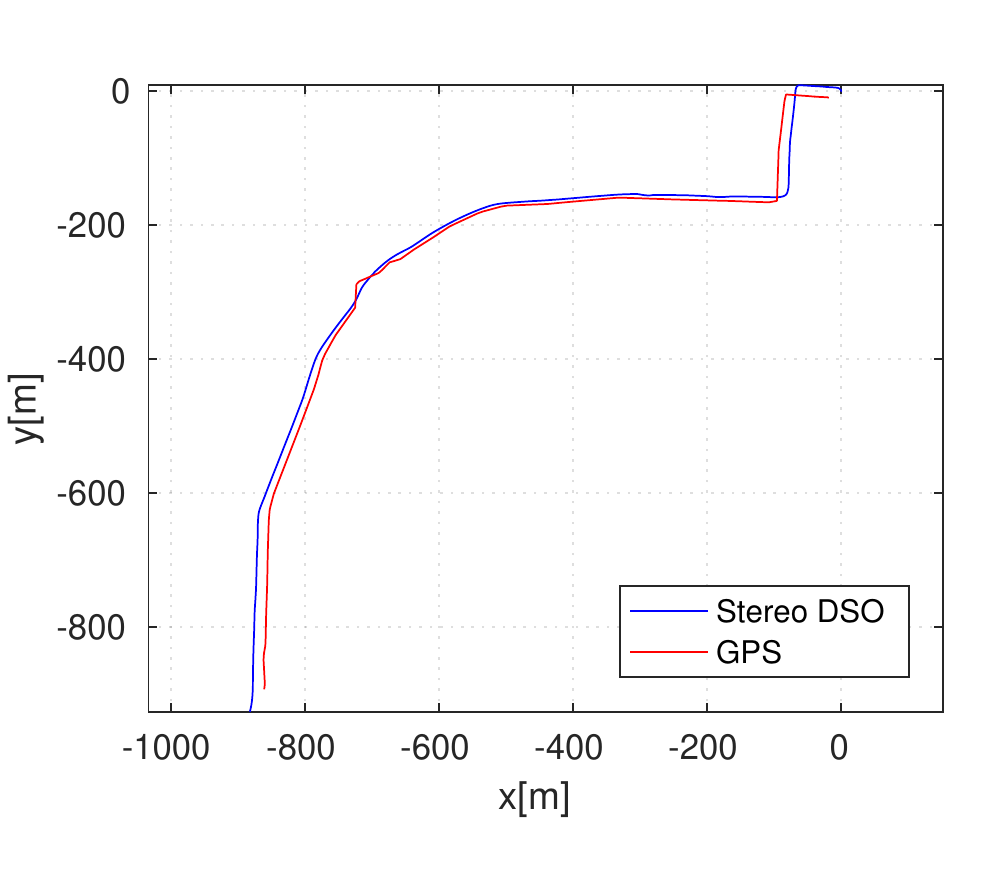}
        \caption{1-6000}\label{fig:11} 
    \end{subfigure} 
    ~
    \begin{subfigure}[t]{0.32\textwidth}
        \centering
        \includegraphics[width=\textwidth]{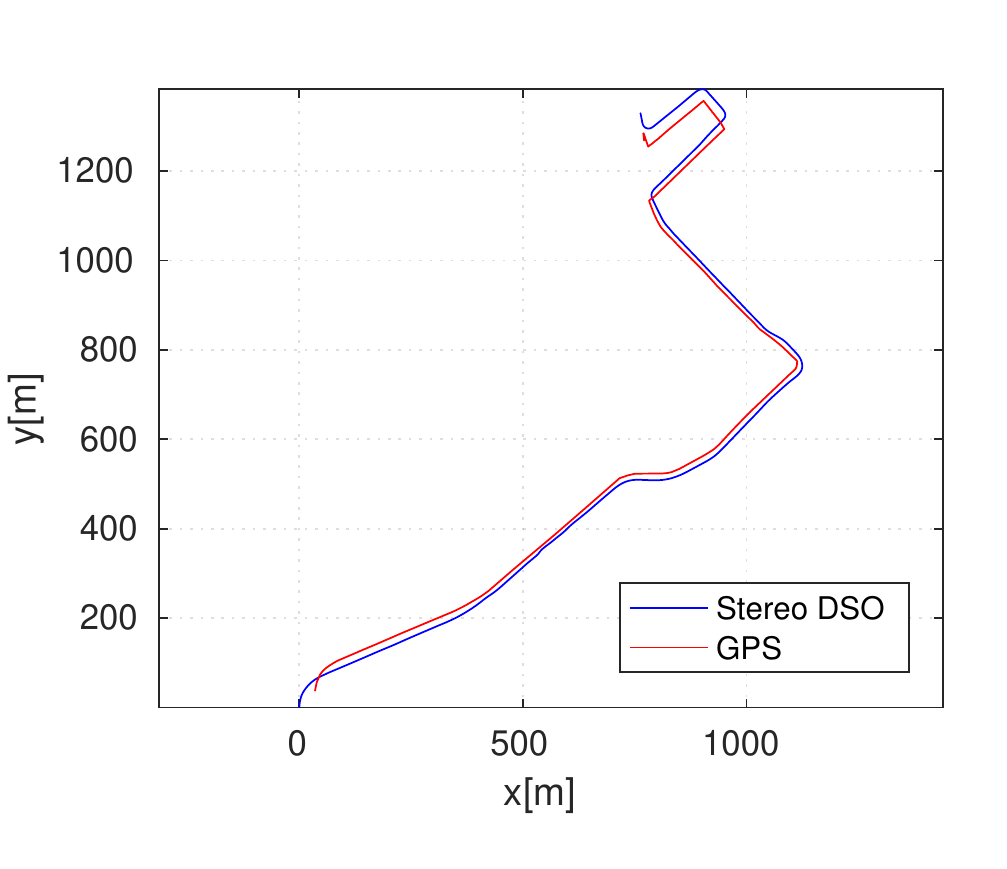}
        \caption{6001-12000}\label{fig:12}
    \end{subfigure} 
    ~
    \begin{subfigure}[t]{0.32\textwidth}
        \centering
        \includegraphics[width=\textwidth]{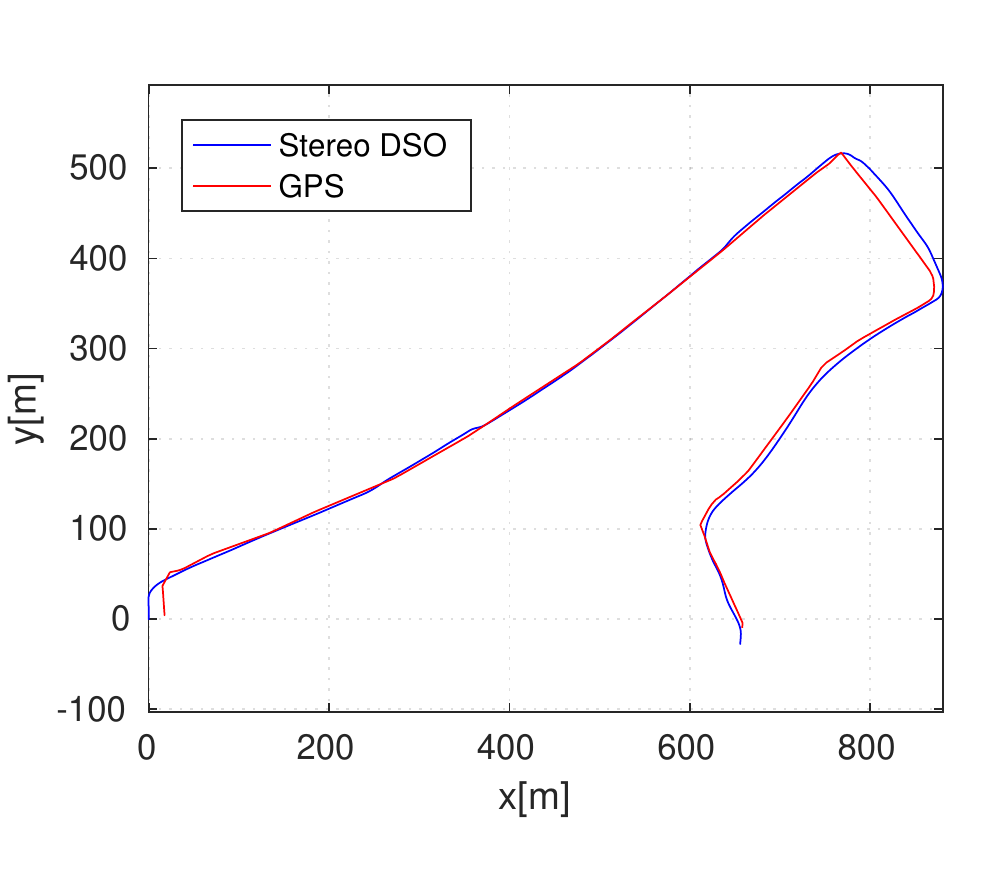}
        \caption{27001-33000}\label{fig:13}
    \end{subfigure} 
    \begin{subfigure}[t]{0.32\textwidth}
        \centering
        \includegraphics[width=\textwidth]{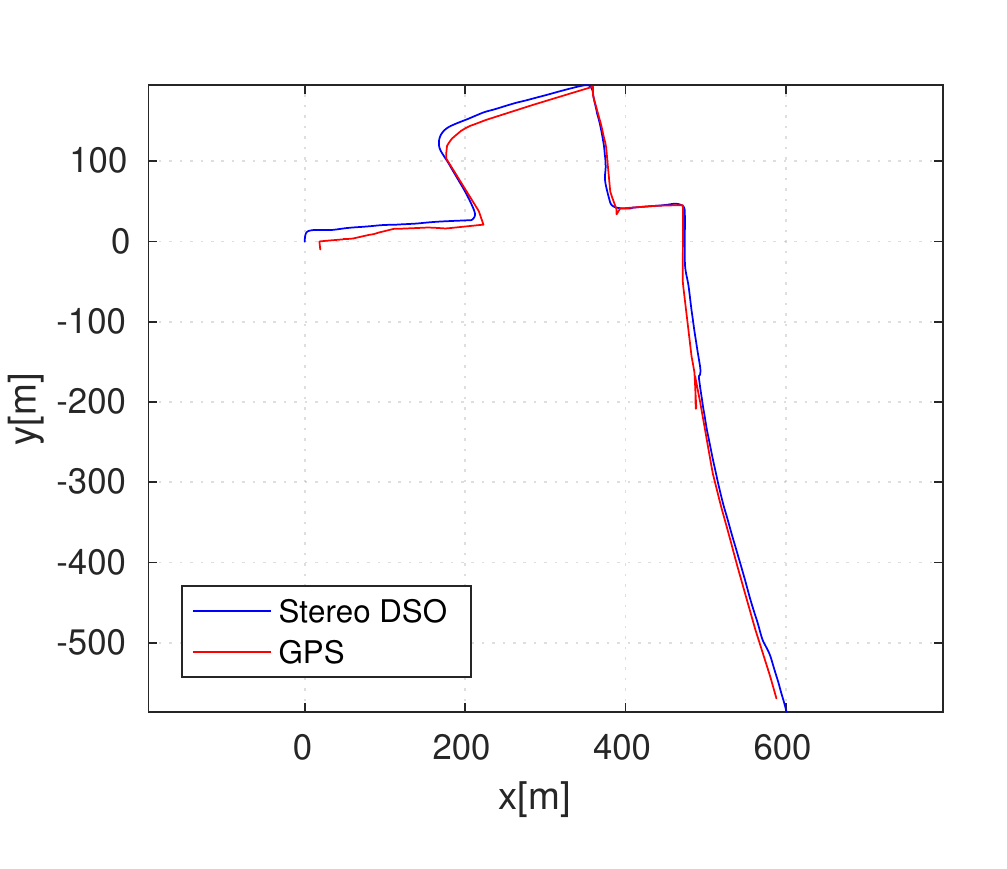}
        \caption{36001-42000}\label{fig:21} 
    \end{subfigure} 
    ~
    \begin{subfigure}[t]{0.32\textwidth}
        \centering
        \includegraphics[width=\textwidth]{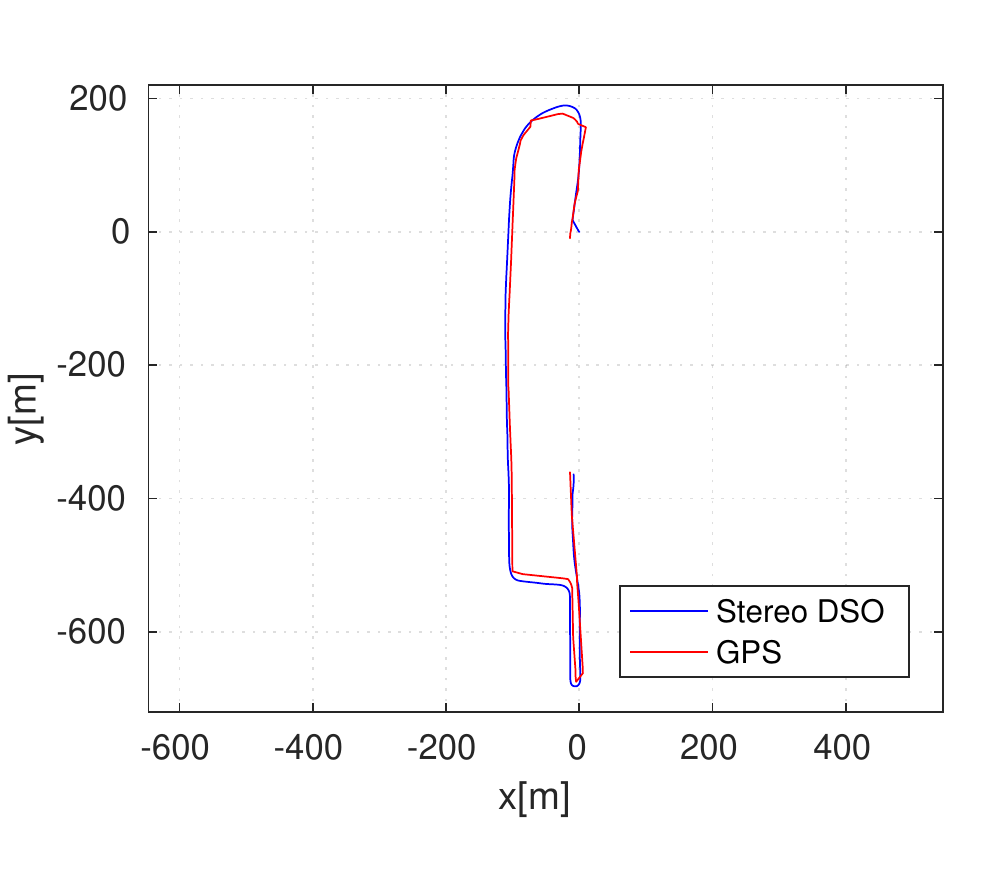}
        \caption{48001-54000}\label{fig:22}
    \end{subfigure} 
    ~
    \begin{subfigure}[t]{0.32\textwidth}
        \centering
        \includegraphics[width=\textwidth]{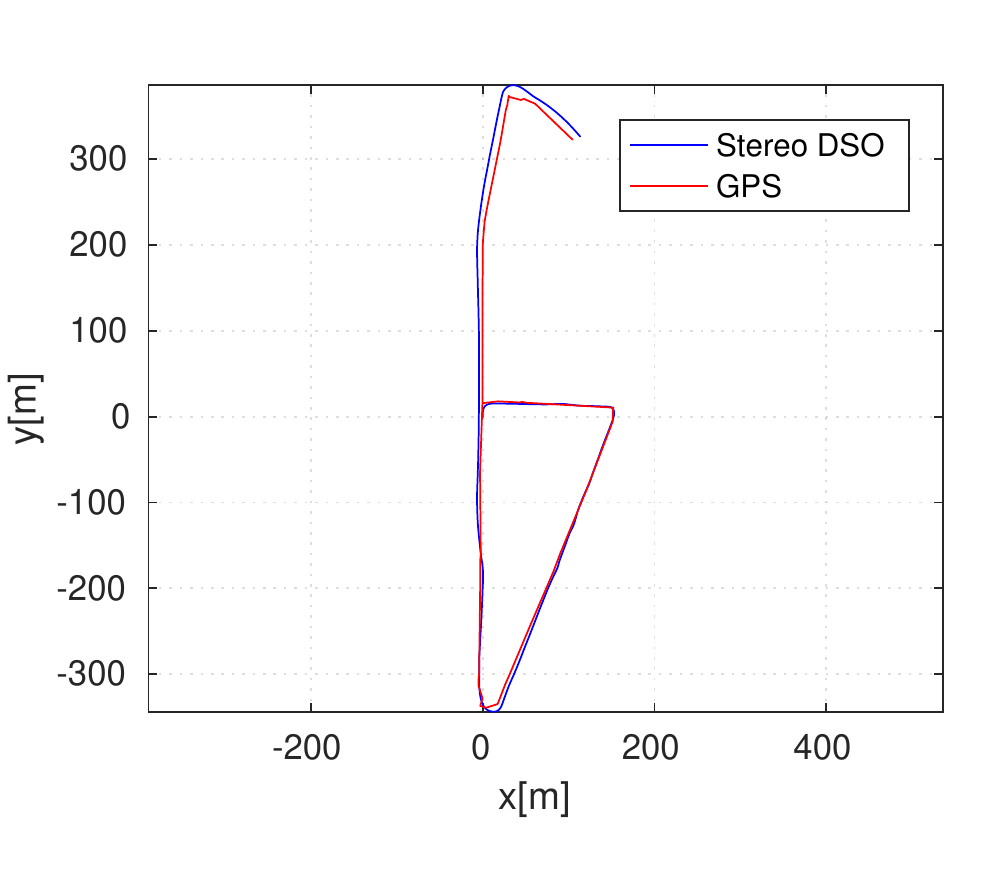}
        \caption{54001-60000}\label{fig:23}
    \end{subfigure} 
    \begin{subfigure}[t]{0.32\textwidth}
        \centering
        \includegraphics[width=\textwidth]{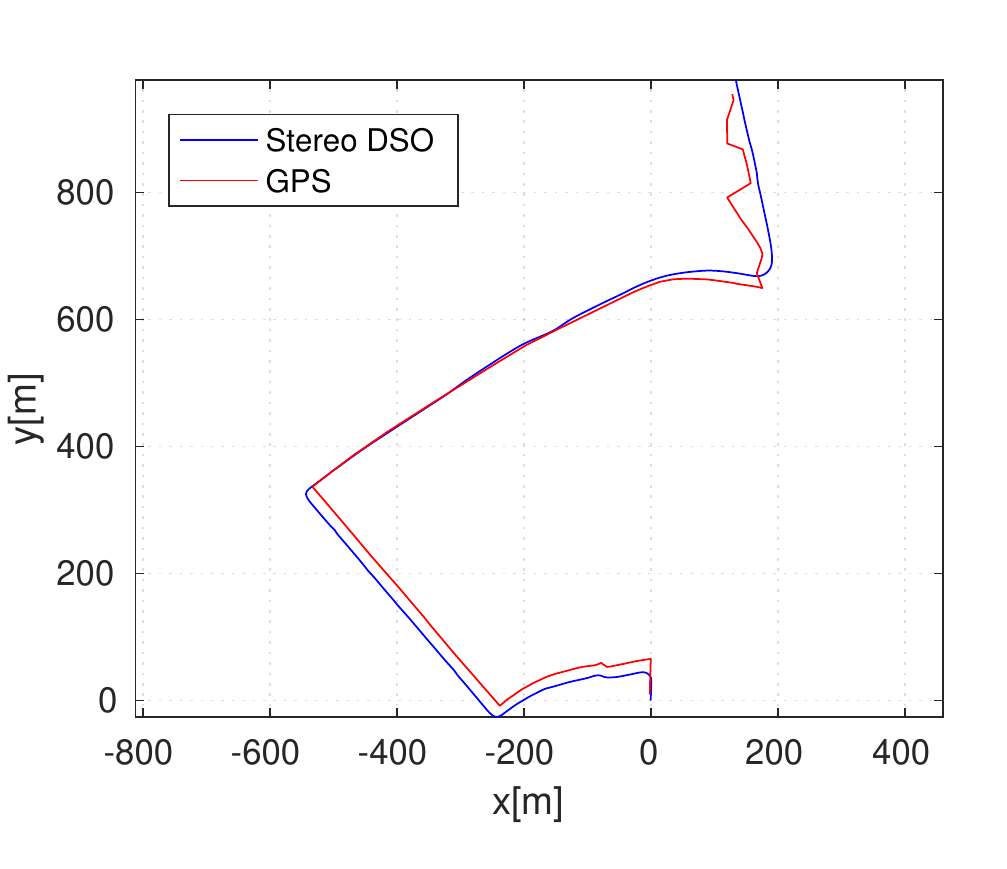}
        \caption{69001-55000}\label{fig:31} 
    \end{subfigure} 
    ~
    \begin{subfigure}[t]{0.32\textwidth}
        \centering
        \includegraphics[width=\textwidth]{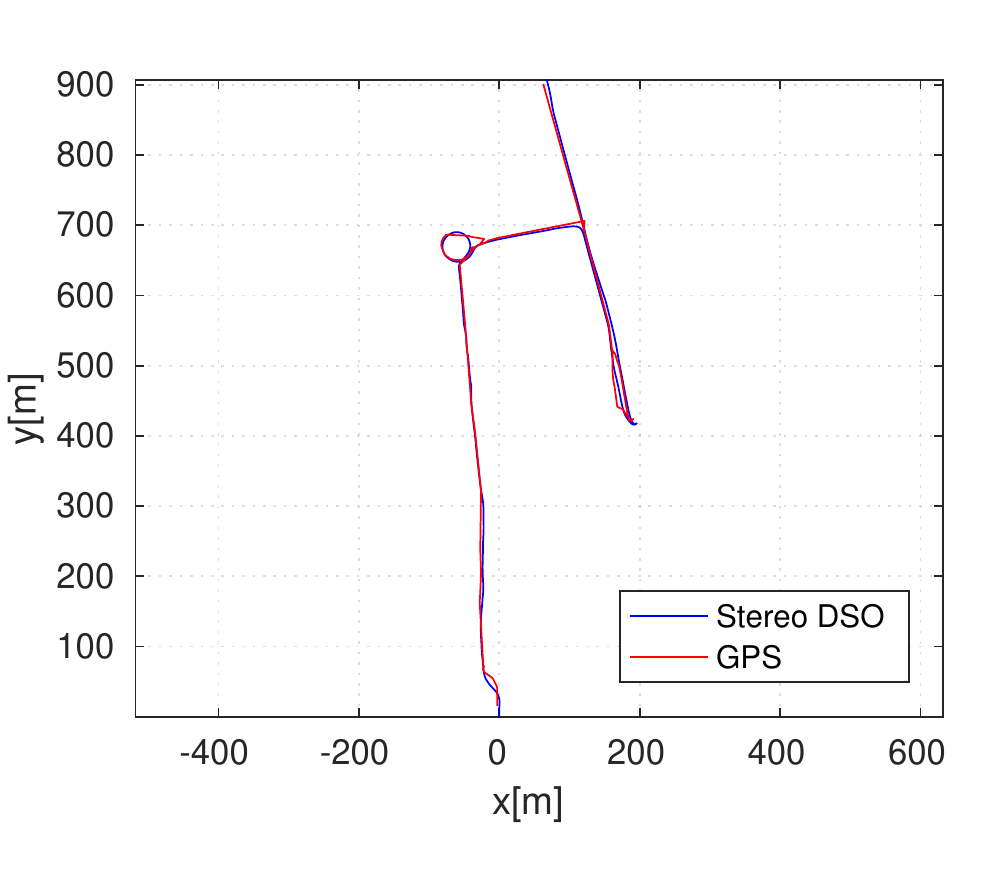}
        \caption{87001-93000}\label{fig:32}
    \end{subfigure} 
    ~
    \begin{subfigure}[t]{0.32\textwidth}
        \centering
        \includegraphics[width=\textwidth]{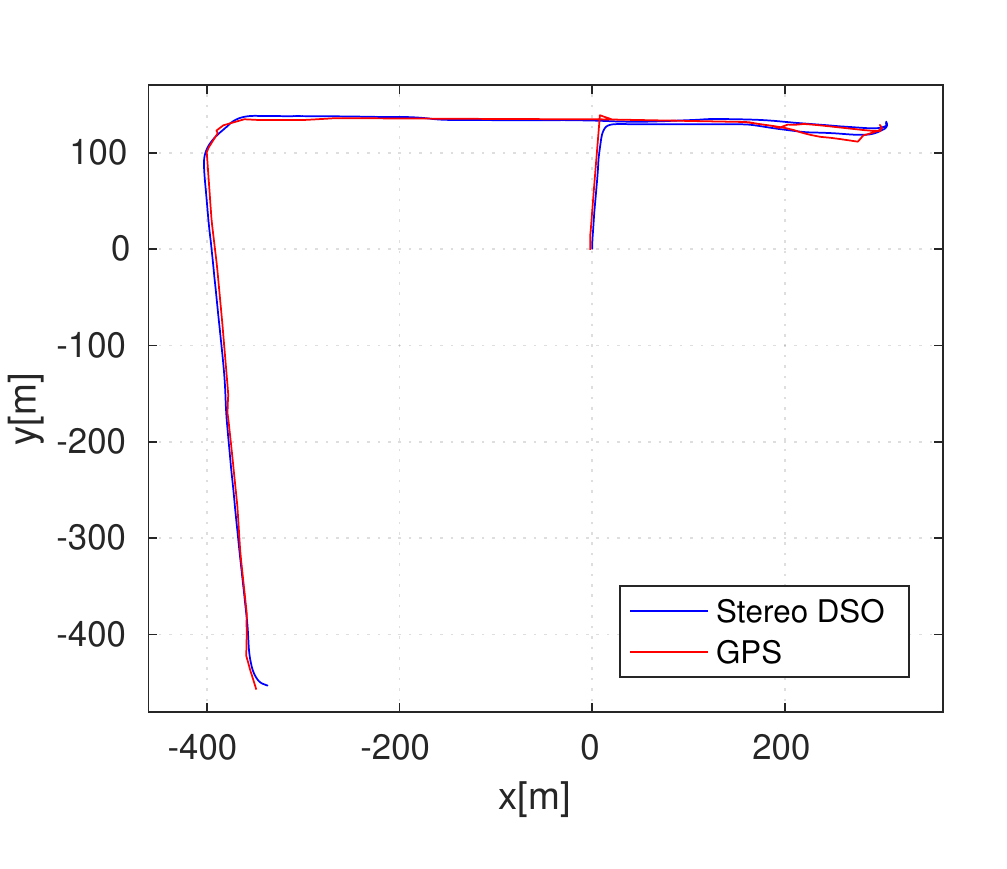}
        \caption{90001-96000}\label{fig:33}
    \end{subfigure} 
   \caption{Estimated camera trajectories on the Cityscapes Frankfurt stereo sequence.
   The sequences are obtained by dividing the full sequence into several smaller sections
   with length of 5000 to 6000 frames and coverages comparable to the sequences of
   KITTI. The sub-captions name the corresponding frame indices in the full sequence. The ground truth poses are calculated from the provided GPS coordinates using the Mercator projection. In some figures, \eg Fig~\ref{fig:31}, the inaccuracies of the GPS are clearly visible.}\label{plot:cs_trajectories}
\end{figure*}

As mentioned in the main paper, without specifically handling moving objects and sudden strong brightness changes, our method is currently
not able to run on the entire Frankfurt sequence (around 107,000 frames). Therefore,
we divide the full sequence into several smaller sections, each with a length of 5000-6000
frames resulting in a comparable coverage to the KITTI sequences. Exemplary results with
ground truth are shown in Fig~\ref{plot:cs_trajectories}. The plots of a few sections reveal that the ground truth poses calculated from the provided GPS coordinates are not always accurate. 
In Fig~\ref{fig:frankfurt} we show the ground truth trajectory of the entire sequence as well as the estimated
trajectories aligned to it. 

\begin{figure*}
    \centering
    \begin{subfigure}[]{0.98\textwidth}
        \includegraphics[width=0.32\textwidth]{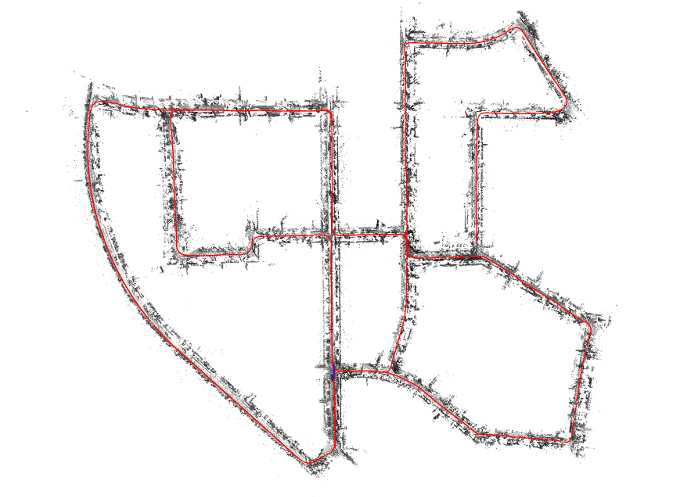}
        \includegraphics[width=0.32\textwidth]{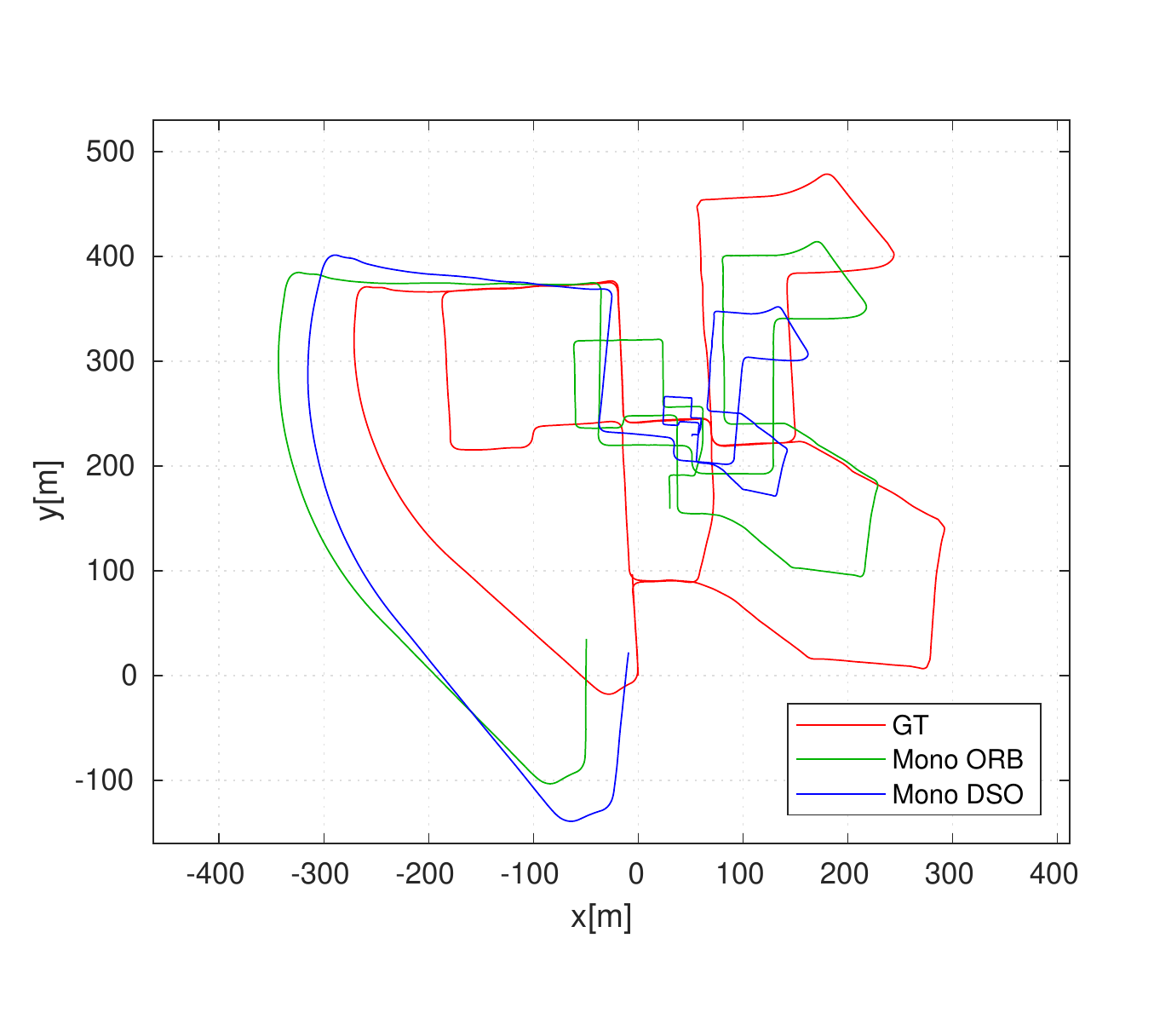}
        \includegraphics[width=0.32\textwidth]{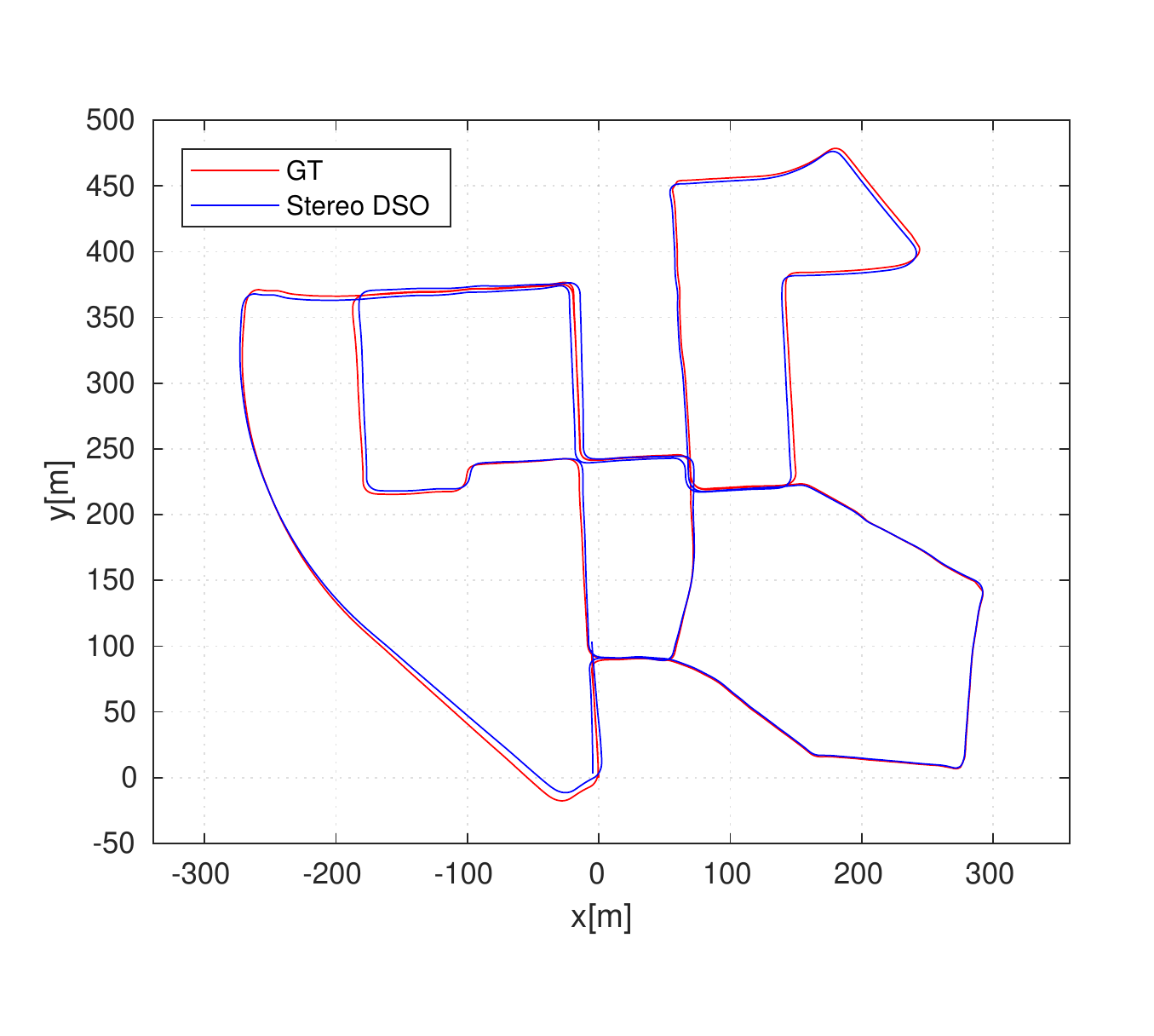}
        \vspace{-1\baselineskip}
        \caption{Seq. 00}
    \end{subfigure} 
    \vspace{-0.25em}

    \begin{subfigure}[]{0.98\textwidth}
        \includegraphics[width=0.32\textwidth]{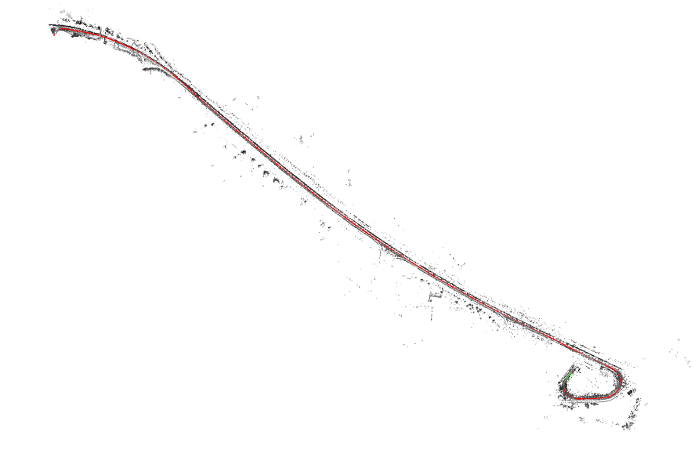}
        \includegraphics[width=0.32\textwidth]{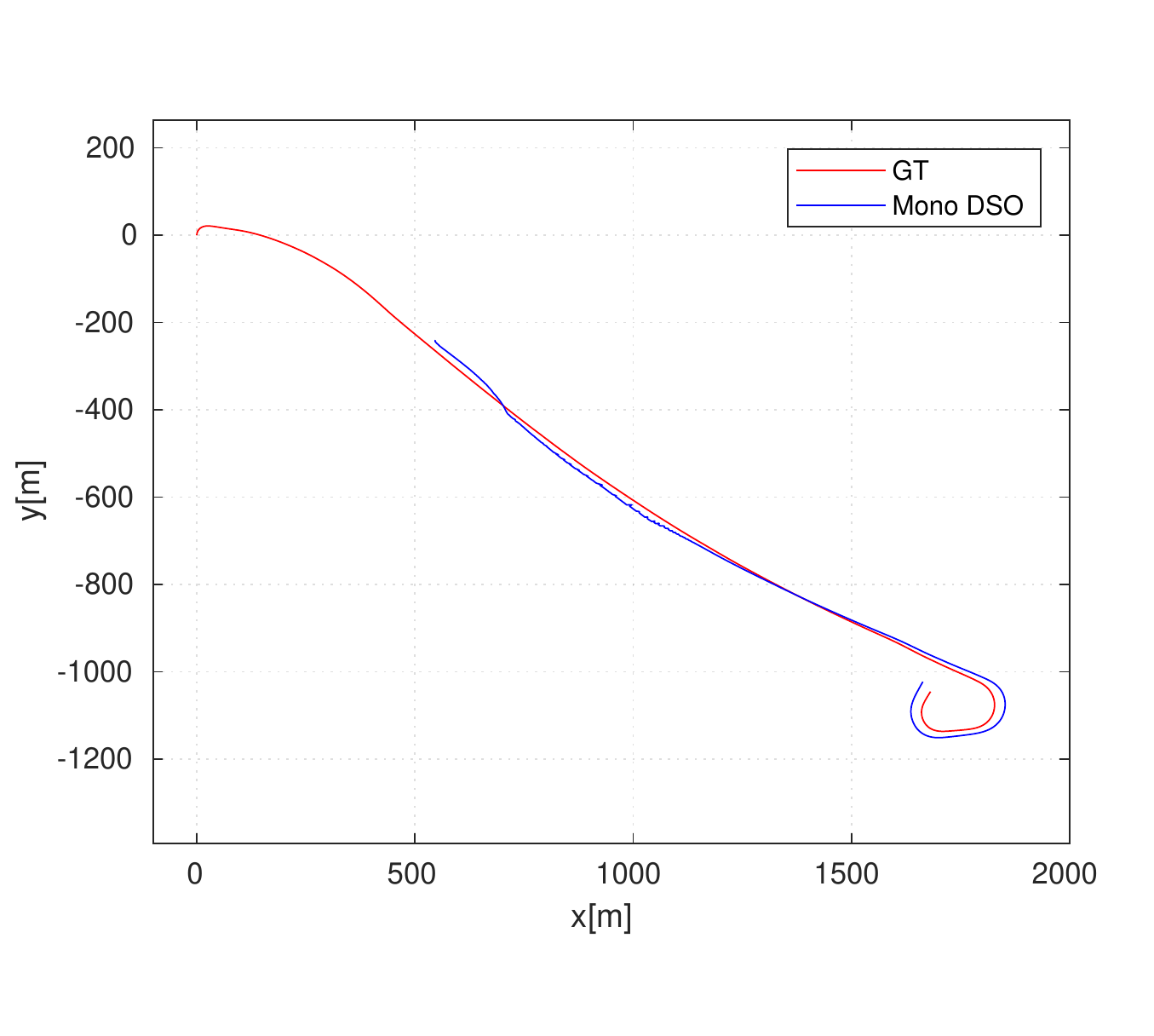}
        \includegraphics[width=0.32\textwidth]{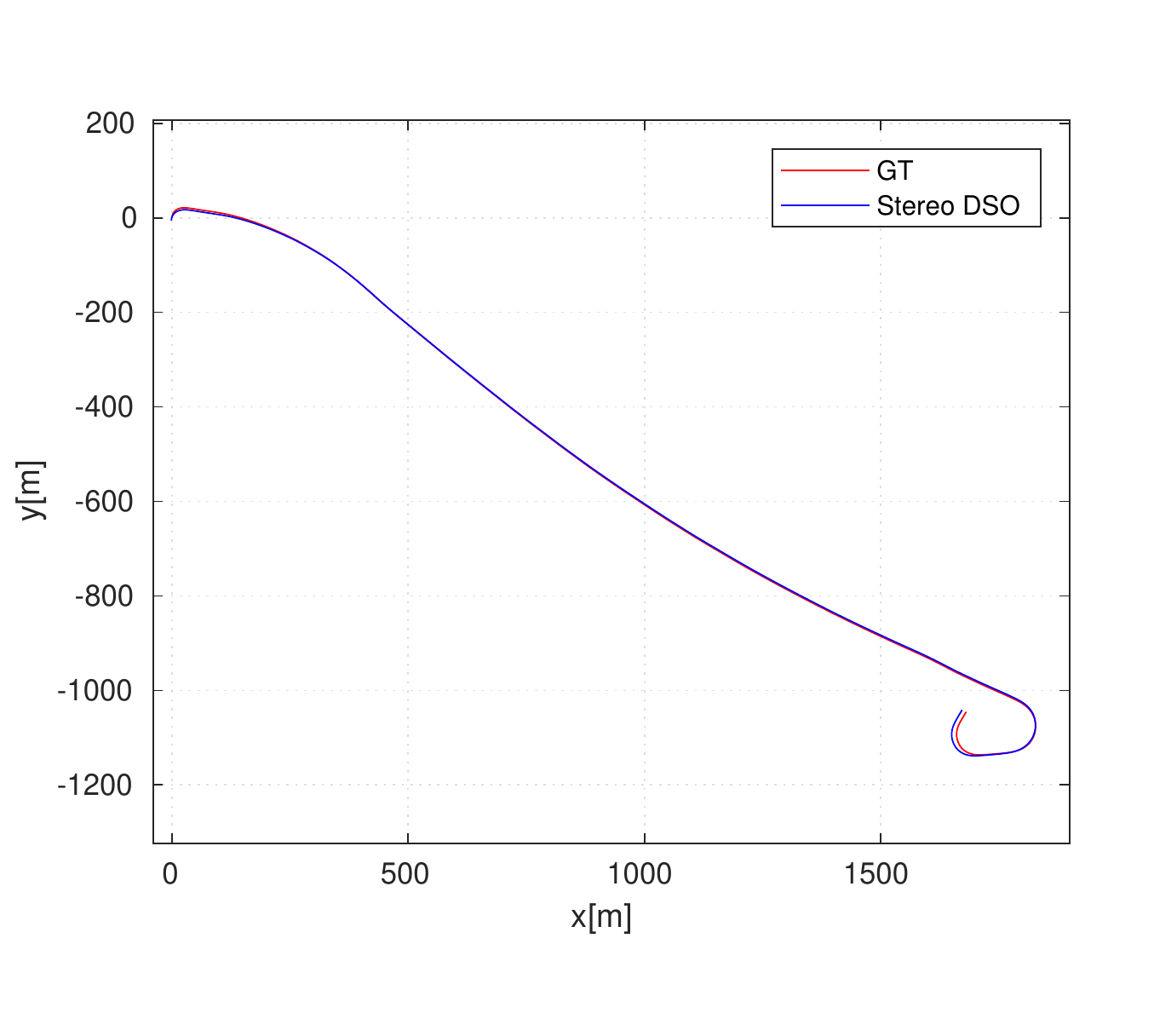}
        \vspace{-1\baselineskip}
        \caption{Seq. 01 (ORB-SLAM fails on this sequence)}
    \end{subfigure} 
    \vspace{-0.25em}

        \begin{subfigure}[]{0.98\textwidth}
        \includegraphics[width=0.32\textwidth]{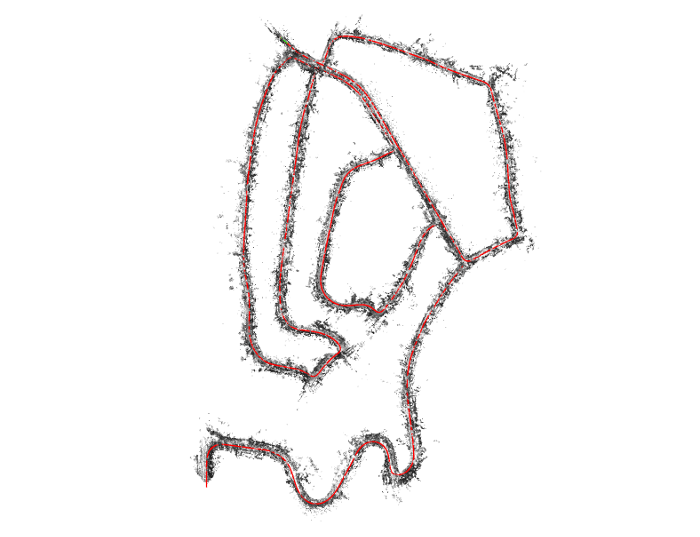}
        \includegraphics[width=0.32\textwidth]{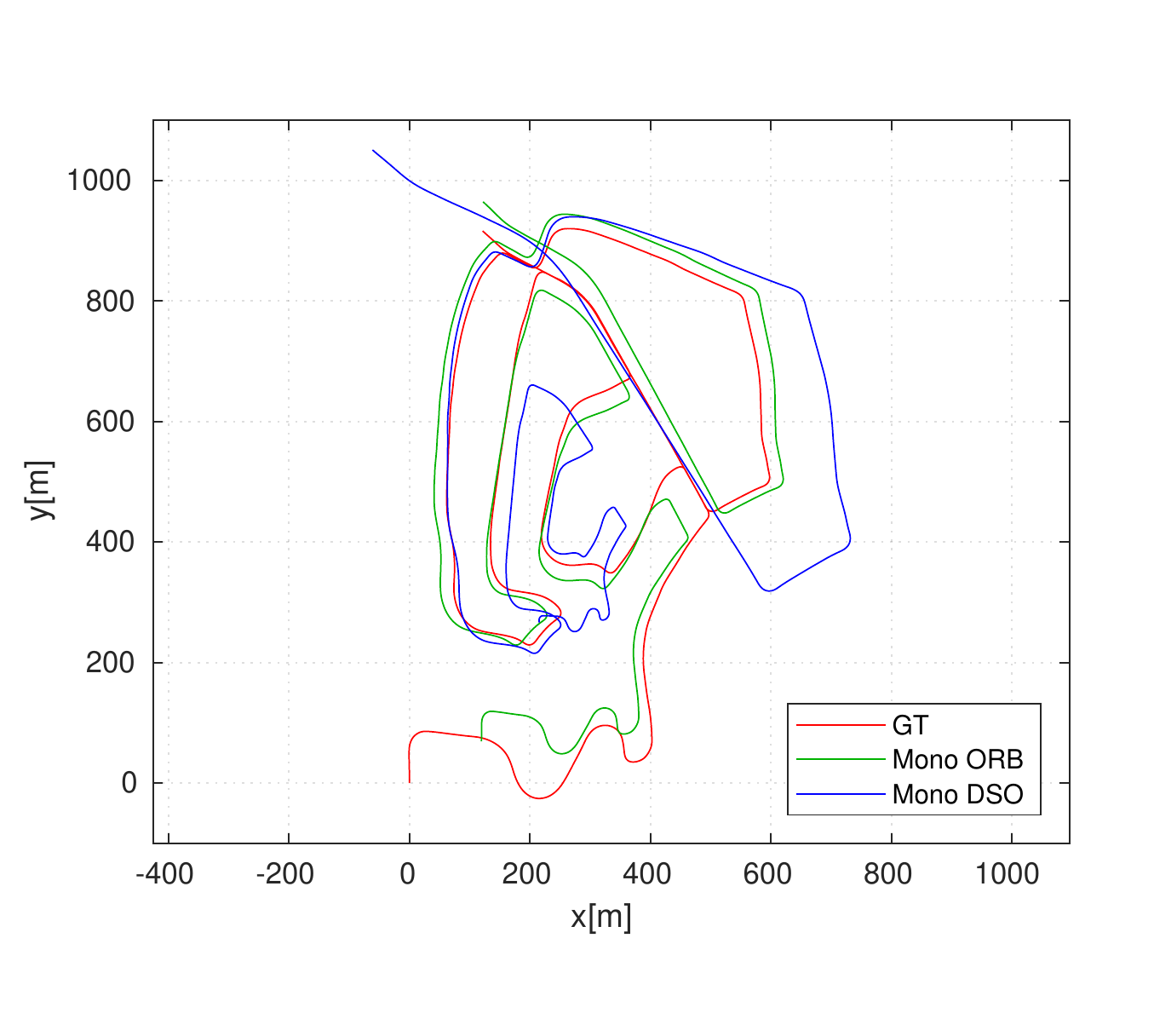}
        \includegraphics[width=0.32\textwidth]{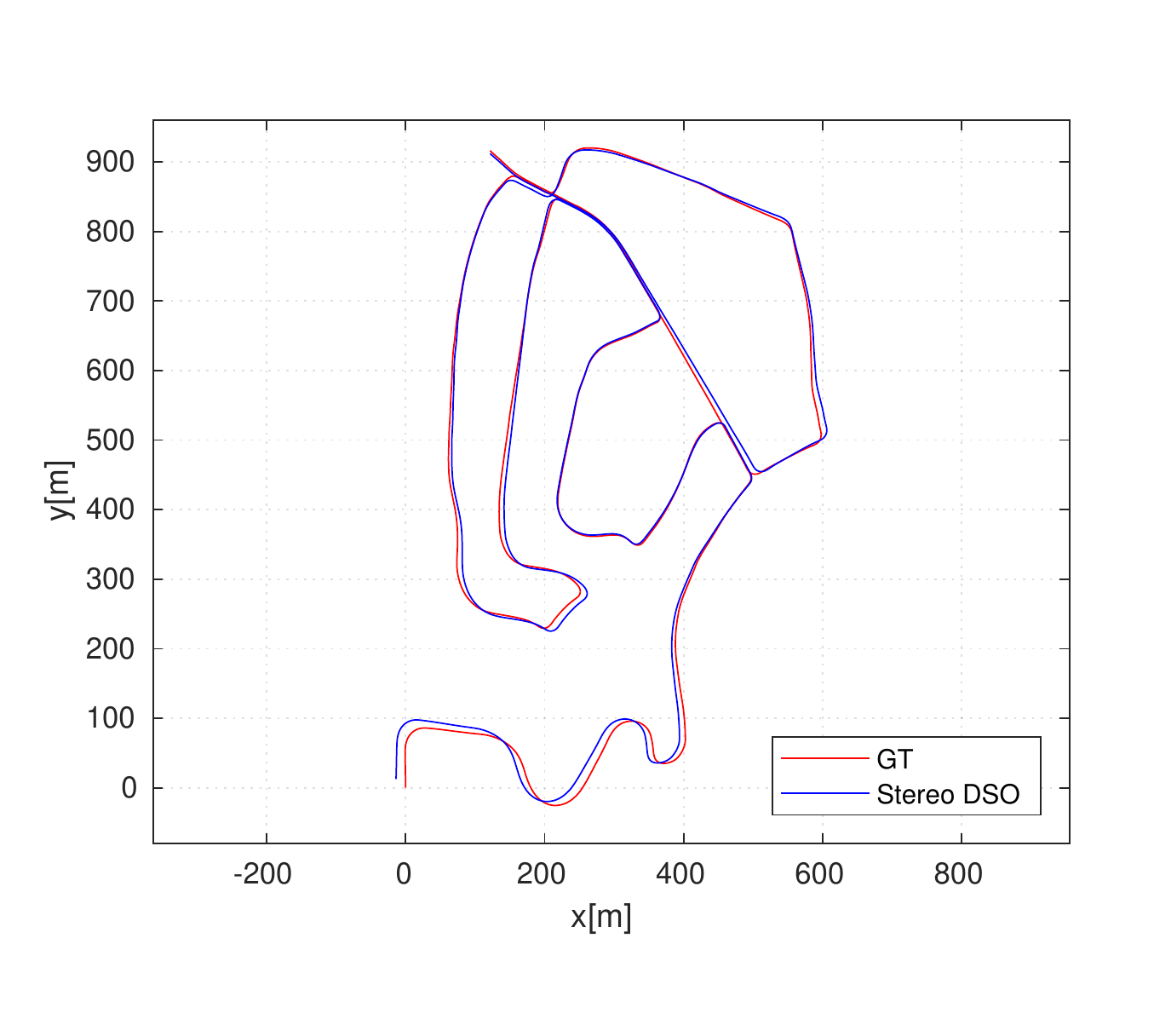}
        \vspace{-1\baselineskip}
        \caption{Seq. 02}
    \end{subfigure} 
    \vspace{-0.25em}

        \begin{subfigure}[]{0.98\textwidth}
        \includegraphics[width=0.32\textwidth]{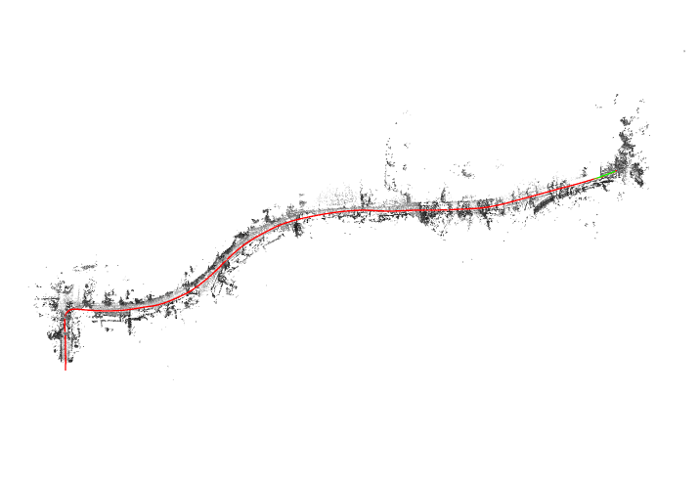}
        \includegraphics[width=0.32\textwidth]{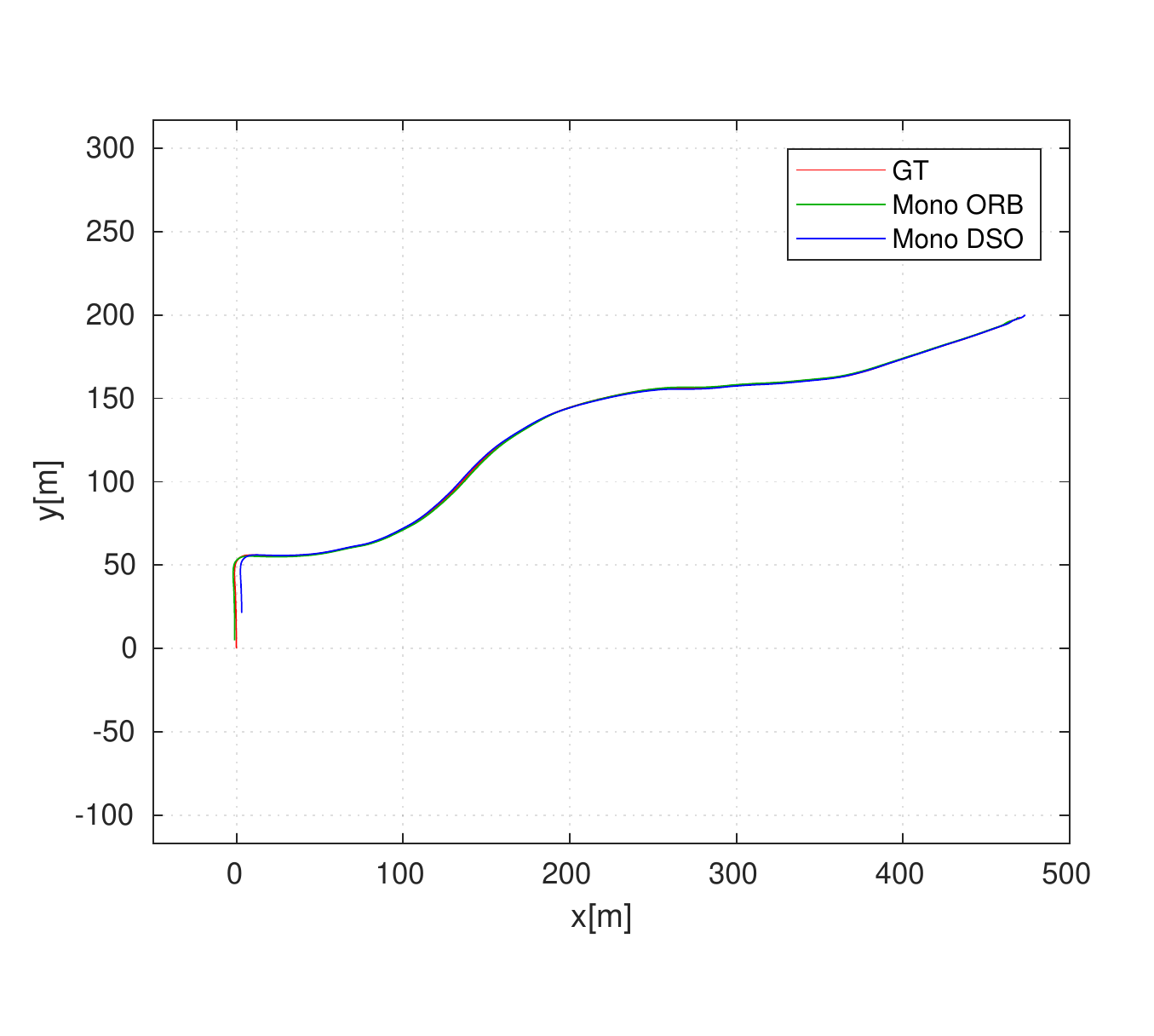}
        \includegraphics[width=0.32\textwidth]{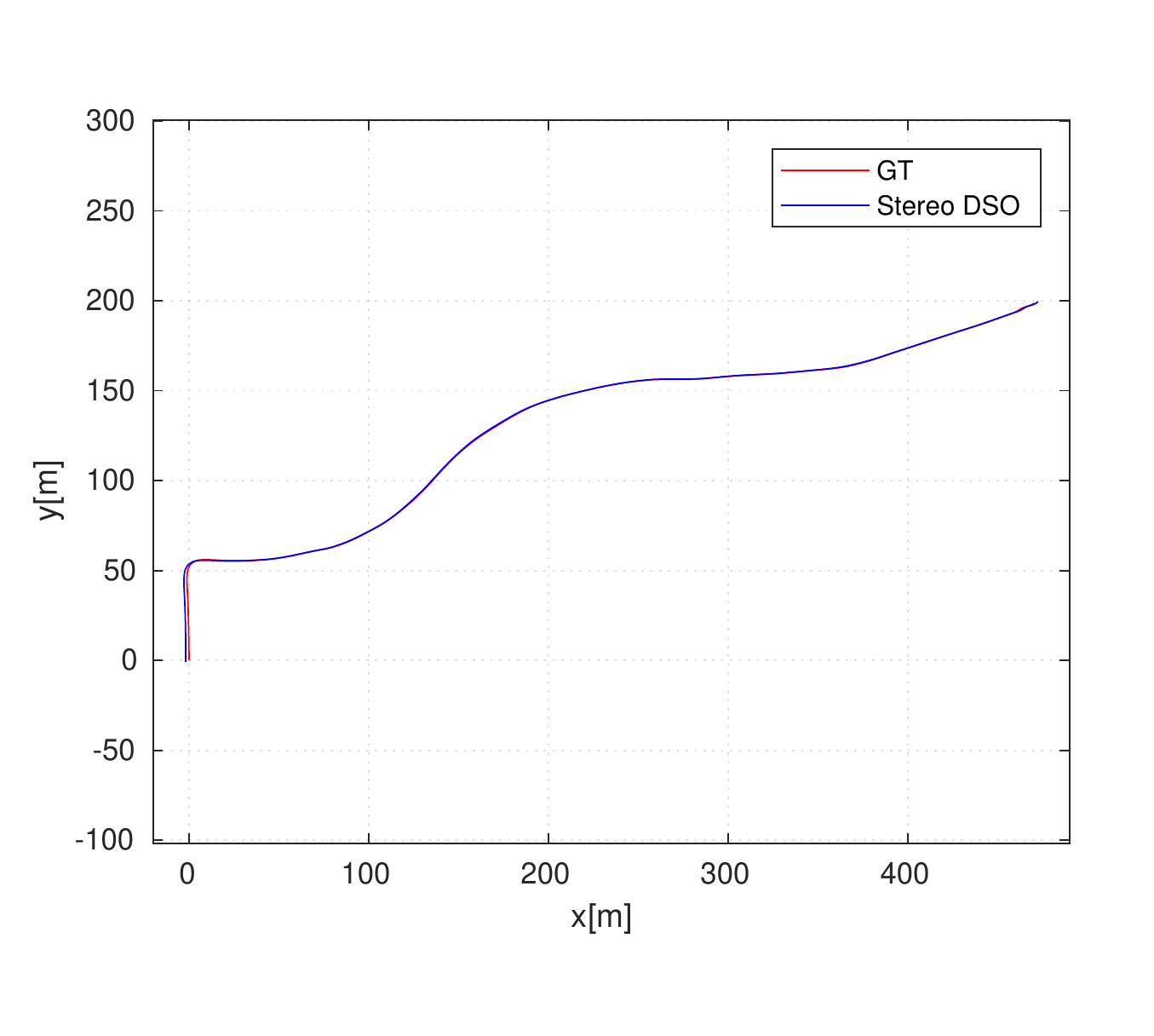}
        \vspace{-1\baselineskip}
        \caption{Seq. 03}
    \end{subfigure} 

\caption{Full results on the KITTI training set. The trajectories estimated by our VO method
are shown in the left column. The comparisons to the ground truth are shown in the right
column. The results of the two state-of-the-art monocular VO methods, namely ORB-SLAM (VO only) and DSO, are shown in the middle. The trajectories are aligned to the ground truth using similarity transformations (7DoF) and rigid-body
transformations (6DoF) for the monocular methods and our stereo method respectively.}\label{plot:kitti_trajectories}
\end{figure*}

\begin{figure*}\ContinuedFloat
    \centering
        \begin{subfigure}[]{0.98\textwidth}
        \includegraphics[width=0.32\textwidth]{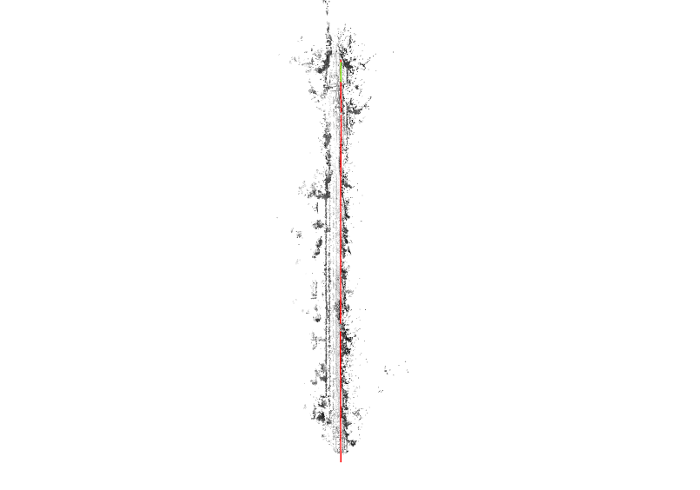}
        \includegraphics[width=0.32\textwidth]{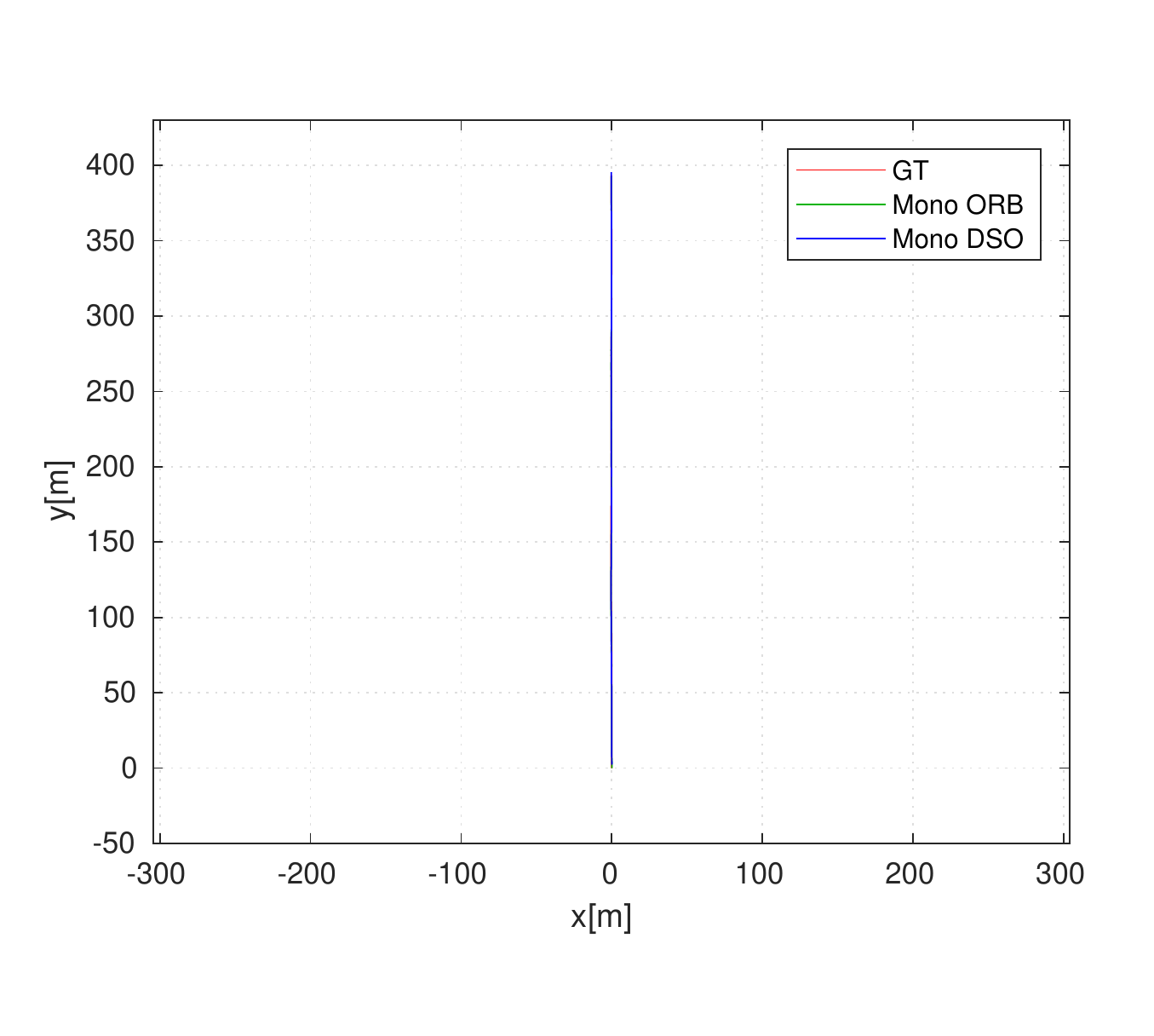}
        \includegraphics[width=0.32\textwidth]{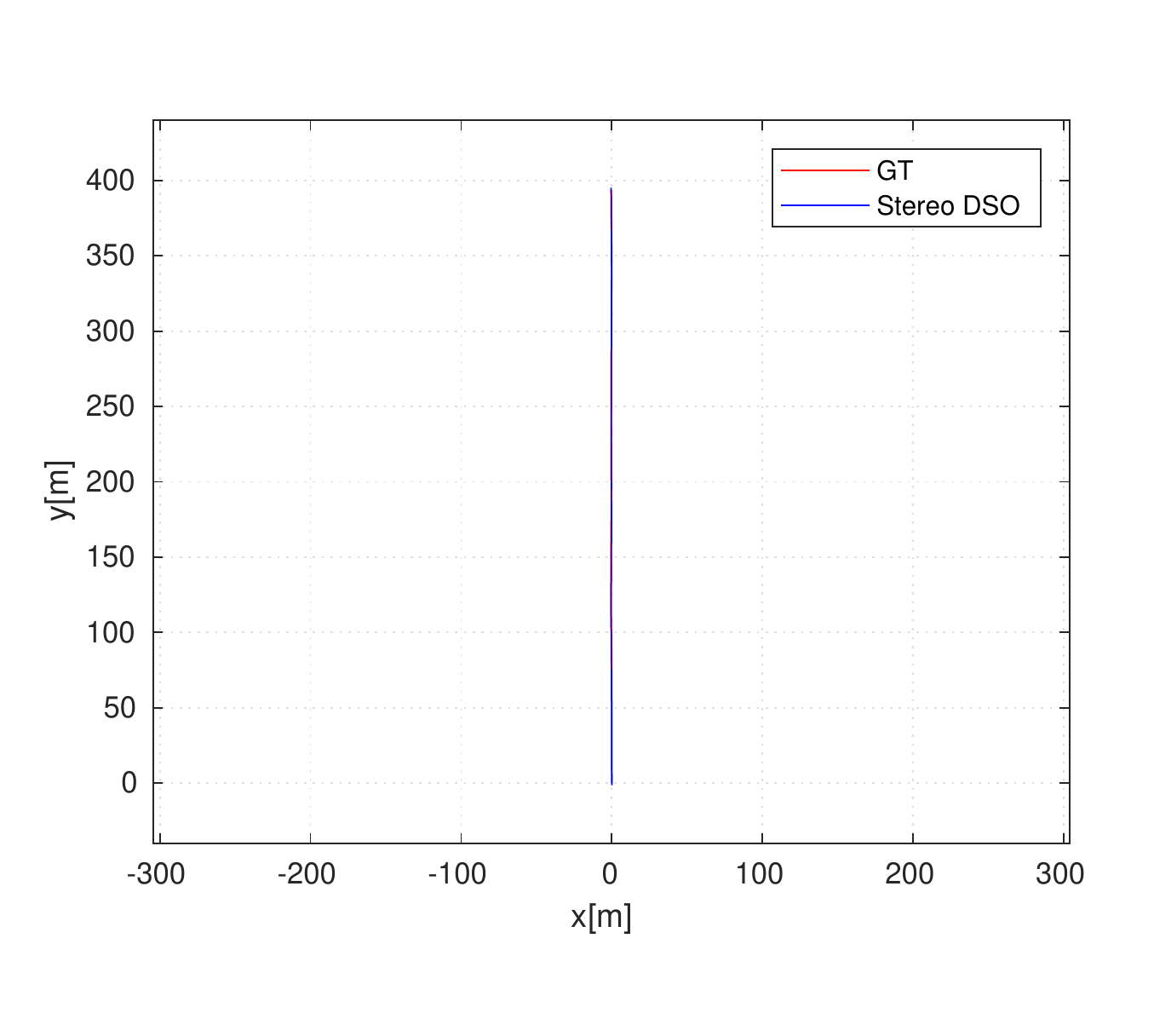}
        \vspace{-1\baselineskip}
        \caption{Seq. 04}
    \end{subfigure} 
    \vspace{-0.25em}

        \begin{subfigure}[]{0.98\textwidth}
        \includegraphics[width=0.32\textwidth]{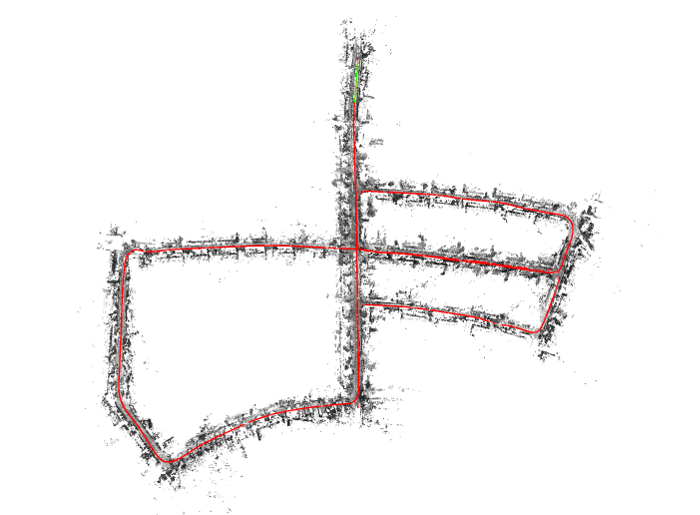}
        \includegraphics[width=0.32\textwidth]{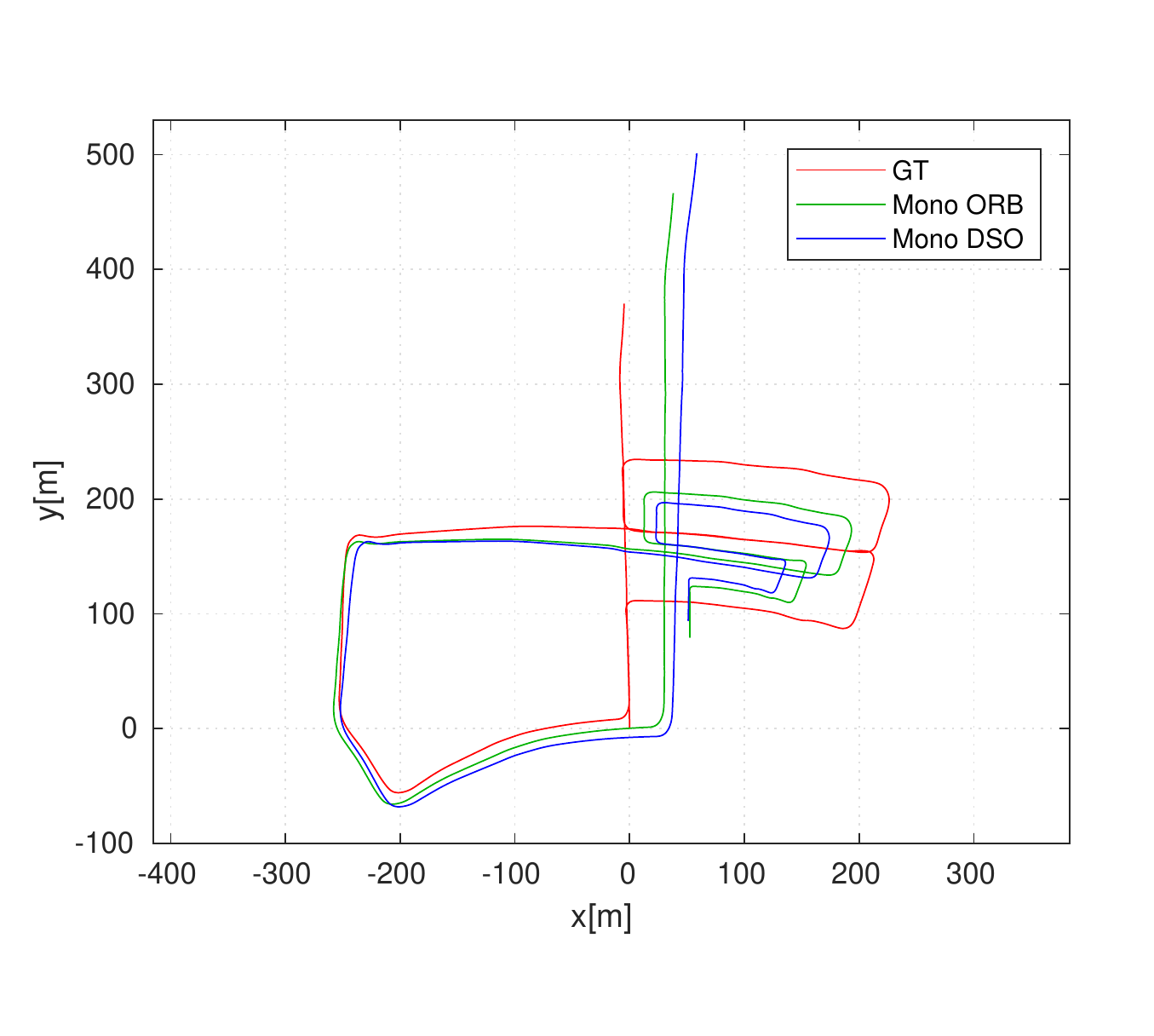}
        \includegraphics[width=0.32\textwidth]{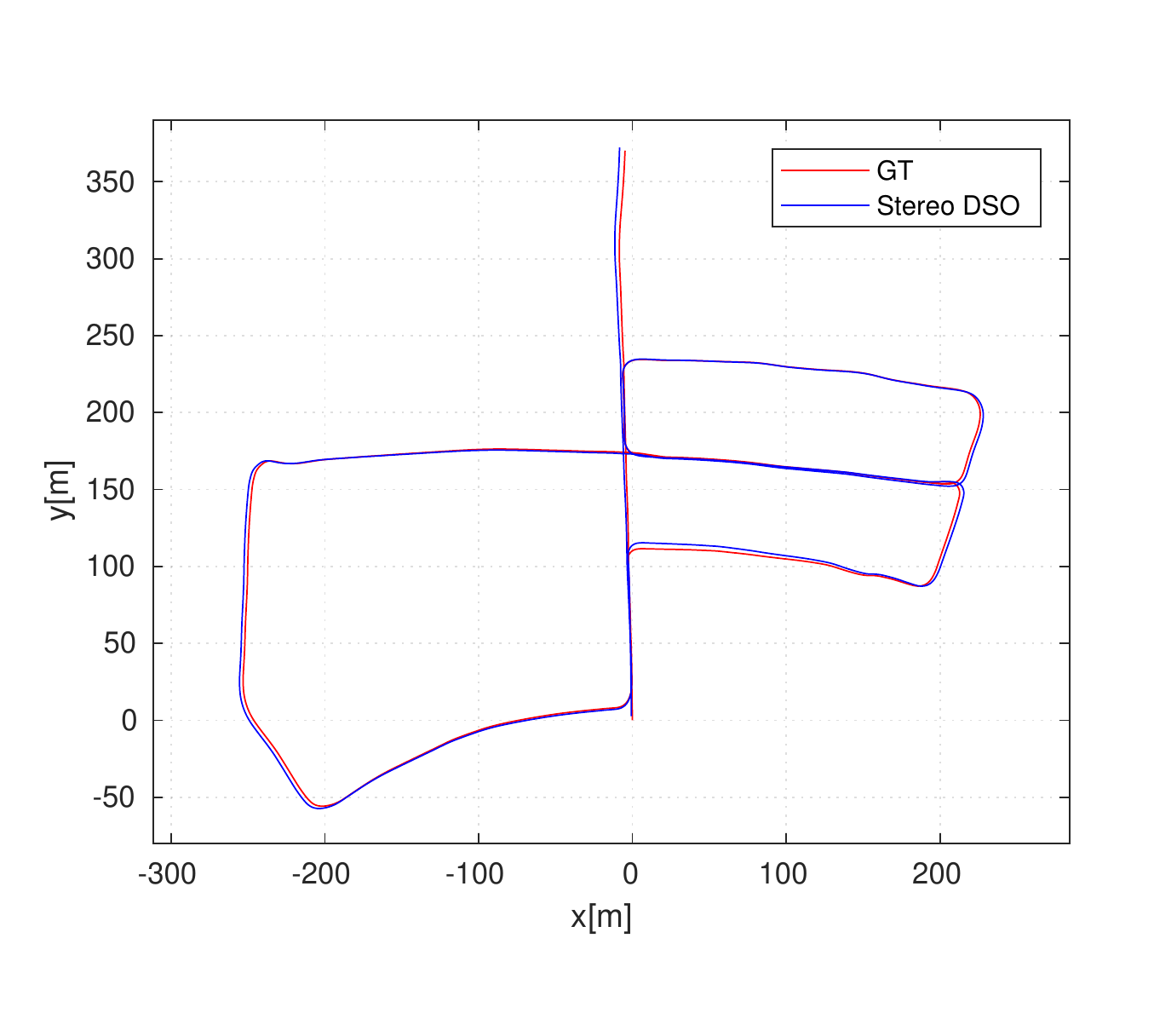}
        \vspace{-1\baselineskip}
        \caption{Seq. 05}
    \end{subfigure} 
    \vspace{-0.25em}

        \begin{subfigure}[]{0.98\textwidth}
        \includegraphics[width=0.32\textwidth]{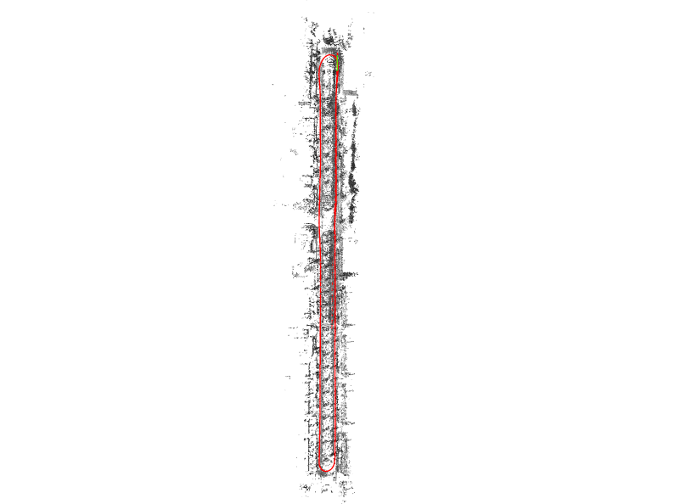}
        \includegraphics[width=0.32\textwidth]{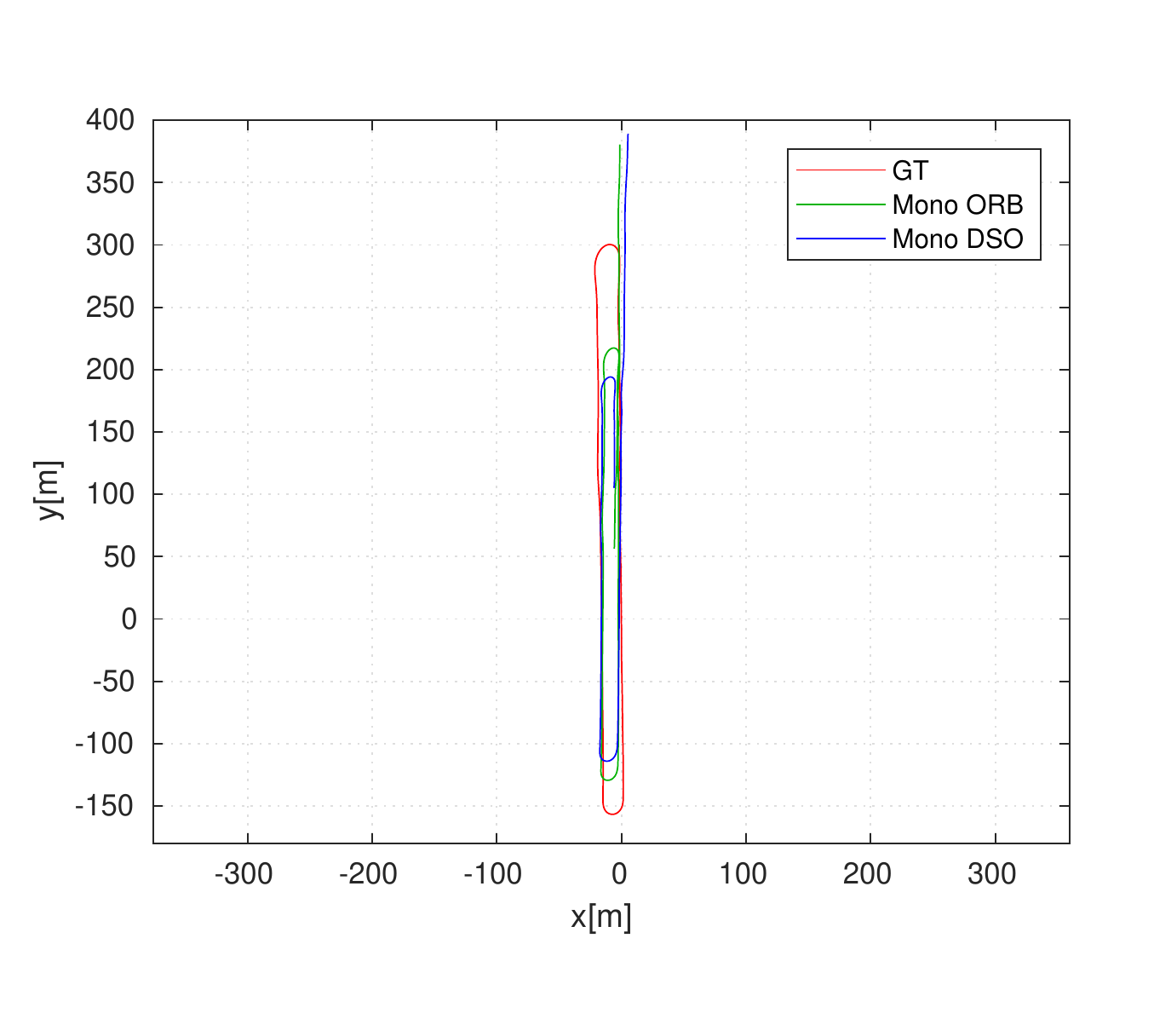}
        \includegraphics[width=0.32\textwidth]{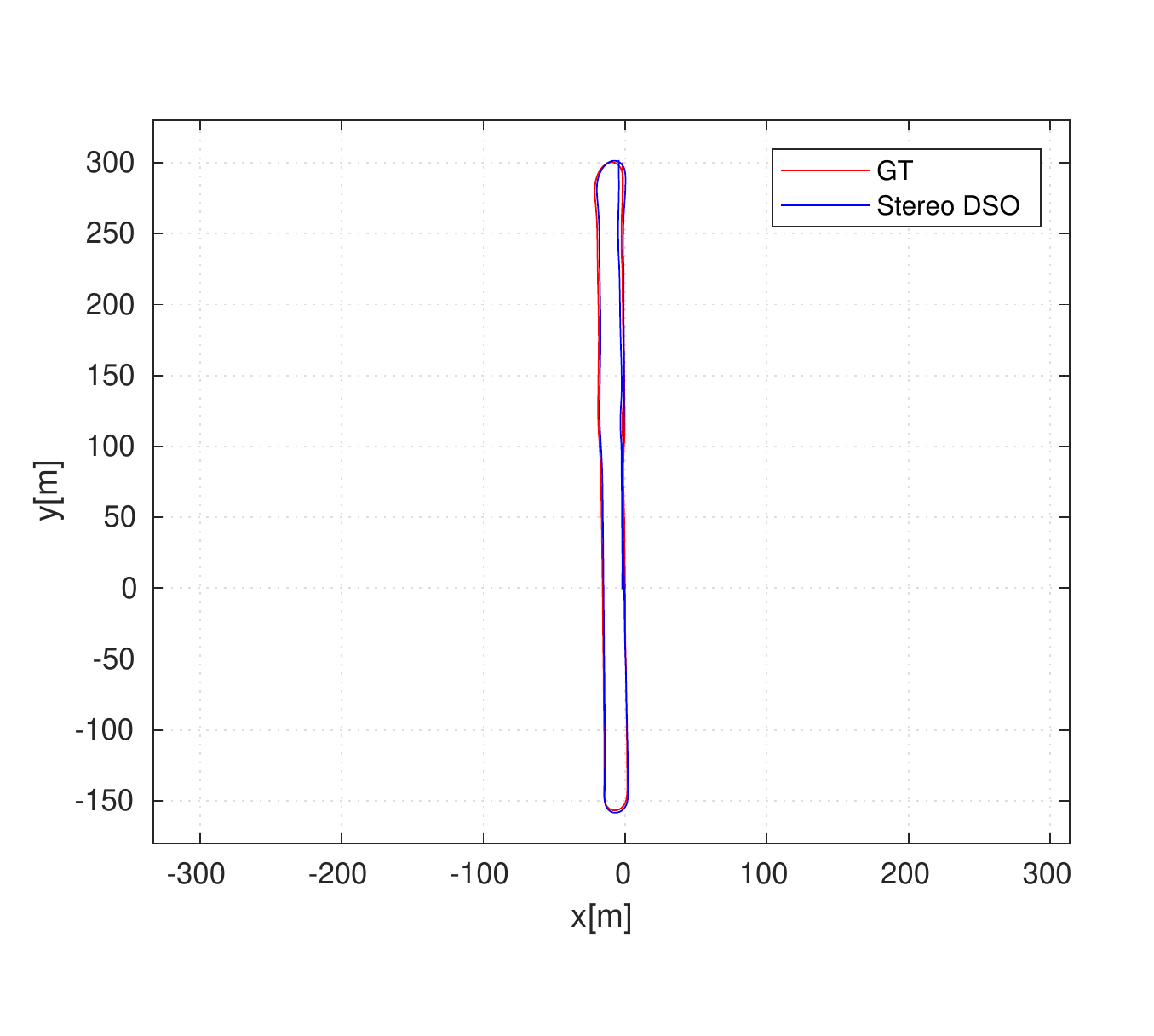}
        \vspace{-1\baselineskip}
        \caption{Seq. 06}
    \end{subfigure} 
    \vspace{-0.25em}

        \begin{subfigure}[]{0.98\textwidth}
        \includegraphics[width=0.32\textwidth]{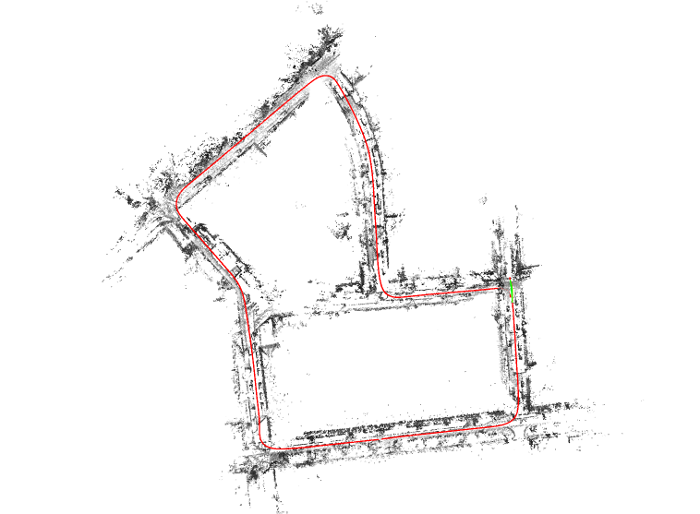}
        \includegraphics[width=0.32\textwidth]{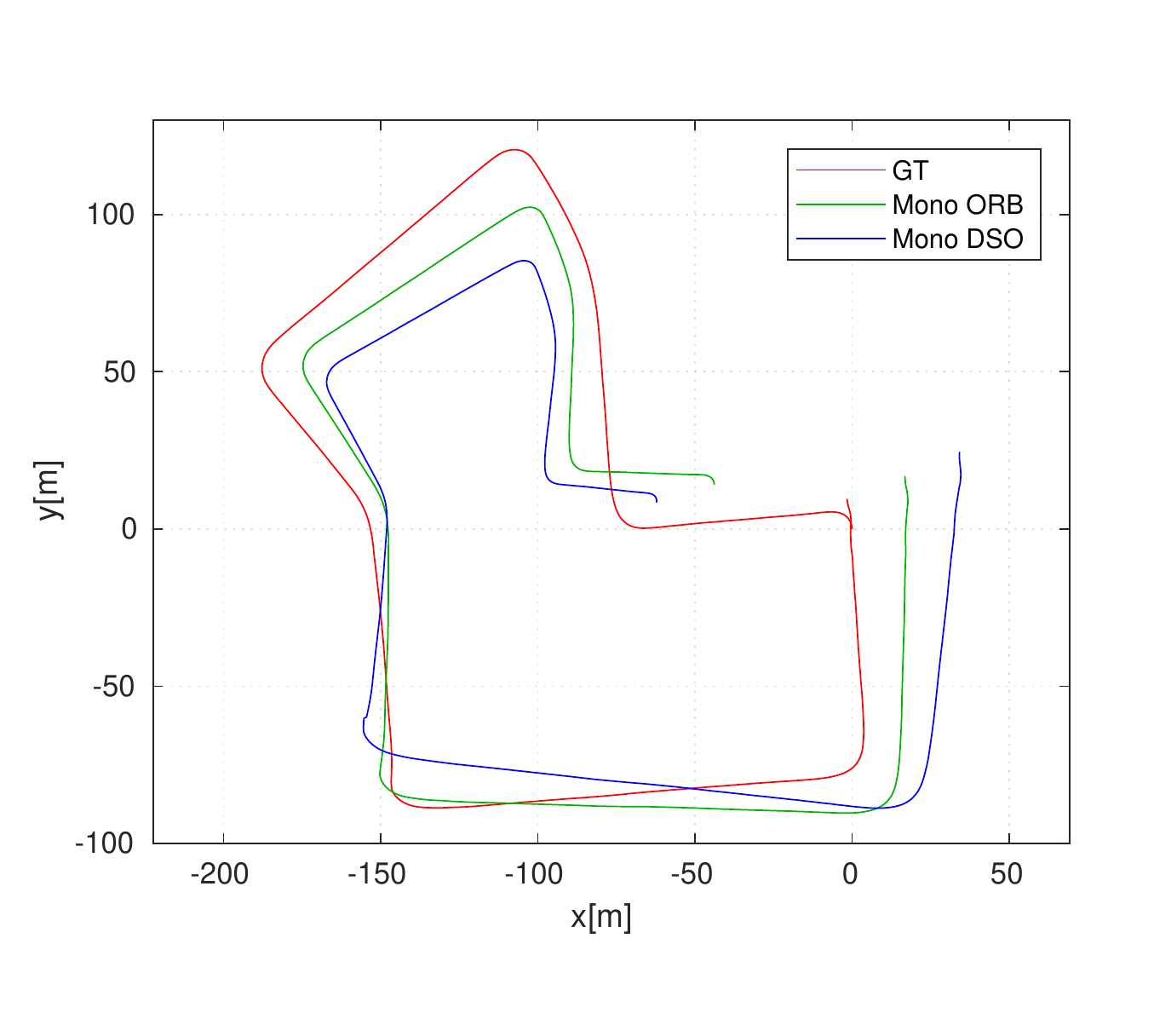}
        \includegraphics[width=0.32\textwidth]{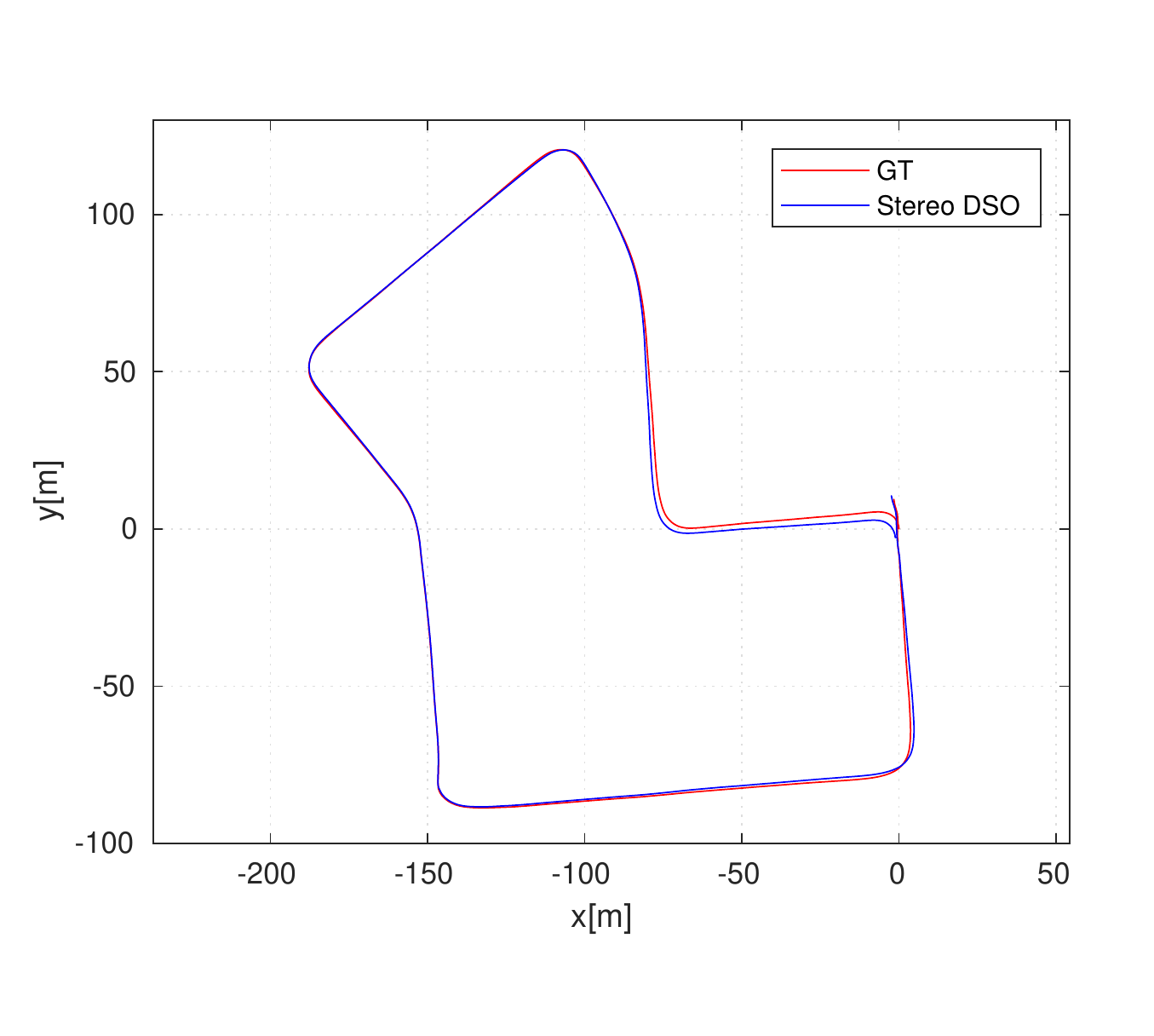}
        \vspace{-1\baselineskip}
        \caption{Seq. 07}
    \end{subfigure} 
\caption{Full results on the KITTI training set (cont.). The trajectories estimated by our VO method
	are shown in the left column. The comparisons to the ground truth are shown in the right
	column. The results of the two state-of-the-art monocular VO methods, namely ORB-SLAM (VO only) and DSO, are shown in the middle. The trajectories are aligned to the ground truth using similarity transformations (7DoF) and rigid-body
	transformations (6DoF) for the monocular methods and our stereo method respectively.}
\end{figure*}

\begin{figure*}\ContinuedFloat
    \centering
        \begin{subfigure}[]{0.98\textwidth}
        \includegraphics[width=0.32\textwidth]{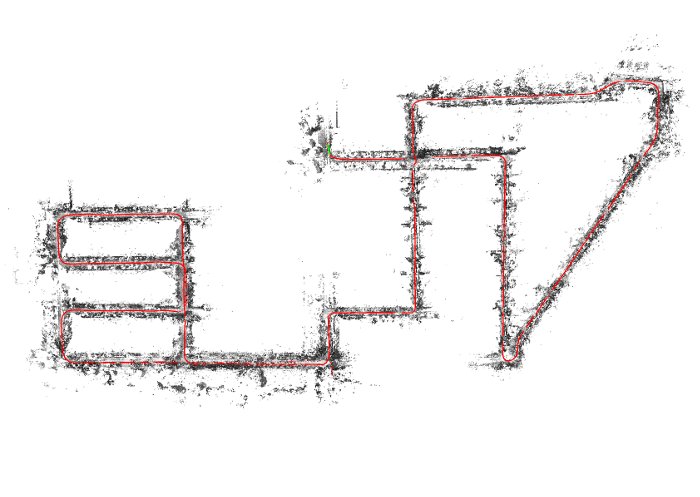}
        \includegraphics[width=0.32\textwidth]{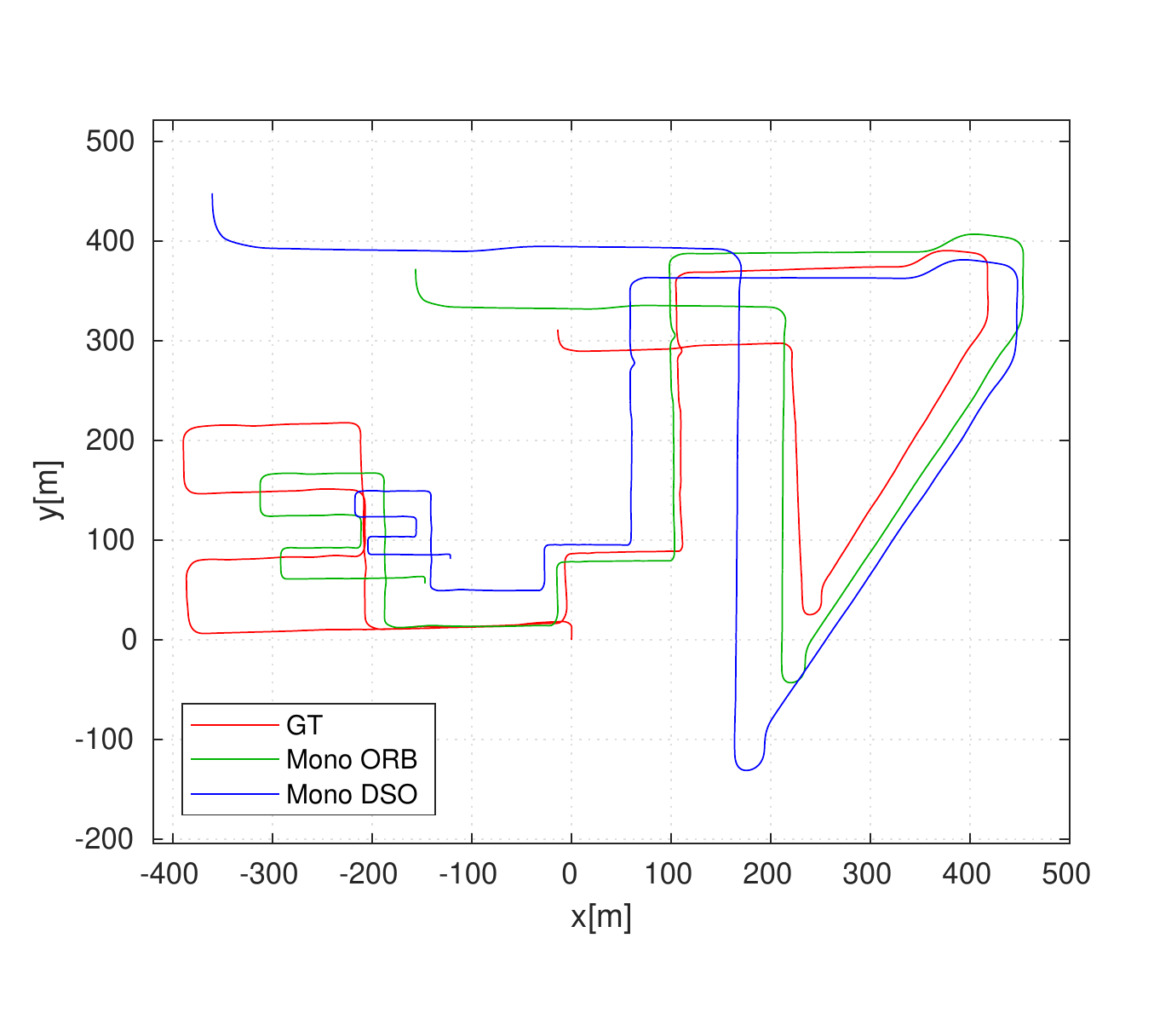}
        \includegraphics[width=0.32\textwidth]{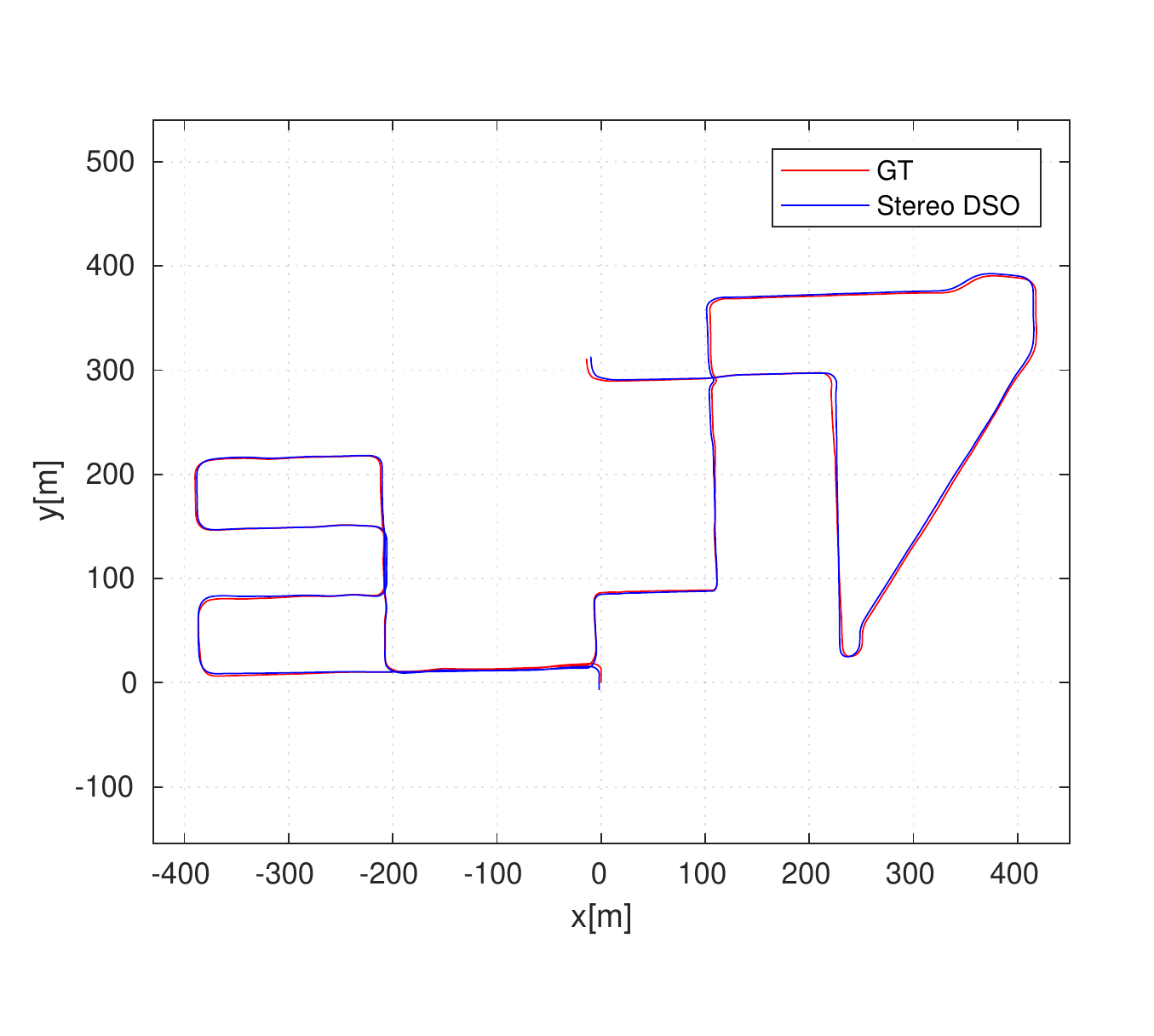}
        \vspace{-1\baselineskip}
        \caption{Seq. 08}
    \end{subfigure} 
    \vspace{-0.25em}

    \begin{subfigure}[]{0.98\textwidth}
        \includegraphics[width=0.32\textwidth]{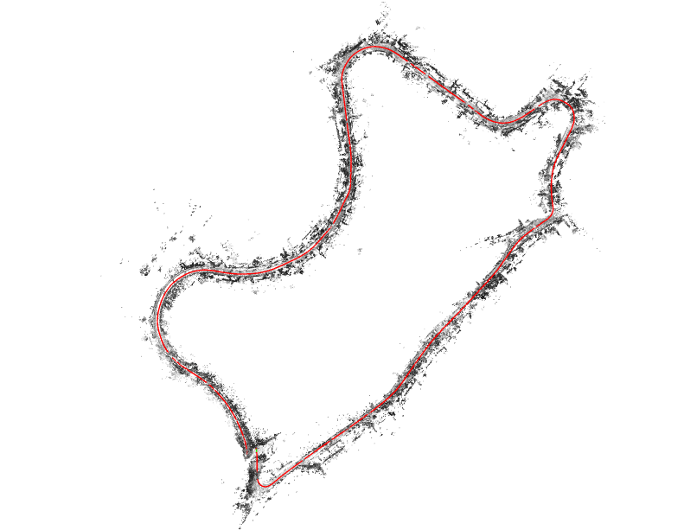}
        \includegraphics[width=0.32\textwidth]{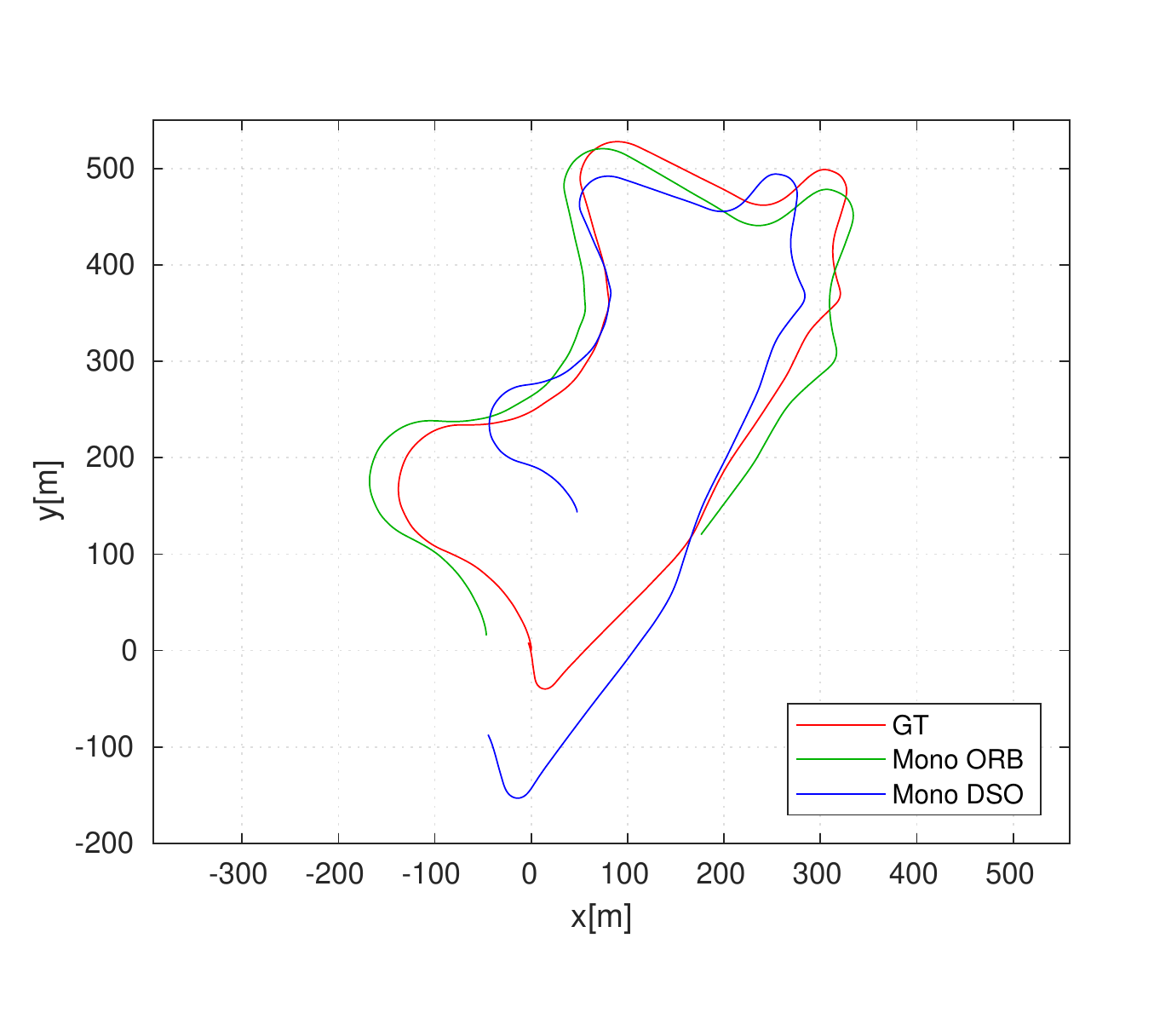}
        \includegraphics[width=0.32\textwidth]{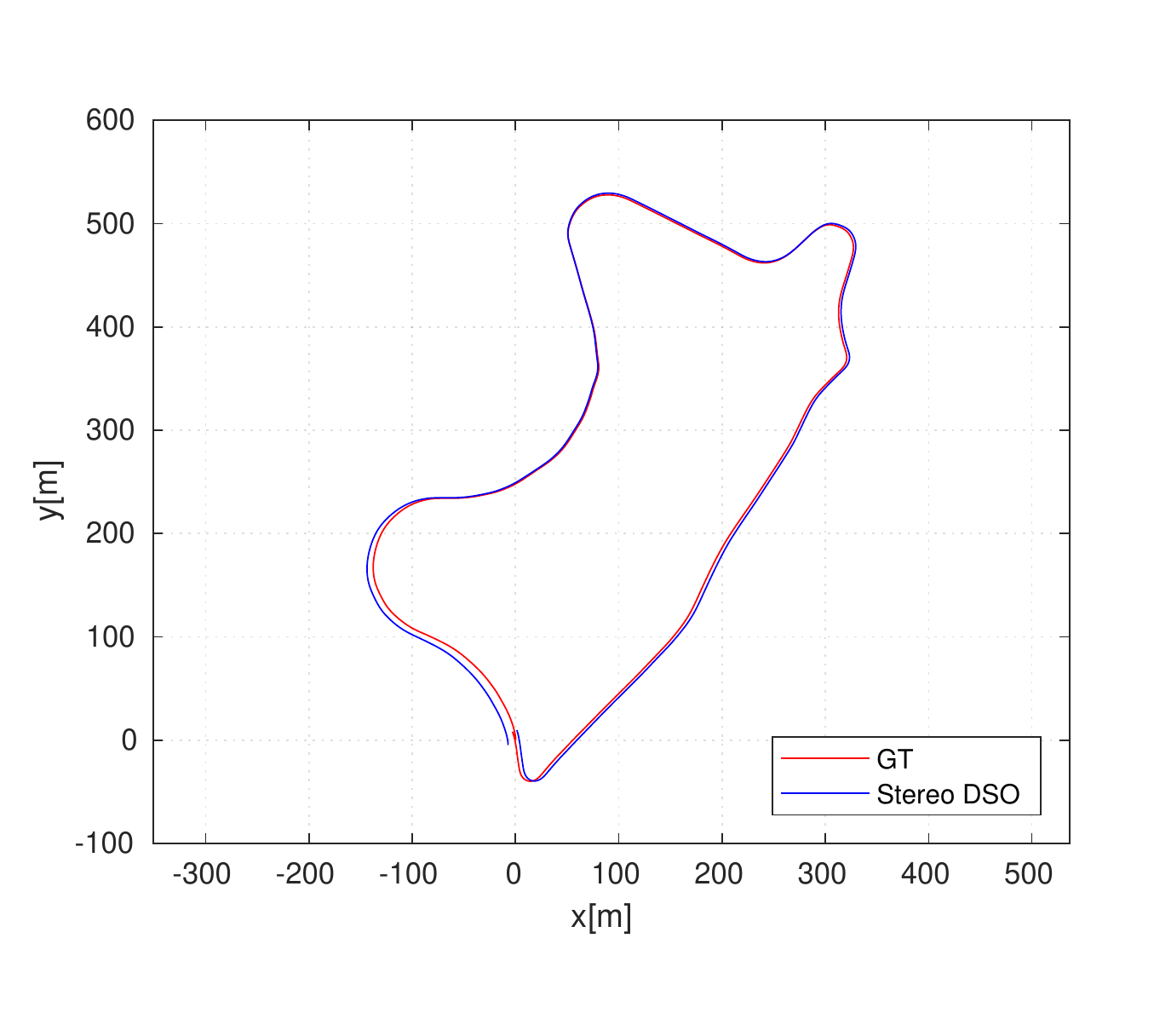}
        \vspace{-1\baselineskip}
        \caption{Seq. 09}
    \end{subfigure} 
    \vspace{-0.25em}

    \begin{subfigure}[]{0.98\textwidth}
        \includegraphics[width=0.32\textwidth]{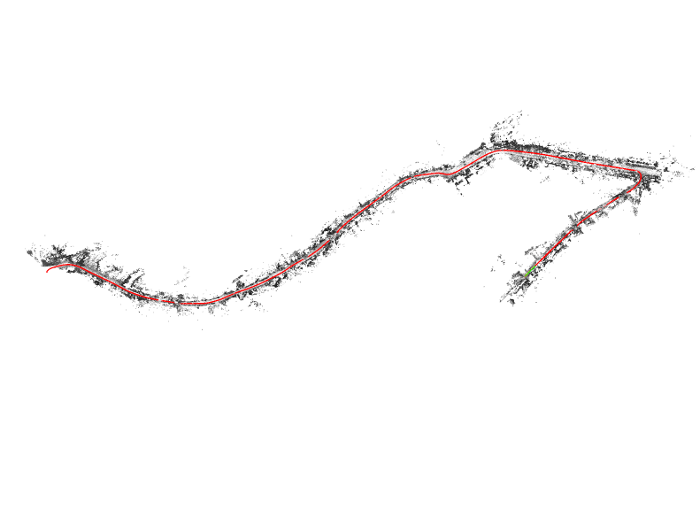}
        \includegraphics[width=0.32\textwidth]{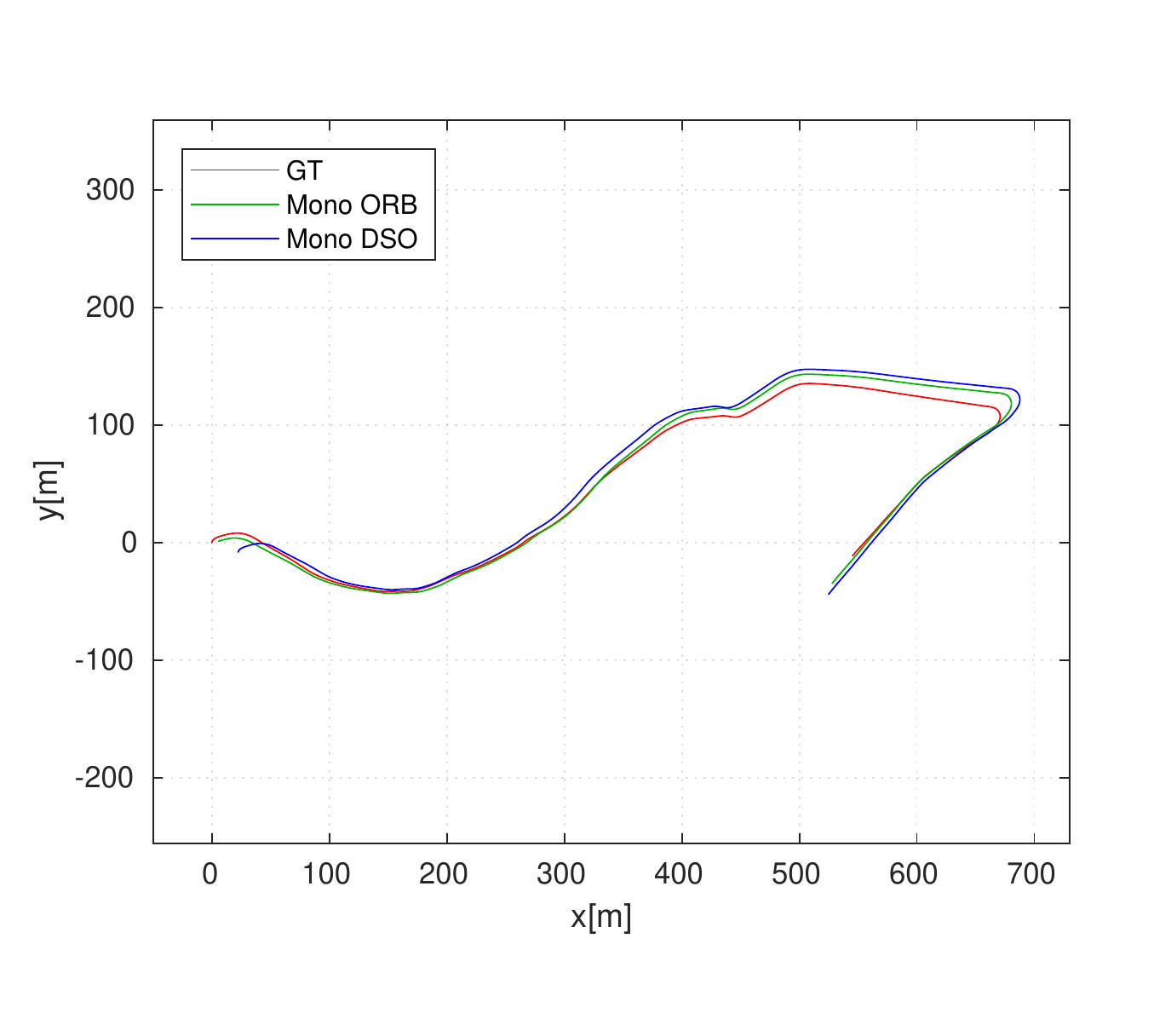}
        \includegraphics[width=0.32\textwidth]{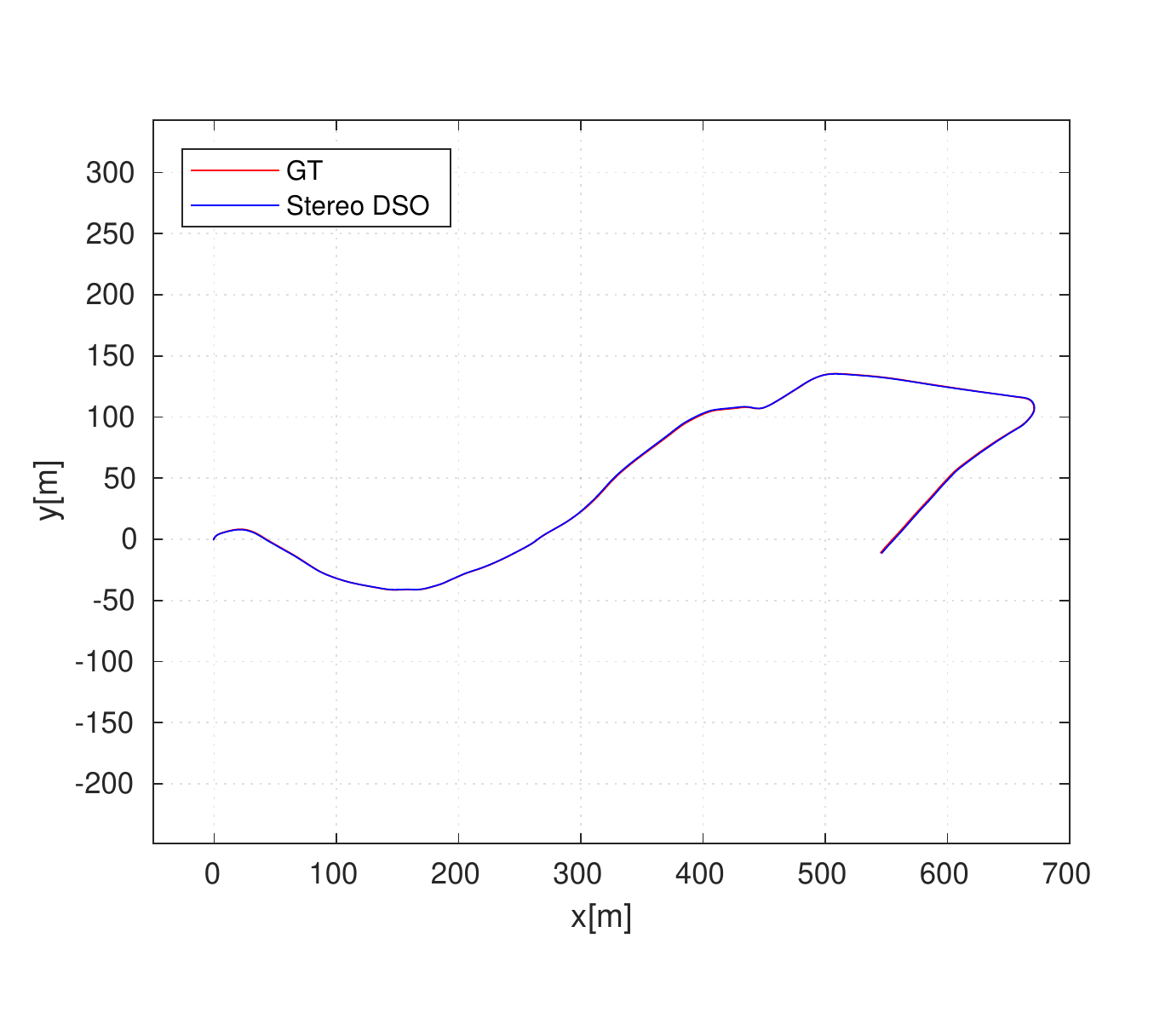}
        \vspace{-1\baselineskip}
        \caption{Seq. 10}
    \end{subfigure} 

   \caption{Full results on the KITTI training set (cont.). The trajectories estimated by our VO method
   	are shown in the left column. The comparisons to the ground truth are shown in the right
   	column. The results of the two state-of-the-art monocular VO methods, namely ORB-SLAM (VO only) and DSO, are shown in the middle. The trajectories are aligned to the ground truth using similarity transformations (7DoF) and rigid-body
   	transformations (6DoF) for the monocular methods and our stereo method respectively.}
\end{figure*}

%As a big advantage of our method over the feature-based methods, precise 3D reconstructions
%can be achieved during the VO approach. In Fig \ref{} we show some further reconstruction
%results on the Frankfurt sequence. As can be seen there, although the reconstructions are
%sparser than the ones from previous dense or semi-dense approaches, they are now much more
%accurate due to the bundle adjustment in the windowed optimization. 

\end{document}